\documentclass[twoside,11pt]{article}

\usepackage{jmlr2e}

\usepackage[utf8]{inputenc} 
\usepackage{url}            
\usepackage{booktabs}       
\usepackage{amsfonts}       
\usepackage{nicefrac}       
\usepackage{microtype}      
\usepackage{color}      

\usepackage{multirow}

\usepackage{comment}
\usepackage{float}
\usepackage{bm,amsmath,amsfonts}
\usepackage[inline]{enumitem}
\usepackage{subcaption}
\usepackage{mathtools}
\def\Re{\mathbb{R}}

\def\Nat{{\rm I\kern\pIR N}}

\newcommand{\EE}[1]{\exptE\left[#1\right]}

\newcommand{\defeq}{\overset{\text{\tiny def}}{=}}

\def\A{{\mathcal{A}}}

\def\H{{\mathcal{H}}}

\def\R{{\mathcal{R}}}
\def\S{{\mathcal{S}}}

\newcommand{\States}{\S}
\newcommand{\Actions}{\A}

\def\vec0{{\boldsymbol{0}}}
\def\vecb{{\boldsymbol{b}}}
\def\vece{{\boldsymbol{e}}}

\def\vecw{{\boldsymbol{w}}}
\def\vecx{{\boldsymbol{x}}}

\def\vecz{{\boldsymbol{z}}}

\newcommand{\ra}{\rightarrow}
\newcommand{\beq}{\begin{equation}}
\newcommand{\eeq}{\end{equation}}
\newcommand{\beqa}{\begin{eqnarray}}
\newcommand{\eeqa}{\end{eqnarray}}
\newcommand{\beqan}{\begin{eqnarray*}}
\newcommand{\eeqan}{\end{eqnarray*}}
\newcommand{\ben}{\begin{eqnarray*}}
\newcommand{\een}{\end{eqnarray*}}

\def\tr{^\top\!}

\def\vece{{\boldsymbol{\bf e}}}
\def\vecw{{\boldsymbol{\bf w}}}
\def\vecz{{\boldsymbol{\bf z}}}
\def\vecx{{\boldsymbol{\bf x}}}
\def\vech{{\boldsymbol{\bf h}}}
\renewcommand{\EE}[2]{\mathbb{E}_{#1\!\!}\left[#2\right]}

\newcommand{\CEE}[3]{\EE{#1}{{#2}~\middle\vert~{#3}}}
\renewcommand{\CEE}[3]{\EE{#1}{{#2}\mid{#3}}}

\def\CEb#1#2{\CEE{b}{#1}{#2}}
\def\CEpi#1#2{\CEE{\pi}{#1}{#2}}

\def\Epi#1{\EE{\pi}{#1}}

\newcommand{\zerovec}{\mathbf{0}}

\def\dtp{\delta_t^\prime}
\def\ztp{\vecz_t^\prime}
\def\dt{\delta_t}
\def\zt{\vecz_t}

\def\ztpp{\vecz_t^\prime}

\newcommand{\vhat}{\hat{v}}
\newcommand{\Glambda}{G^\lambda}

\firstpageno{1}

\begin{document}

\title{Online Off-policy Prediction}




\author{\name Sina Ghiassian \email ghiassia@ualberta.ca 
       \AND
       \name Andrew Patterson \email ap3@ualberta.ca
       \AND
       \name Martha White \email whitem@ualberta.ca 
       \AND
       \name Richard S. Sutton \email rsutton@ualberta.ca 
       \AND
       \name Adam White \email amw8@ualberta.ca \\
       \addr Reinforcement Learning and Artificial Intelligence Laboratory\\
      Department of Computing Science, University of Alberta\\
      Edmonton, Alberta, Canada T6G 2E8\\
}

\editor{}

\maketitle
\vspace{5mm}
\begin{abstract}
This paper investigates the problem of online prediction learning, where prediction, action, and learning proceed continuously as the agent interacts with an unknown environment. The predictions made by the agent are contingent on a particular way of behaving---specifying what would happen if the agent behaved in a particular way---represented as a value function. However, the behavior used to select actions and generate the behavior data might be different from the behavior used to define the predictions, and thus the samples are generated off-policy. The ability to learn behavior-contingent predictions online and off-policy has long been advocated as a key capability of predictive-knowledge learning systems (Sutton et al, 2011, Ring, 2016), but has remained an open algorithmic challenge for decades. The fundamental issue lies with the temporal difference learning update at the heart of most value-function learning algorithms: combining bootstrapping, off-policy sampling, and fixed-basis function approximation may cause the value estimate to diverge to infinity (e.g., Q-learning with linear function approximation). A major breakthrough came with the development of a new objective function that admitted light-weight stochastic gradient descent variants of temporal difference learning. Since then, many sound online off-policy prediction algorithms have been developed, but there has been limited empirical work investigating the relative merits of all these variants. This paper aims to fill these empirical gaps in the literature as well as provide clarity on the key ideas behind each method. We first summarize the large body of literature on off-policy learning, focusing on (1) methods that use computation linear in the number of features and are convergent under off-policy sampling, and (2) other methods which have proven useful with non-fixed, nonlinear function approximation. We provide a large empirical study of online, off-policy prediction methods in two challenging microworlds with fixed-basis feature representations. We report each method's sensitivity to hyper-parameters, update variance, empirical convergence rate, and asymptotic performance, providing new insights that should enable practitioners to successfully extend these new methods to challenging large-scale applications.
\end{abstract}

\begin{keywords}
  Off-policy learning, Policy evaluation, Temporal difference learning,\\ Reinforcement learning, Artificial Intelligence.
\end{keywords}

\section{A short history of off-policy temporal difference learning}
The story of off-policy learning begins with one of the best-known algorithms of reinforcement learning, called Q-learning, and the classic exploration-exploitation tradeoff. 
Off-policy learning poses an elegant solution to the exploration-exploitation tradeoff: the agent makes use of an independent exploration policy to select actions while learning the value function for the optimal policy.
The exploration policy does not maximize reward, but instead selects actions in order to generate data that improves the optimal policy through learning. 
Ultimately, the full potential of Q-learning---and this ability to learn about one policy from a data generated by a totally different exploration---proved limited. Baird's famous counter-example (1995) provided a clear illustration of how, under function approximation, the weights learned by Q-learning can become unstable.\footnote{The action-value star MDP can be found in the errata of Baird's paper (1995).}  Baird's counter-example highlights that divergence can occur when updating off-policy with function approximation and with bootstrapping (as in temporal difference (TD) learning);
even when learning the value function of a fixed target policy.


The instability of TD methods is caused by how we correct the updates to the value function to account for the potential mismatch between the target and exploration policies.  Off-policy training involves estimating the expected future rewards (the value function) that would be observed while selecting actions according to the target policy with training data (states, actions, and rewards) generated while selecting actions according to an exploration policy.
One approach to account for the differences between the data produced by these two policies is based on using importance sampling corrections: scaling the update to the value function based on the agreement between the target and exploration policy at the current state. If the target and exploration policy would select the same action in a state, then they completely agree. Alternatively, if they never take the same action in a state they completely disagree. More generally, there can be degrees of agreement. We call this approach {\em posterior corrections} because the corrections account for the mismatch between policies ignoring the history of interaction up to the current time step---it does not matter what the exploration policy has done in the past.

Another approach, called {\em prior corrections}, uses the history of agreement between the exploration and target policy in the update.
The likelihood that the trajectory could have occurred under the target policy is used to scale the update. The most extreme version of prior corrections uses the trajectory of experience from the beginning of time, corresponding to what has sometimes been referred to as the \emph{alternative life} framework.
Prior and posterior corrections can be combined to achieve stable Off-policy TD updates (Precup et al., 2000), though finite variance of the updates cannot be guaranteed (Precup et al., 2001). The perceived variance of these updates, as well as a preference for the excursions framework discussed below, led to a different direction years later for obtaining sound off-policy algorithms (Sutton et al., 2009).

Learning about many different policies in parallel has long been a primary motivation for stable off-policy learning, and this usage suggested that perhaps prior corrections are not essential.
Several approaches require learning many value functions or policies in parallel, including approaches based on option models (Sutton, Precup \& Singh, 1999), predictive representations of state (Littman, Sutton \& Singh, 2002; Tanner and Sutton, 2005; Sutton et al., 2011), and auxiliary tasks (Jaderberg et al., 2016). In a parallel learning setting, it is natural to estimate the future reward achieved by following each target policy until termination from the states encountered during training---the value of taking \emph{excursions} from the behavior policy.
When value functions or policies estimated off-policy will be used, they will be used starting from states visited by the behavior policy.
In such a setting, therefore, it is not necessarily desirable to obtain alternative life solutions.


The first major breakthrough came with the formalization of this excursion model as an objective function, which then enabled development of an online stochastic gradient descent algorithm.
The resultant family of {\em Gradient}-TD methods use posterior corrections via importance sampling, and are guaranteed to be stable under function approximation (Sutton et al.,\ 2009). This new excursion objective has the same fixed point as TD, and thus Gradient-TD methods converge to the same solution in the cases for which TD converges. Prior attempts to create an objective function for off-policy learning, namely the mean-squared Bellman error due to Baird (1995), resulted in algorithms that converge to different and sometimes less desirable fixed points (see Sutton \& Barto, 2018 for an in depth discussion of these issues). The Gradient-TD methods have extensions for incorporating eligibility traces (Maei \& Sutton, 2010), non-linear function approximation such as with a neural network (Maei, 2011), and learning optimal policies (Maei \& Sutton, 2010). Although guaranteed stable, the major critiques of these methods are (1) the additional complexity due to a second set of learned parameters, and (2) the variance due to importance sampling corrections.

The second major family of off-policy methods revisits the idea of using prior corrections. The idea is to
incorporate prior corrections, starting only from the beginning of the excursion. In this way, the values of states that are more often visited under the target policy are emphasized, but the high variance of full prior corrections---to the beginning of the episode---is avoided. An incremental algorithm, called {\em Emphatic} TD($\lambda$), was developed to estimates these emphasis weightings (Sutton, Mahmood \& White, 2016), with a later extension to further improve variance of the emphasis weights (Hallak et al., 2015).
These Emphatic-TD methods are guaranteed stable under both on-policy and off-policy sampling with linear function approximation (Sutton, Mahmood \& White, 2016; Yu, 2015; Hallak et al., 2015).

Since the introduction of these methods, several refinements have been introduced, largely towards improving sample efficiency. These include (1) \emph{Hybrid}-TD methods that behave like TD when sampling is on-policy, (2) \emph{Saddlepoint} methods for facilitating application of improved stochastic approximation algorithms and (3) variance reduction methods for posterior corrections, using different eligibility trace parameters.
The Hybrid TD methods can be derived with a simple substitution in the gradient of the excursion objective. The resultant algorithms perform conventional TD updates when data is generated on-policy (Hackman 2012; White \& White, 2016), and are stable under off-policy sampling. Initial empirical studies suggested that TD achieves better sample efficiency than Gradient-TD methods when the data is sampled on-policy, though later evaluations found little difference (White \& White, 2016).

Another potential improvement on Gradient-TD can be derived by reformulating the excursion objective into a saddlepoint problem, resulting in several new methods (Liu et al., 2015; Liu et al., 2016; Du et al., 2017; Touati et al., 2018). This saddlepoint formulation enables use of optimization strategies for saddlepoint problems, including finite sample analysis (Touati et al., 2018) and accelerations (Liu et al., 2015; Du et al., 2017). Though most are applicable to online updating, some acceleration strategies are restricted to offline batch updating (Du et al., 2017). As with the hybrid methods, comparative studies to date remain inconclusive about the advantages of these methods over their vanilla Gradient-TD counterparts (Mahadevan et al., 2014; White \& White, 2016).

Finally, several algorithms have been proposed to mitigate variance from importance sampling ratios in the posterior corrections. High magnitude importance sampling corrections introduce variance and slow learning, dramatically reducing the benefits of off-policy learning. In parallel learning frameworks with many target policies, the likelihood of large importance sampling corrections increases as the number of target policies increases. In practice one might use small stepsizes, or avoid eligibility traces to mitigate the variance. The Retrace algorithm solves this issue by truncating the importance sampling ratio and a bias correction, thus avoiding large updates when the exploration and the target policy differ significantly. This approach can diverge with function approximation (Touati et al., 2018). Nevertheless, Retrace has been used in several deep-learning systems with non-linear function approximation (Munos et al., 2016; Wang, 2016). The Tree Backup algorithm (Precup, 2000) mitigates variance without importance sampling corrections by only using the probability of the selected action under the target policy. Both Retrace and Tree Backup can be viewed as adapting the eligibility trace to reduce variance. The related ABQ algorithm achieves stable off-policy updates without importance sampling corrections by varying the amount of bootstrapping in an action-dependent manner (Mahmood, Yu \& Sutton, 2017). Empirical studies suggest Retrace based deep-learning systems can outperform systems based on Tree Backup and Q-learning. However, more targeted experiments are needed to understand the benefits of these adaptive bootstrapping methods over Gradient and Emphatic-TD methods.

\subsection{Outlining the empirical study}

Our theoretical understanding of off-policy temporal difference learning has evolved significantly, but our practical experience with these methods remains limited. Several variants of Gradient and Emphatic-TD have asymptotic performance guarantees (Mahadevan et al., 2014; Yu 2015, 2016, and 2017), and most of the methods discussed above (besides Tree Backup, V-trace) achieve the slightly weaker standard of convergence in expectation. In practice, stable off-policy methods are not commonly used. Q-learning and other potentially divergent Off-policy TD methods have been at the forefront of recent progress in scaling up reinforcement learning (Mnih et al., 2015; Lillicrap et al., 2015; Munos et al., 2016; Wang et al., 2016; Gruslys et al., 2018; Espeholt et al., 2018). To date, there have been no successful demonstrations of Gradient-TD methods in Atari, or any other large-scale problems. This is largely because these  methods are not well understood empirically and many basic questions remain open. (1) How do methods from the two major families---Gradient and Emphatic-TD---compare in terms of asymptotic performance and speed of learning? (2) Does Emphatic-TD's prior correction result in better asymptotic error, and does its single weight vector result in better sensitivity to its tunable parameters? (3) Does the hypothesis that Hybrid methods can learn faster than Gradient-TD hold across domains? (4) Do posterior correction methods exhibit significant variance compared with V-trace, Tree Backup, and ABQ? (5) What is the best way to incorporate posterior importance sampling corrections?

In this paper, we provide a comprehensive survey and empirical comparison of modern linear off-policy, policy evaluation methods. Prior studies have provided useful insights into some of these methods (Dann et al., 2014; Geist \& Scherrer, 2014; White \& White, 2016). Here we take the next step towards practical and stable online off-policy prediction.

In this paper, we restrict attention to policy evaluation methods and linear function approximation. There are several reasons for this choice.
This setting is simpler and yet still includes key open problems.
Focusing on policy evaluation allows us to lay aside a host of issues including maintaining sufficient exploration and chattering near optimal policies (see Bertsekas 2012, Chapter 6, for an overview).  Another reason for focusing on policy evaluation is that many methods for policy optimization involve evaluation as an intermediate, repeated step; solving policy evaluation better can be expected to lead to better optimization methods. Although many of the recent large-scale demonstrations of reinforcement learning make use non-linear function approximation via artificial neural networks, the linear case requires further treatment. Several recently proposed off-policy methods can diverge, even with linear function approximation. The majority of the methods we consider here have not been extended to the non-linear case, and the extensions are not trivial. Most importantly, conducting empirical comparisons of neural network learning systems is challenging due to extreme parameter sensitivity, and sensitivity to initial conditions. The development of sound methodologies for empirical comparisons of neural network learning systems is still very much in its infancy, and beyond the scope of this paper (Henderson et al., 2017).


\section{Problem Definition and Background}
\label{sct:ProblemDefinition}

We consider the problem of learning the value function for a given policy under the Markov Decision Process (MDP) formalism. The agent interacts with the environment over a sequence of discrete time steps, $t=1, 2, 3, \ldots$. On each time step the agent observes a partial summary of the state $S_t \in \S$ and selects an action $A_t \in \A$.  In response, the environment transitions to a new state $S_{t+1}$, according to transition function $P(S_{t+1} | S_t, A_t)$, and emits a scalar reward $R_{t+1} \in \R$. The agent selects actions according to a stochastic, stationary {\em target policy} $\pi: \S \times \A \rightarrow [0,1]$.

We study the problem of {\em policy evaluation}: the computation or estimation of the expected discounted sum of future rewards for policy $\pi$ from every state. The {\em return} at time $t$, denoted $G_t \in \mathbb{R}$, is defined as the discounted sum of future rewards. The discount factor can be variable, dependent on state: $\gamma: \States \rightarrow [0,1]$, with $\gamma_{t+1} \defeq \gamma(S_t)$.
The return is defined as
\begin{align}
G_t &= R_{t+1} + \gamma_{t+1}R_{t+2} + \gamma_{t+1}\gamma_{t+2}R_{t+3} + \gamma_{t+1}\gamma_{t+2}\gamma_{t+3}R_{t+4} + \hdots \nonumber \\
&= \sum_{k=0}^{\infty} \left({\prod_{i=1}^{k}{\gamma_{t+i}}}\right)R_{t+k+1}.\nonumber
\end{align}
When $\gamma_t$ is constant, we get the familiar return $G_t = R_{t+1} + \gamma R_{t+2} + \gamma^2R_{t+3} + ...$, where we overload $\gamma$ here to indicate a scalar, constant discount. Otherwise, variable $\gamma_t$ can discount per state, including encoding termination when it is set to zero.
The \emph{value function} $v:\S\ra\Re$ maps each state to the expected return under policy $\pi$ starting from that state
\begin{equation}
v_{\pi}(s) \doteq \CEpi{G_t}{S_t=s} = \CEpi{\sum_{k=0}^{\infty} {\gamma^k R_{t+k+1}} }{S_t=s} \text{ , for all $s\in\S$} \label{eq:value}
\end{equation}
where the expectation operator $\Epi{\cdot}$ reflects that the expectation over future states, actions, and rewards uses the distribution over actions given by $\pi$, and the transition dynamics of the MDP.

In this paper, we are interested in problems where the value of each state cannot be stored in a table; instead the agent must approximate the value with a parameterized function.
The approximate value function can have arbitrary form, as long as it is everywhere differentiable with respect to the parameters.
An important special case is when the approximate value function is linear in the parameters and in features of the state. In particular, the current state $S_t$ is converted into feature vector $\vecx_t \in \Re^d$ by some fixed mapping $\vecx: \S \rightarrow \Re^d$. The value of the state can then be approximated with an inner product: $$\hat{v}(s_t, \vecw) \doteq \vecw\tr\vecx_t \approx v(s_t), \hspace{1cm}{\text{for all}}~s_t\in\S
,
$$
where $\vecw\in \Re^d$ is a vector of weights/parameters which are modified by the learning process to better approximate $v_\pi$.  Henceforth, we refer to $\vecw$ exclusively as the \emph{weights}, or weight vector, and reserve the word \emph{parameter}
for variables like the discount-rate and stepsize parameters. Typically the number of components in $\vecw$ is much less than the number of possible states ($d\ll|\S|$), and thus $\hat{v}$ will generalize values across many states in $\S$.

We first describe how to learn this value function for the on-policy setting, where the behavior policy equals the target policy.
Temporal difference learning (Sutton, 1988) is perhaps the best known and most successful approach for estimating $\hat{v}$ directly from samples generated while interacting with the environment. Instead of waiting until the end of a trajectory to update the value of each state, the TD($\lambda$) algorithm adjusts its current estimate of the weights toward the difference between the discounted estimate of the value in the next state and the estimated value of the current state plus the reward along the way:
\begin{equation}\label{eq:delta}
\delta_t \doteq \delta(S_t, A_t, S_{t+1}) \doteq R_{t+1} + \gamma \vecw_t\tr\vecx_{t+1} - \vecw_t\tr\vecx_t
.
\end{equation}
We use the value function's own estimate of future reward as a placeholder for the future rewards defining $G_t$ that are not available on time-step $t+1$. In addition, the TD($\lambda$) algorithm also maintains an eligibility trace vector $\vecz_t \in \Re^d$ that stores a fading trace of recent feature activations. The components of  $\vecw_t$ are updated on each step proportional to the magnitude of the trace vector. This simple scheme allows update information to more quickly propagate in domains when the rewards are often zero, such as a maze with a reward of one upon entering the terminal state and zero otherwise.

The update equations for TD($\lambda$) are straightforward:
\begin{align*}
\vecw_{t+1} \leftarrow& ~\vecw_t + \alpha \delta_t z_t\\
\vecz_t \leftarrow& ~\gamma \lambda \vecz_{t-1}  +\vecx_t
,
\end{align*}
where $\alpha\in\Re$ is the scalar stepsize parameter that controls the speed of learning, and $\lambda\in\Re$ controls the length of the eligibility trace. If $\lambda$ is one, then the above algorithm performs an incremental version of Monte-Carlo policy evaluation. On the other-hand, when $\lambda$ is zero the TD($\lambda$) algorithm updates the value of each state using only the reward and the estimated value of the next state---often referred to as full one-step bootstrapping. In practice, intermediate values of $\lambda$ between zero and one often perform best. The TD($\lambda$) algorithm has been shown to converge with probability one to the best linear approximation of the value function under quite general conditions.



These updates need to be modified for the {\em off-policy case},
where the agent selects actions according to a {\em behavior policy} $b: \States \times \Actions \rightarrow [0,1]$ that is different from the target policy.
The value function for target policy $\pi$ is updated using experience generated from a behavior policy that is {\em off}, away, or distant from the target policy.
For example, consider the most well-known off-policy algorithm, Q-learning. The target policy might be the one that maximizes future discounted reward, while the behavior is nearly identical to the target policy, but instead selects an exploratory action with some small probability. More generally, the target and behavior policies need not be so closely coupled. The target policy might be the shortest path to one or more goal states in a gridworld, and the behavior policy might select actions in each state uniform randomly. The main requirement linking these two policies is that the behavior policy {\em covers} the actions selected by the target policy in each state visited by $b$, that is: $b(a|s) > 0$ for all states and actions in which $\pi(a|s) > 0$.

An important difference between these two settings is in the stability and convergence of the algorithms.
One of the most distinctive aspects of off-policy learning and function approximation is that it has been shown that Q-learning and TD($\lambda$), appropriately modified for off-policy updates, and even Dynamic Programming can diverge (Sutton \& Barto, 2018).
In the next two sections, we will discuss different ways to adapt TD-style algorithms with linear function approximation to the off-policy setting. We will highlight convergence issues and issues with solution quality, and discuss different ways recent algorithms proposed to address these issues.

\section{Off-policy Corrections}
The key problem in off-policy learning is to estimate the value function for the target policy, conditioned on samples produced by actions selected according to the behavior policy. This is an instance of the problem of estimating an expected value under some target distribution from samples generated by some other behavior distribution. In statistics, we address this problem with importance sampling, and indeed most methods of off-policy reinforcement learning use such corrections.

We can either account for the differences between which actions the target policy would choose in each state, or account for which states are more likely to be visited under the target policy. More precisely, there are two distributions that we could consider correcting: the distribution over actions, given the state, and the distribution over states. When observing a transition $(S,A,S',R)$ generated by taking the action according to $b(\cdot | S)$, we can consider correcting the update for that transition so that in expectation it is as if actions were taken according to $\pi(\cdot | S)$. However, these updates would still be different than if we evaluated $\pi$ on-policy, because the frequency of visiting state $S$ under $b$ will be different than under $\pi$. All methods correct for the distribution over actions (posterior corrections), given the state, but several methods correct for the distribution over states (prior corrections) in slightly different ways.

In this section, we first provide an intuitive explanation of the differences between methods that use only posterior correction and those that additionally incorporate prior corrections. We then discuss the optimization objective used by Off-policy TD methods, and highlight how the use of prior corrections corresponds to different weightings in this objective. This generic objective will then allow us to easily describe the differences between the algorithms in Section \ref{sct:Algorithms}.

\subsection{Posterior Corrections}
The most common approach to developing sound Off-policy TD algorithms makes use of posterior corrections based on importance sampling. One of the simplest examples of this approach is {\em Off-policy TD($\lambda$)}. The procedure is easy to implement and requires constant computation per time step, given knowledge of both the target and behavior policies. On the transition from $S_t$ to $S_{t+1}$ via action $A_t$, we compute the ratio between $\pi$ and $b$:
\begin{equation}
\rho_t \doteq \rho(A_t | S_t) \doteq \frac{\pi(A_t|S_t)}{b(A_t|S_t)}
.
\end{equation}
%
These importance sampling corrections are then simply added to the eligibility trace update on each time step:
\begin{align}\label{eq:offTD}
\vecw_{t+1} \leftarrow& ~\vecw_t + \alpha \delta_t \vecz^\rho_t \nonumber \\
\vecz^\rho_t \leftarrow& ~\rho_t(\gamma \lambda \vecz^\rho_{t-1}  +\vecx_t)
,
\end{align}
where $\delta_t$ is defined in Equation \ref{eq:delta}. This way of correcting the sample updates ensures that the approximate value function $\hat{v}$ estimates the expected value of the return as if the actions were selected according to $\pi$.
Posterior correction methods use the target policy probabilities for the selected action to correct the update to the value of state $S_t$ using only the
data from time step $t$ onward. 
Values of $\pi$ from time steps prior to $t$ have no impact on the correction. Combining importance sampling with eligibility trace updates, as in Off-policy TD($\lambda$), is the most common realization of posterior corrections.
\begin{figure}[]
      \centering
      \includegraphics[width=0.4\linewidth]{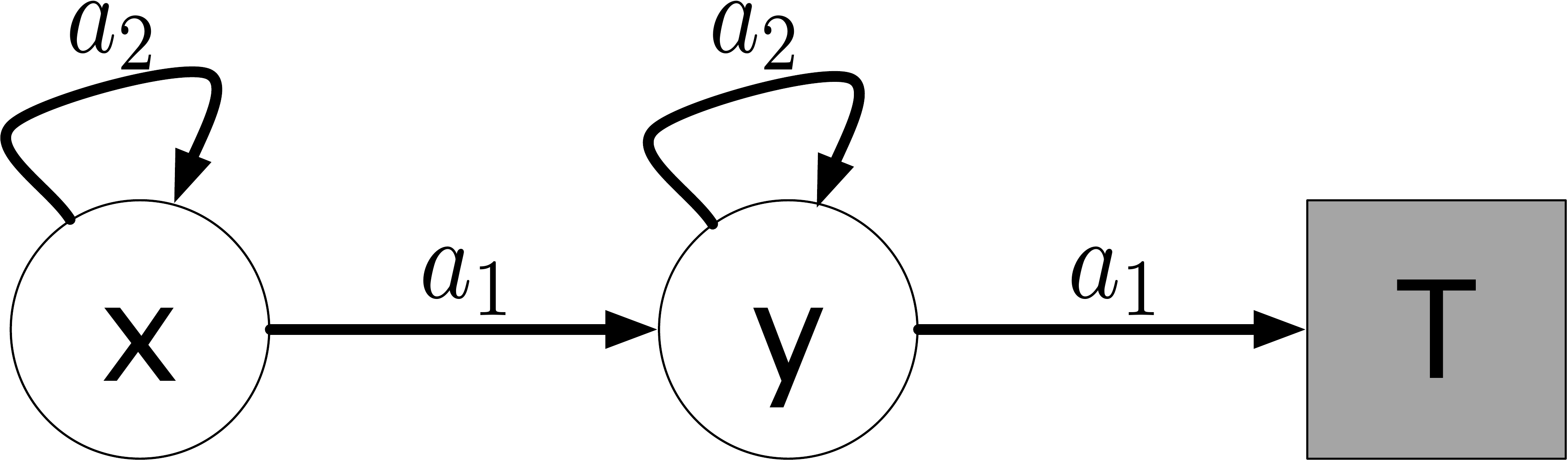}
      \caption{A simple MDP to understand the differences between prior corrections and posterior corrections in Off-policy TD algorithms with importance sampling.}
      \label{fig:SimpleMDP}
\end{figure}

To help understand the implications of posterior corrections, consider the MDP depicted in Figure \ref{fig:SimpleMDP}. Each episode starts in the leftmost state denoted `x' and terminates on transition into the terminal state denoted with `T', and each state is  represented with a unique tabular state encoding: x$: [1,0]$, y$: [0,1]$.
In each state there are two possible actions and the behavior policy chooses each action in each state with 0.5 probability. The target policy chooses action $a_1$ in all states. A posterior correction method like Off-policy TD($\lambda$), will always update the value of a state if action $a_1$ is taken. For example if the agent experiences the transition $y\rightarrow T$, Off-policy TD($\lambda$) will update the value of state $y$; no matter the history of interaction before entering state $y$.

\newcommand{\vhatt}[2]{\hat{v}_{#2}(#1)}

Although the importance sampling corrections product in the eligibility trace update, Off-policy TD($\lambda$) does not use importance sampling corrections computed from prior time-steps to update the value of the current state. This is easy to see with an example. For simplicity we assume $\gamma_t$ is a constant $\gamma \in [0,1)$. Let's examine the updates and trace contents for a trajectory where $b$'s action choices perfectly agree with $\pi$:
$$x\rightarrow y\rightarrow T.$$
After the transition from $x\rightarrow y$, Off-policy TD($\lambda$) will update the value estimate corresponding to $x$:
\begin{align*}
\begin{bmatrix}
           \vhatt{x}{1} \\
           \vhatt{y}{1} \\
         \end{bmatrix} \leftarrow
\begin{bmatrix}
           0 \\
           0 \\
         \end{bmatrix}
+ \alpha\delta_1 \vecz^\rho_1 = \alpha \delta_1
\begin{bmatrix}
           \frac{\pi(a_1|x)}{b(a_1|x)}\gamma\lambda \\
           0 \\
         \end{bmatrix}
,
\end{align*}
where $\hat v_1(x)$ denotes the estimated value of state $x$ on time step $t=1$ (after the first transition), and as usual $\vecz^\rho_0$ and $\hat v$ are initialized to zero. After the second transition, $y\rightarrow T$, the importance sampling corrections will product in the trace, and the value estimates corresponding to both $x$ and $y$ are updated:
\begin{align*}
\begin{bmatrix}
           \vhatt{x}{2} \\
           \vhatt{y}{2} \\
         \end{bmatrix}
\leftarrow
\begin{bmatrix}
           \vhatt{x}{1} \\
           \vhatt{y}{1} \\
         \end{bmatrix}
+ \alpha\delta_2
\begin{bmatrix}
           \frac{\pi(a_1|y)}{b(a_1|y)}\frac{\pi(a_1|x)}{b(a_1|x)}\gamma^2\lambda^2 \\
            \frac{\pi(a_1|y)}{b(a_1|y)}\gamma\lambda \\
         \end{bmatrix}
.
\end{align*}
The estimated value of state $y$ is only updated with importance sampling corrections computed from state transitions that occur after the visit to $y$: using $\frac{\pi(a_1|y)}{b(a_1|y)}$, but not $\frac{\pi(a_1|x)}{b(a_1|x)}$.

Finally, consider another trajectory that deviates from the target policy's choice on the second step of the trajectory:  $$x\rightarrow y \rightarrow y\rightarrow T.$$
On the first transition the value of $x$ is updated as expected, and no update occurs as a result of the second transition. On the third, transition the estimated value of state $x$ is {\em not} updated; which is easy to see from inspecting the eligibility trace on each time-step:
\begin{align*}
\vecz^\rho_1 = \begin{bmatrix}
            \frac{\pi(a_1|x)}{b(a_1|x)}\gamma\lambda  \\
           0
         \end{bmatrix};~
\vecz^\rho_2 = \vec 0;~
\vecz^\rho_3 = \begin{bmatrix}
            0  \\
           \frac{\pi(a_1|y)}{b(a_1|y)}\gamma\lambda
         \end{bmatrix}.
\end{align*}
The eligibility trace is set to zero on time step two, because the target policy never chooses action $a_2$ in state $y$ and thus $\frac{\pi(a_2|y)}{b(a_2|y)} = 0$.
The value of state $S_t$ is never updated using importance sampling corrections computed on time steps prior to $t$.

Many modern off-policy prediction methods use some form of posterior corrections including the Gradient-TD methods, Tree Backup($\lambda$), V-trace($\lambda$), and Emphatic TD($\lambda$). In fact, all off-policy prediction methods with stability guarantees make use of posterior corrections via importance sampling. Only correcting the action distribution, however, does not necessarily provide stable updates, and Off-policy TD($\lambda$) is not guaranteed to converge (Baird, 1995). To obtain stable Off-policy TD($\lambda$) updates, we need to consider corrections to the state distribution as well; as we discuss next.


\subsection{Prior Corrections}
We can also consider correcting for the differences between the target and behavior policy by using the agreement between the two over a trajectory of experience. {\em Prior correction} methods keep track of the product of either $\prod_{k=1}^t \pi(A_k|S_k)$ or $\prod_{k=1}^t \frac{\pi(A_k|S_k)}{b(A_k|S_k)}$, and correct the update to the value of $S_t$ using the current value of the product. Therefore, the value of $S_t$ is only updated if the product is not zero, meaning that the behavior policy never selected an action for which $\pi(A_k|S_k)$ was zero---the behavior never completely deviated from the target policy.

To appreciate the consequences of incorporating these prior corrections into the TD update consider a state-value variant of Precup et al's (2001) Off-policy TD($\lambda$) algorithm:
\begin{align}\label{eq:precup2001}
\vecw_{t+1} \leftarrow& ~\vecw_t + \alpha \delta_t \vecz^\rho_t \nonumber\\
\vecz^\rho_t \leftarrow& ~\rho_t \left(\gamma \lambda \vecz_{t-1}  +\prod_{k=1}^{t-1} \rho_k \vecx_t \right)
\end{align}
where $\vecz^\rho_0 = \zerovec$.
We will refer to the above algorithm as {\em Alternative-life TD($\lambda$)}.
The product in Equation \ref{eq:precup2001} includes all the $\rho_t$ observed during the current episode. Note that experience from prior episodes does not impact the computation of the eligibility trace, as the trace is always reinitialized at the start of the episode.

Now consider the updates performed by Alternative-life TD($\lambda$) using different trajectories from our simple MDP (Figure \ref{fig:SimpleMDP}). If the agent ever selects action $a_2$, then none of the following transitions will result in further updates to the value function. For example, the trajectory $x\rightarrow y\rightarrow y \rightarrow y \cdots y \rightarrow T$ will update $\hat{v}(s)$ corresponding to the first $x \rightarrow y$ transition, but $\hat{v}(y)$ would never be updated due to the product in Equation \ref{eq:precup2001}. In contrast, the Off-policy TD($\lambda$) algorithm described in Equation \ref{eq:offTD} would update $\hat{v}(s)$ on the first transition, and also update $\hat{v}(y)$ on the last transition of the trajectory.

The Alternative-life TD($\lambda$) algorithm has been shown to converge under linear function approximation, but in practice exhibits unacceptable variance (Precup et al., 2001). The Emphatic TD($\lambda$) algorithm, on the other hand, provides an alternative form for the prior corrections, that is lower variance but still guarantees convergence. 
To more clearly explain why, next we will discuss how different prior corrections account for different weightings in optimizing the mean-squared Projected Bellman Error (MSPBE).

\subsection{Objective functions for posterior and prior corrections}
\label{sct:ASecondPerspectiveOnOffPolicyCorrections}

In this section, we describe how different prior corrections, or no prior corrections, correspond to optimizing similar objectives, but with different weightings over the state. This section introduces the notation required to explain all the algorithms, and clarifies convergence properties of algorithms, including which algorithms converge and to which fixed point.

We begin by considering a simplified setting, with $\lambda = 0$, and a simplified variant of the MSPBE, called the NEU (norm of the expected TD update (Sutton, 2009))
%
\begin{align}
\text{NEU}(\vecw) = \Big\|\sum_{s\in\S}d(s)\CEpi{\delta(S, A, S')\vecx(S)}{S=s}\Big\|_2^2
,
\label{eq:onPolicyTDObjective}
\end{align}
where $d: \States \rightarrow [0, \infty)$ is a positive weighting on the states, and
we explicitly write $\delta(S, A, S')$ to emphasize that randomness in the TD-error is due to the underlying randomness in the transition $(S, A, S')$. Equation \ref{eq:onPolicyTDObjective} does not commit to a particular sampling strategy. If the data is sampled on-policy, then $d = d_\pi$, where $d_\pi: \States \rightarrow [0,1]$ is the stationary distribution for $\pi$ which represents the state visitation frequency under behavior $\pi$ in the MDP. If the data is sampled off-policy, then the objective is instead weighted by the state visitation frequency under $b$, i.e., $d = d_b$. As discussed for ETD($\lambda$) in Section \ref{sec:etd}, other weightings $d$ are also possible; for now, we focus on $d = d_\pi$ or $d = d_b$.

We first consider how to sample the NEU for a given a state.
The behavior selects actions in each state $s$, so the update $\delta_t \vecx_t$ needs to be corrected for the action selection probabilities of $\pi$ in state $s$. Importance sampling is one way to correct these action probabilities from a given state $S_t = s$
%
%
%
\begin{align}
\CEpi{\delta(S_t, A_t, S_{t+1}) \vecx(S_t)}{S_t=s}
&=\sum_{a\in\A}\pi(a|s)\sum_{s'\in\S}P(s' | s, a)\delta(s, a, s')\vecx(s)\nonumber\\
&=\sum_{a\in\A}\frac{b(a|s)}{b(a|s)}\pi(a|s)\sum_{s'\in\S}P(s' | s, a)\delta(s, a, s')\vecx(s)\nonumber\\
&=\sum_{a\in\A}b(a|s)\sum_{s'\in\S}P(s' | s, a)\frac{\pi(a|s)}{b(a|s)}\delta(s, a, s')\vecx(s)\nonumber\\
&=\CEb{\rho(A_t| S_t)\delta(S_t, A_t, S_{t+1})\vecx(S_t)}{S_t=s}\label{eq:EqualityOfBehaviorAndTargetExpectations}
.
\end{align}
Therefore, the update $\rho_t \delta_t \vecx_t$ provides an unbiased sample of the desired
expected update $\CEpi{\delta(S_t, A_t, S_{t+1}) \vecx(S_t)}{S_t=s}$.
%
All off-policy methods use these posterior corrections.

%

We can also adjust the state probabilities from $d_b$ to $d_\pi$, using prior corrections.
Alternative-life TD($\lambda$) uses such prior corrections to ask: what would the value be if the data had been generated according to $\pi$ instead of $b$. In such a scenario, the state visitation would be according to $d_\pi$, and so we need to correct both action probabilities in the updates as well as the distribution from which we update. Prior corrections adjust the likelihood of reaching a state. Consider the expectation using prior corrections, when starting in state $s_0$ and taking two steps following $b$:
\begin{align*}
&\CEb{\rho_0 \rho_1 \CEpi{\delta(S_t, A_t, S_{t+1}) \vecx(S_t)}{S_t = S_2}}{S_0 = s_0}\\
&=\CEb{\rho_0 \sum_{a_1 \in \Actions} b(a_1 | S_{1}) \sum_{s_2 \in \States} P(s_2| S_1, a_1) \rho(a_1 | S_1)  \CEpi{\delta(S_t, A_t, S_{t+1}) \vecx(S_t)}{S_t=s_2}}{S_0 = s_0}\\
&=\CEb{\rho_0 \sum_{a_1 \in \Actions} \pi(a_1 | S_1) P(s_1 | S_1, a_1) \CEpi{\delta(S_t, A_t, S_{t+1}) \vecx(S_t)}{S_t=s_2}}{S_0 = s_0}\\
&=\CEb{\rho_0 \CEpi{\delta(S_t, A_t, S_{t+1}) \vecx(S_t)}{S_{t-1}=S_1}}{S_0 = s_0}\\
&=\sum_{a_0 \in \Actions} \pi(s_0, a_0) \sum_{s_1 \in \States} P(s_1 | s_0, a_0) \CEpi{\delta(S_t, A_t, S_{t+1}) \vecx(S_t)}{S_{t-1}=s_1}\\
&=\CEpi{\delta(S_t, A_t, S_{t+1}) \vecx(S_t)}{S_0 = s_0}
.
\end{align*}
%
More generally, we get
\begin{align*}
\mathbb{E}_b \biggl[ \rho_1 \ldots \rho_{t-1} \CEpi{\delta(S_t, A_t, S_{t+1}) \vecx(S_t)}{S_t=s}|S_0 = s_0\biggr ]
&=\CEpi{\delta(S_t, A_t, S_{t+1}) \vecx(S_t)}{S_0 = s_0}
.
\end{align*}
These corrections adjust the probabilities of the sequence from the beginning of the episode to make it as if policy $\pi$ had taken actions $A_1, \ldots, A_{t-1}$ to get to state $S_t$, from which we do the TD($\lambda$) update.

A natural question is which objective should be preferred: the alternative-life ($d \propto d_\pi$) or the excursions objective ($d \propto d_b$). As with all choices for objectives, there is not an obvious answer. The alternative-life objective is difficult to optimize, because prior corrections can become very large or zero---causing data to be discarded---and is high variance. On the other hand, the fixed-point solution to the excursion objective can be arbitrarily poor compared with the best value function in the function approximation class if there is a significant mismatch between the behavior and target policy (Kolter, 2011).
Better solution accuracy can be achieved using an excursion's weighting that includes $d_b$, but additionally reweights to make the states distribution closer to $d_\pi$, as is done with Emphatic TD($\lambda$). We postpone the discussion of this alternative weighting and its corresponding fixed point until after we have properly described Emphatic TD($\lambda$), with the rest of the algorithms in the next section.

\newcommand{\Cmat}{\mathbf{C}}
\newcommand{\Amat}{\mathbf{A}}
\newcommand{\inv}{{-1}}
\newcommand{\sneg}{\mathrm{-}}

The above discussion focused on a simplified variant of the MSPBE with $\lambda = 0$, but the intuition is the same for the MSPBE and $\lambda > 0$. To simplify notation we introduce a conditional expectation operator:
\begin{equation*}
\mathbb{E}_d [Y] = \sum_{s\in\States} d(s) \mathbb{E}_\pi [Y~|~S=s]
.
\end{equation*}
We can now define
\begin{align*}
\Cmat &\doteq \mathbb{E}_d [\vecx(S) \vecx(S)^\top]\\
\Amat &\doteq -\mathbb{E}_d [(\gamma(S') \vecx(S') - \vecx(S)) \vecz(S)^\top]\\
\vecb &\doteq  \mathbb{E}_d [R(S,A,S') \vecz(S)^\top]
\end{align*}
where the eligibility trace $\vecz(S)\in\mathbb{R}^k$ is defined recursively as $\vecz(S) \doteq \vecx(S) + \gamma(S) \lambda \mathbb{E}_\pi[\vecz(S_{t-1}) | S_t = S]$.
We can write the TD($\lambda$) fixed point residual as:
\begin{equation}
\mathbb{E}_d [\delta(S, A, S')\vecz(S)] = \sneg\Amat \vecw + \vecb
\label{eq_td_A}
\end{equation}
so called because $\mathbb{E}_{d_\pi} [\delta(S, A, S')\vecz(S)]=\zerovec$ at the fixed point solution for on-policy TD($\lambda$).
%
The MSPBE can be defined simply, given the definition above:
\begin{equation}
\text{MSPBE}(\vecw) \doteq  (\sneg\Amat \vecw + \vecb)^\top \Cmat^\inv (\sneg\Amat \vecw + \vecb)
.\label{eq:MSPBE}
\end{equation}
The only difference compared with the NEU is the weighted $\ell_2$ norm, weighted by $\Cmat^\inv$, instead of simply $\|\sneg\Amat \vecw + \vecb \|_2^2$.
The extension to $\lambda > 0$ requires that posterior corrections also correct future actions from the state $S$, resulting
in a product of importance sampling ratios in the eligibility trace, as described in the previous section. The conclusions about the choice of state probabilities $d$ in defining the objective, however, remain consistent. In the next section, we discuss how different off-policy methods optimize the different variants of the MSPBE. 

\section{Algorithms}
\label{sct:Algorithms}


In this section, we describe the methods used in the empirical study that follows next. In particular, we discuss the optimization objective, and provide detailed update equations highlighting how prior or posterior corrections are used in each method. We begin with the Gradient-TD family of methods that minimize the excursion variant of the MSPBE. We then discuss modifications on GTD($\lambda$)---namely the Hybrid methods and the Saddlepoint methods. Then we discuss the second family of off-policy methods, the Emphatic methods. We conclude with a discussion of several methods that reduce variance of posterior corrections, using action-dependent bootstrapping. The algorithms, categorized according to weightings, are summarized in Table \ref{tab_methods}.

\definecolor{mygray}{gray}{0.6}

\begin{table}
\begin{tabular}{ |p{2cm}||p{4cm}|p{8cm}|  }
 \hline
 & \multicolumn{2}{|c|}{Objective function weight $d$ includes} \\
 \hline
  & $d_\pi$ & $d_b$\\
 \hline
Posterior corrections
&
\begin{enumerate}[wide, topsep=0pt, itemsep=0pt]
      \item[] N/A. Alternative life algorithms cannot only do posterior corrections.
\end{enumerate}
&
\begin{enumerate}[wide, topsep=0pt, itemsep=0pt]
      \item[] Off-Policy TD($\lambda$)
      \item[] GTD($\lambda$) (Sutton et al., 2010),
      \item[] Hybrid TD($\lambda$) (Maei, 2011; White \& White, 2016)
      \item[] Action-dependent bootstrapping, including Tree Backup($\lambda$) (Precup et al., 2000), V-trace($\lambda$) (Espeholt et al., 2018), AB-Trace($\lambda$) (Mahmood, Yu \& Sutton, 2017)
      \item[] Saddlepoint methods for GTD2($\lambda$), including GTD2-MP($\lambda$) (Liu et al., 2015), SVRG and SAGA for policy evaluation (Du et al., 2017) and Gradient Tree Backup($\lambda$) (Touati et al., 2018)
\end{enumerate} \\
\hline
Prior + Posterior corrections
&
\begin{enumerate}[wide, topsep=0pt, itemsep=0pt]
      \item[] Alternative-life TD($\lambda$)
      \item[] {\color{mygray} Alternative-life GTD($\lambda$), HTD($\lambda$), and Saddlepoint methods}
\end{enumerate}
&
\begin{enumerate}[wide, topsep=0pt, itemsep=0pt]
      \item[] ETD($\lambda$) (Sutton, Mahmood \& White, 2016)
      \item[] ETD($\lambda, \beta$) (Hallak et al., 2015)
      \item[] {\color{mygray} Emphatic GTD($\lambda$), HTD($\lambda$), and Saddlepoint \newline methods}
\end{enumerate}
\\
\hline
\end{tabular}
\caption{A summary of off-policy policy evaluation methods, based on weightings in the objective and whether they incorporate both prior and posterior corrections. The algorithms in grey are hypothetical algorithms that can easily be derived by applying the same derivations as in their original works, but with alternative weightings.}\label{tab_methods}
\end{table}




\subsection[GTD and GTD2]{Gradient Temporal Difference Learning}

Gradient-TD methods were the first to achieve stability with function approximation using gradient descent (Sutton et al., 2009). This breakthrough was achieved by creating an objective function, the MSPBE, and a strategy to sample the gradient of the MSPBE.
The negative of the gradient of the MSPBE, with weighting $d = d_b$, can be written:

%
\begin{align}\label{eq:GradientOfMSPBE}
\nabla \text{MSPBE}(\vecw) &= \mathbb{E}_{d_b}\bigl[\delta(S,A,S') \vecz(S)\bigr] \\
&- \mathbb{E}_{d_b}\bigl[\gamma(S')\vecx(S') \vecx(S)^{\tr}\bigr]  \mathbb{E}_{d_b}\bigl[\vecx(S) \vecx(S)^{\tr}\bigr]^{-1}  \mathbb{E}_{d_b}\bigl[\delta(S,A,S') \vecz(S)\bigr] \nonumber
.
\end{align}
Sampling this gradient is not straightforward due to the product of expectations. To resolve this issue, a second weight vector, $\vech$, can be used to estimate $\mathbb{E}_{d_b}[ \vecx_t\vecx_t^\top]^{-1}\mathbb{E}_{d_b}[\delta_t\vecz_t]$ and avoid the need for two independent samples. The resultant method, called GTD($\lambda$), can be thought of as approximate stochastic gradient descent on the MSPBE and is specified by the following updated equations:
\begin{align}\label{eq:GTD}
\vech_{t+1} \leftarrow& ~\vech_t + \alpha_h\bigl[\delta_t\vecz^\rho_t - (\vech_t^{\tr}\vecx)\vecx_{t+1}  \bigr] \nonumber\\\
\vecw_{t+1} \leftarrow& ~\vecw_t + \alpha \delta_t \vecz^\rho_t - \underbrace{\alpha \gamma_{t+1} (1-\lambda)(\vech_t^{\tr}\vecz^\rho_t)\vecx_{t+1}}_{\text{correction term}}
\end{align}
The GTD($\lambda$) algorithm has several important details that merit further discussion.
The most notable characteristic is the second weight vector $\vech\in\mathbb{R}^k$ that forms a quasi-stationary estimate of the last two terms in the gradient of the MSPBE.
The corresponding two-timescale analysis highlights that the learning rate parameter $\alpha_h\in\mathbb{R}$ should be larger than $\alpha$, where the weights $\vecw$ change slower to enable $\vech$ to obtain such a quasi-stationary estimate (Sutton et al., 2009). In practice, the best values of $\alpha$ and $\alpha_h$ are problem dependent, and the practitioner must tune them independently to achieve good performance (White, 2015; White \& White, 2016). Another important detail is that the first part of the update to $\vecw$ corresponds to Off-policy TD($\lambda$). When $\lambda = 1$, the second term---the correction term---is removed, making GTD(1) = TD(1). Otherwise, for smaller $\lambda$, the correction term plays a bigger role.


The GTD($\lambda$) algorithm has been shown to be stable with linear function approximation. The GTD($\lambda$) with $\lambda=0$, also known as TDC, has been shown to converge in expectation with i.i.d sampling of states (Sutton et al., 2009). The convergence of Gradient-TD methods with $\lambda>0$ was later shown in the Markov noise case with constant stepsize and stepsizes that approach zero in the limit (Yu, 2018).

The GTD2($\lambda$) algorithm is related to GTD($\lambda$), and can be derived starting from the gradient of the excursion MSPBE in Equation \ref{eq:GradientOfMSPBE}. The gradient of the MSPBE given in Equation \ref{eq:GradientOfMSPBE} is an algebraic rearrangement of:
\begin{equation*}
\nabla \text{MSPBE}(\vecw) ~=~\mathbb{E}_{d_b} \bigl[(\vecx(S) - \gamma(S') \vecx(S'))\vecz(S)^{\tr}\bigr]  \mathbb{E}_{d_b}\bigl[\vecx(S) \vecx(S)^{\tr}\bigr]^{-1}  \mathbb{E}_{d_b}\bigl[\delta(S,A,S') \vecz(S)\bigr]
.
\end{equation*}
As before, the last two terms can again be replaced by a secondary weight vector $\vech \in \mathbb{R}^k$. The resultant expression
\begin{equation*}
\mathbb{E}_{d_b} \bigl[(\vecx(S) - \gamma(S') \vecx(S'))\vecz(S)^{\tr}\bigr]  \vech
,
\end{equation*}
can be sampled resulting in an algorithm that is similar to GTD($\lambda$), but differs in its update to the primary weights:
\begin{equation}\label{eq:GTD2}
\vecw_{t+1} \leftarrow ~\vecw_t + \alpha (\vech_t^{\tr}\vecx_t)\vecx_t - \alpha \gamma_{t+1} (1-\lambda)(\vech_t^{\tr}\vecz^\rho_t)\vecx_{t+1}
.
\end{equation}
This update does not make use of the TD-error $\delta_t$, except through the secondary weights $\vech$. The GTD2($\lambda$) algorithm performs stochastic gradient descent on the MSPBE, unlike GTD($\lambda$), which uses an approximate gradient, as we discuss further in Section \ref{sec_saddlepoint} when describing the Saddlepoint methods.

\subsection[Hybrid]{Hybrid TD methods}
The Hybrid TD methods were created to achieve the data efficiency of TD($\lambda$) when data is sampled on-policy, and the stability of Gradient-TD methods when the data is sampled off-policy. Early empirical experience with TD(0) and GTD(0) in on-policy problems suggested that TD(0) might be more sample efficient (Sutton et al., 2009). Later studies highlighted the need for additional empirical comparisons to fully characterize the relative strengths of GTD($\lambda$) compared with TD($\lambda$) (Dann et al., 2014; White \& White, 2016).

\newcommand{\Bmat}{\mathbf{B}}

Hybrid TD methods were first proposed by Maei (2011) and Hackman (2012) and were further developed to make use of eligibility traces by White and White (2016). The derivation of the method starts with the gradient of the excursion MSPBE.
Recall from Equation \eqref{eq:MSPBE} that the MSPBE can be written $(\vecb \sneg\Amat \vecw)^\top \Cmat^\inv (\vecb \sneg\Amat \vecw)$. The matrix $\Cmat$ is simply the weighting in the squared error for $\vecb \sneg\Amat \vecw$. In fact, because we know every solution to the MSPBE satisfies $\vecb \sneg\Amat \vecw = \zerovec$, the choice of $\Cmat$ asymptotically is not relevant, as long as it is positive definite. The gradient of the MSPBE, $\sneg\Amat^\top  \Cmat^\inv (\vecb \sneg\Amat \vecw)$ can therefore be modified to $\sneg\Amat^\top  \Bmat (\vecb \sneg\Amat \vecw)$, for any positive definite $\Bmat$, and should still converge to the same solution(s).


In order to achieve a hybrid learning rule, this substitution must result in an update that reduces to the TD($\lambda$) update when $\pi = b$.
This can be achieved by setting $\Bmat = \mathbb{E}_{d_b} \biggl[ \bigl(\vecx_t - \gamma_{t+1} \vecx_{t+1}\bigr)\vecz_t\biggr]$, which is the $\Amat$ matrix for the behavior. Because this $\Bmat$ is estimated with on-policy samples---since we are following $b$---we know $\Bmat$ is positive semi-definite (Sutton, 1989), and positive definite under certain assumptions on the features. Further, when $b = \pi$, we have that   $\Bmat = \Amat^{-\top}$, giving update
$\sneg\Amat^\top  \Bmat (\vecb \sneg\Amat \vecw) = \vecb \sneg\Amat \vecw$. The TD($\lambda$) update is a stochastic sample of expected update $\vecb \sneg\Amat \vecw$, and so when HTD($\lambda$) uses a stochastic sample of $\sneg\Amat^\top  \Bmat (\vecb \sneg\Amat \vecw)$ when $b = \pi$, it is in fact using the same update as TD($\lambda$).

The HTD($\lambda$) algorithm is:
\begin{align}\label{eq:HTD}
\vech_{t+1} \leftarrow& ~\vech_t + \alpha_h\biggl[\delta_t\vecz^\rho_t - (\vecx_t - \gamma_{t+1}\vecx_{t+1})(\vech_t^{\tr}\vecz_t)  \biggr] \nonumber\\\
\vecw_{t+1} \leftarrow& ~\vecw_t + \alpha\biggl[\delta_t\vecz^\rho_t - (\vecx_t - \gamma_{t+1}\vecx_{t+1})(\vecz^\rho_t -\vecz_t)^{\tr}\vech_t  \biggr]
\end{align}
HTD($\lambda$) has two eligibility trace vectors, with $\vecz$ being a conventional accumulating eligibility trace for the behavior policy. If $\pi=b$, then all the $\rho_t$ are 1 and $\vecz_t = \vecz^\rho_t$, which causes the last term in the $\vecw$ update to be zero and the overall update reduces to the TD($\lambda$) algorithm. The last term in the $\vecw$ update applies a correction to the usual Off-policy TD($\lambda$).

Like GTD($\lambda$), the HTD($\lambda$) algorithm is a posterior correction method that should converge to the minimum of the excursion variant of the MSPBE. No formal stochastic approximation results have been published, though the expected update is clearly convergent because $\Amat^\top \Bmat \Amat$ is positive semi-definite. This omission is likely due to the mixed empirical results achieved with Hybrid TD methods Markov chains and random MDPs (Hackman, 2012; White \& White, 2016).


%
%
%
%

\subsection{Gradient-TD methods based on a saddlepoint formulation}\label{sec_saddlepoint}

Optimization of the MSPBE can be reformulated as a saddle point problem, yielding another family of stable Off-policy TD methods based on gradient descent.
These include the original Proximal-GTD methods (Liu et al., 2015; Liu et al., 2016) methods, stochastic variance reduction methods for policy evaluation (Du et al., 2017), and gradient formulations of Retrace and Tree Backup (Touati et al., 2018).
The MSPBE can be rewritten using convex conjugates:
\begin{equation}\label{eq:saddleMSPBE}
\text{MSPBE}(\vecw) = \min_{\vech} (\vecb -\Amat \vecw)^\top \vech - \tfrac{1}{2} \| \vech \|_\Cmat^2
\end{equation}
where the weighted norm $\| \vech \|_\Cmat^2 = \vech^\top \Cmat \vech$.


The utility of this saddlepoint formulation is that it removes the product of expectations, with the explicit addition of an auxiliary variable.
This avoids the double sampling problem, since for a given $\vech$, it is straightforward to sample $(\vecb -\Amat \vecw)^\top \vech$ (see Equation \eqref{eq_td_A}) with sample $\delta_t {\vecz_t^{\rho}}^\top \vech$. It is similarly straightforward to sample the gradient of this objective for a given $\vech$.
Now this instead requires that this auxiliary variable $\vech$ be learned. The resulting algorithm is identical to GTD2(0) when using stochastic gradient descent for this saddle point problem. This result is somewhat surprising, because GTD2(0) was derived from the gradient of the MSPBE using a quasi-stationary estimate of a proportion of the gradient.

The saddle point formulation---because it is a clear convex-concave optimization problem---allows for many algorithmic variants. For example, stochastic gradient descent algorithm for this convex-concave problem can incorporate accelerations, such as mirror-prox---as used by Liu et al. (2015)---or variance reduction approaches---as used (Du et al., 2017). This contrasts the original derivation for GTD2($\lambda$), which used a quasi-stationary estimate and was not obviously a standard gradient descent technique. One such accelerated algorithm is Proximal GTD2($\lambda$), described by the following update equations:
\begin{align}\label{Proximal GTD2}
\vech_{t+\frac{1}{2}} \leftarrow& ~\vech_{t} + \alpha_\vech \biggl[\delta_{t} \vecz_t^{\rho} - (\vech_{t}^{\tr}\vecx_t)\vecx_t\biggr]\\
\vecw_{t+\frac{1}{2}} \leftarrow& ~\vecw_t +\alpha (\vech_t^{\tr}\vecx_t)\vecx_t -\alpha \gamma_{t+1} (1 - \lambda_{t+1}) (\vech_t^{\tr}\vecz_{t}^{\rho} ) \vecx_{t+1}\\
\delta_{t+\frac{1}{2}} ~\defeq~& R_{t+1} + \gamma_{t+1} \vecw_{t+\frac{1}{2}}^{\tr}\vecx_{t+1} - \vecw_{t+\frac{1}{2}}^{\tr}\vecx_{t}\\
\vech_{t+1} \leftarrow& ~\vech_{t} + \alpha_\vech \biggl[\delta_{t+\frac{1}{2}} \vecz_t^\rho - (\vech_{t+\frac{1}{2}}^{\tr}\vecx_t)\vecx_t\biggr]\\
\vecw_{t+1} \leftarrow& ~\vecw_t +\alpha (\vech_{t+\frac{1}{2}}^{\tr}\vecx_t)\vecx_t -\alpha  \gamma_{t+1} (1 - \lambda_{t+1}) (\vech_{t+\frac{1}{2}}^{\tr}\vecz_{t}^{\rho} ) \vecx_{t+1}
\end{align}
%
The double update to $\vecw$ and $\vech$, denoted by subscripts ${t+\frac{1}{2}}$ and $t+1$, is produced by applying the Stochastic Mirror-Prox acceleration (Juditsky et al., 2011) to the gradient descent update derived from Equation \ref{eq:saddleMSPBE}. We will refer to this algorithm by the shorthand name PGTD2 in the figures.

The saddle point formulation cannot be applied to derive an accelerated version of GTD($\lambda$). 
Recall that GTD($\lambda$) was obtained by reordering expectations in the gradient of the MSPBE, and then using quasi-stationary estimates of different expected values. This alternative formulation cannot obviously be written as a saddle point problem---though it has nonetheless been shown to be convergent. 
Nevertheless, a heuristic approximation of accelerated Proximal GTD($\lambda$) has been proposed (Liu et al., 2015), and its update equations are similar to that of Proximal GTD2($\lambda$) with difference in updating the weight vector $\vecw$:
\begin{align}\label{Proximal GTD}
\vecw_{t+\frac{1}{2}} \leftarrow& ~\vecw_t +\alpha \delta_{t} \vecz_t^{\rho} -\alpha \gamma_{t+1} (1 - \lambda_{t+1}) (\vech_{t}^{\tr}\vecz_{t}^{\rho} ) \vecx_{t+1} \\
\vecw_{t+1} \leftarrow& ~\vecw_t +\alpha \delta_{t+\frac{1}{2}} \vecz_t^\rho - \alpha \gamma_{t+1} (1 - \lambda_{t+1}) (\vech_{t+\frac{1}{2}}^{\tr}\vecz_t^\rho) \vecx_{t+1}
\end{align}
We will refer to this algorithm by the shorthand name PGTD in the figures.

Both Proximal GTD($\lambda$) and Proximal GTD2($\lambda$) minimize the excursion variant of MSPBE, as they assume $d = d_b$.
The idea of the saddlepoint formulation, however, is more general and alternatives weightings could be considered, such as $d = d_\pi$ (shown in Table 1). The expectations in the MSPBE would simply change, and prior corrections would need to be incorporated
to get an unbiased sample of $\vecb -\Amat\vecw$ weighted by $d_\pi$.

The practical utility of these methods for online estimation is still not well understood. Several of the accelerations mentioned above, such as the use of stochastic variance reduction strategies (Du et al., 2017), assume a batch learning setting. The online algorithms, as mentioned, all use variants of GTD2($\lambda$), which seems to perform more poorly than GTD($\lambda$) in practice (Touati et al., 2018). This saddle point formulation, however, does enable continued advances in online convex optimization to be ported to reinforcement learning. Additionally, this formulation allows analysis tools from optimization to be applied to the analysis of TD learning methods. For example,  Touati et al. (2018) provided the first finite sample analysis for GTD2($\lambda$), which is not possible with the original GTD2($\lambda$) derivation based on the quasi-stationary secondary weights. 

%
%


\subsection{Off-policy learning with action-dependent boostrapping}
\label{subsct:ABTDDerivation}

A common concern with using importance sampling ratios is the possibility for high variance, due to large ratios.\footnote{We would like to note that, to the best of our knowledge, variance issues due to importance sampling ratios have not been concretely demonstrated in the literature. This concern, therefore, is based on intuition and should be considered a hypothesis rather than a known phenomenon.}
Several methods have been introduced that control this variance, either by explicitly or implicitly avoiding the product of importance sampling ratios in the traces. The Tree Backup($\lambda$) algorithm, which we call TB($\lambda$), was the first off-policy method that did not explicitly use importance sampling ratios (Precup et al., 2000). This method decays traces more, incurring more bias; newer algorithms such as V-trace($\lambda$) and ABQ($\zeta$) attempt to reduce variance but without decaying traces as much, and improve performance in practice. In this section, we describe the state-value prediction variants of TB($\lambda$), V-trace($\lambda$), and ABQ($\zeta$) that we investigate in our empirical study.

These three methods can all be seen as Off-policy TD($\lambda$) with $\lambda$ generalized from a constant to a function of state and action.
This unification was highlighted by Mahmood et al. (2017) when they introduced ABQ. This unification makes explanation of the algorithms straightforward: each method simply uses a different action-dependent trace function $\lambda: \States \times \Actions \rightarrow [0,1]$. All three methods were introduced for learning action-values; we present the natural state-value variants below.
 %

We begin by providing the generic Off-policy TD algorithm with action-dependent traces.
The key idea is to set $\lambda_t \defeq \lambda(S_{t-1}, A_{t-1})$ such that $\rho_{t-1} \lambda_t$ is well-behaved. The Off-policy TD($\lambda$) algorithm for this generalized trace function can be written\footnote{This update explicitly uses $\rho_t$ in the update to $\vecw_{t+1}$. This contrasts the earlier Off-policy TD updates in Equation \eqref{eq:offTD}, which have $\rho_t$ in the trace. These two forms are actually equivalent, in that the update to $\vecw$ is exactly the same. We show this equivalence in Appendix~\ref{app_importancesampling}. We use this other form here, to more clearly highlight the relationship between $\rho_{t-1}$ and $\lambda_t$.}
\begin{align}
\vecw_{t+1} & = \vecw_t +\alpha \rho_t \delta_t \vecz_t \nonumber \\
\vecz_t & = \gamma_t \rho_{t-1} \lambda_t \vecz_{t-1} + \vecx_{t}\label{eq:OffTDTrace}
,
\end{align}
Now we can specify different algorithms using this generic variant of Off-policy TD($\lambda$), by specifying different implementations of the $\lambda$ function.
Like Off-policy TD($\lambda$), these algorithms all perform only posterior corrections.

TB($\lambda$) is Off-policy TD($\lambda$) with $\lambda_t = b_{t-1} \lambda$, for some tuneable constant $\lambda \in [0,1]$. Replacing $\lambda_t$ with $b_{t-1} \lambda$ in the eligibility trace update in Equation \ref{eq:OffTDTrace} simplifies as follows:
\begin{align}
\vecz_t & = \gamma_t \frac{\pi_{t-1}}{b_{t-1}} b_{t-1}\lambda \vecz_{t-1} + \vecx_{t} \nonumber \label{eq:TBTrace}\\
&=\gamma_t \pi_{t-1}\lambda\vecz_{t-1}+\vecx_{t},
\end{align}
and gives the state-value variant of TB($\lambda$).

A simplified variant of the V-trace($\lambda$) algorithm (Espeholt et al., 2018) can be derived with a similar substitution: $\lambda_t = \min\left(\frac{\bar{c}}{\pi_{t-1}},\frac{1}{b_{t-1}}\right) \lambda b_{t-1}$, where $\bar{c} \in \mathbb{R}^+$ and $\lambda \in [0,1]$ are both tuneable constants. The eligibility trace update becomes:
\begin{align}
\vecz_t & = \gamma_t \min \left(\frac{\bar{c}}{\pi_{t-1}},\frac{1}{b_{t-1}}\right) \lambda b_{t-1} \frac{\pi_{t-1}}{b_{t-1}} \vecz_{t-1} + \vecx_{t} \nonumber\\
&= \gamma_t \min \left(\frac{\bar{c}}{\pi_{t-1}},\frac{1}{b_{t-1}}\right) \lambda \pi_{t-1} \vecz_{t-1} + \vecx_{t} \nonumber\\
&= \gamma_t \min \left(\frac{\bar{c}\pi_{t-1}}{\pi_{t-1}},\frac{\pi_{t-1}}{b_{t-1}}\right) \lambda \vecz_{t-1} + \vecx_{t} \nonumber\\
&= \gamma_t \min \left(\bar{c},\rho_{t-1}\right) \lambda \vecz_{t-1} + \vecx_{t}
\label{eq:V-traceTrace}
,
\end{align}
The parameter $\bar{c}$ is used to cap importance sampling ratios in the trace.
Note that it is not possible to recover the full V-trace($\lambda$) algorithm in this way. The more general V-trace($\lambda$) algorithm uses an additional parameter, $\bar{\rho} \in \mathbb{R}^+$ that caps the $\rho_t$ in the update to $\vecw_{t+1}$: $\min(\bar{\rho},\rho_t) \delta_t \vecz_t$. When $\bar{\rho}$ is set to the largest possible importance sampling ratio, it does not affect $\rho_t$ in the update to $\vecw_t$ and so we obtain the equivalence above. For smaller $\bar{\rho}$, however, V-trace($\lambda$) is no longer simply an instance of Off-policy TD($\lambda$). In the experiments that follow, we investigate this simplified variant of V-trace($\lambda$) that does not cap $\rho_t$ and set $\bar{c}=1$ as done in the original Retrace algorithm.

ABTD($\zeta$) for $\zeta \in [0,1]$ uses $\lambda_t = \nu_{t-1}b_{t-1}$, with the following eligibility trace update:
\begin{align}
\vecz_t & = \gamma_t \frac{\nu_{t-1}}{b_{t-1}} b_{t-1}\lambda \vecz_{t-1} + \vecx_{t} \nonumber \label{eq:ABTDTrace}\\
&=\gamma_t \nu_{t-1}\pi_{t-1}\vecz_{t-1}+\vecx_{t}.
\end{align}
with the following scalar parameters to define $\nu_t$
\begin{align*}
\nu_t &\defeq \nu(\psi(\zeta), s_t, a_t) \defeq \min \left(\psi(\zeta),\frac{1}{\max(b(a_t|s_t),\pi(a_t|s_t))}\right)\\
\psi(\zeta) &\defeq 2\zeta\psi_0 + \max(0,2\zeta - 1) (\psi_{\text{max}} - 2\psi_0)\\
\psi_{0} &\defeq \frac{1}{\max_{s, a}\max(b(a|s),\pi(a|s))}\\
\psi_{\textnormal{max}} &\defeq \frac{1}{\min_{s, a}\max(b(a|s),\pi(a|s))}
.
\end{align*}

The convergence properties of all three methods are similar to  Off-policy TD($\lambda$). They are not guaranteed to converge under off-policy sampling with weighting $d_b$ and function approximation. With the addition of gradient corrections similar to GTD($\lambda$), these algorithms are convergent. For explicit theoretical results, see Mahmood et al. (2017) for ABQ with gradient correction and Touati et al. (2018) for convergent versions of Retrace and Tree Backup.

\subsection[ETD and ETDBeta]{Emphatic-TD learning}
\label{sec:etd}

Emphatic Temporal Difference learning, ETD($\lambda$), provides an alternative strategy for obtaining stability under off-policy sampling without computing gradients of the MSPBE.  The key idea is to incorporate some prior corrections so that the weighting $d$ results in a positive definite matrix $\Amat$. Given such an $\Amat$, a TD($\lambda$) algorithm---a semi-gradient algorithm---can be shown to converge. Importantly, this allows for a stable off-policy algorithm with only a single set of weights. Gradient-TD methods, on the other hand, use two stepsize parameters and two weight vectors to achieve stability.
In this section, we describe two different variants of Emphatic-TD methods: ETD($\lambda$) and ETD($\lambda,\,\beta$), which was introduced to reduce the variance of ETD($\lambda$).




ETD($\lambda$) minimizes a variant of the MSPBE defined in Equation \ref{eq:MSPBE}, where the weighting $d$ is defined based on the {\em followon} weighting. The followon reflects (discounted) state visitation under the target policy when doing excursions from the behavior: starting from states sampled according to $d_b$. The followon is defined as
\begin{equation}
f(s_t)\doteq d_b(s_t) + \gamma(s_t) \sum_{s_{t-1}, a_{t-1}}d_b(s_{t-1})\pi(a_{t-1}|s_{t-1})P(s_t | s_{t-1}, a_{t-1}) + \hdots \label{eq:ETDF}
.
\end{equation}
The emphatic weighting then corresponds to $m(s_t) = d_b(s_t) \lambda + (1-\lambda) f(s_t)$. This is the weighting used in the MSPBE in Equation \ref{eq:MSPBE}, setting $d(s) = m(s)$.

The Emphatic TD($\lambda$) algorithm is specified by the following equations:
\begin{align*}
F_t \leftarrow& ~\rho_{t-1}\gamma_t F_{t-1} + 1\\
M_{t}  \leftarrow& \lambda_t + (1 - \lambda_t)F_t\\
\vecz_t^{\rho} \leftarrow& ~\rho_t \left(\gamma_t \lambda \vecz_{t-1}^{\rho} + M_t \vecx_{t}\right)\\
\vecw_{t+1} \leftarrow& ~ \vecw_t +\alpha \delta_t \vecz_t^{\rho}
,
\end{align*}
\noindent with $F_0=1$ and $\vecz_0^{\rho}=\vec0$. The scalar estimate $F_t$ is used to include the weighting defined in Equation \ref{eq:ETDF}.
 %
 To gain some intuition for this weighting, consider a setting where $\gamma_t =\gamma$ is constant and $\lambda = 0$. Then $M_t = F_t =  \sum_{j=0}^{t}{\gamma^j\prod_{i=1}^{j}\rho_{t-i}} $, giving trace $\vecz_t^{\rho} \leftarrow ~\rho_t \left(\gamma_t \lambda \vecz_{t-1}^{\rho} + \sum_{j=0}^{t}{\gamma^j\prod_{i=1}^{j}\rho_{t-i}} \vecx_{t}\right)$.
There are some similarities to the weighting in the Alternative-life TD($\lambda$) trace in Equation \ref{eq:precup2001}, where $\vecz_t^{\rho} \leftarrow ~\rho_t \left(\gamma_t \lambda \vecz_{t-1}^{\rho} + \prod_{i=1}^{t}\rho_{i} \vecx_{t}\right)$. Both adjust the weighting on $\vecx_t$ to correct for---or adjust---the state distributions. Alternative-Life TD more aggressively downweights states that would not have been visited under the target policy. ETD, on the other hand, reweights based on how frequently the states would be seen when starting $\pi$ as an excursion from $b$.

Emphatic TD($\lambda$) has strong convergence guarantees in the case of linear function approximation. The ETD($\lambda$) under off-policy training has been shown to converge in expectation using the same expected update analysis used to show that TD($\lambda$) converges under on-policy training. Later, Yu (2015) extended this result to show that ETD($\lambda$) converges with probability one. Perhaps more practically relevant, this weighting also resolves the issues raised by Kolter's example (2011). Kolter's example demonstrated that for a particular choice of $\pi$ and $b$, the solution to the MSPBE could result in arbitrarily bad error compared with the best possible approximation in the function class. In other words, even if the true value function can be well approximated by the function class, the off-policy fixed point from the MSPBE with weighting $d = d_b$ can result in an arbitrarily poor approximation to the values. Hallak et al. (2015) showed that the fixed points of the MSPBE with the emphatic weighting, on the other hand, do not suffer from this problem (see their Corollary 1). This result was actually generally shown for an extended ETD($\lambda$) method, called ETD($\lambda, \beta$), which we describe next.

%
%


ETD($\lambda$) was extended to include an additional scalar tuneable parameter $\beta$, to further control variance due to prior corrections. The ETD($\lambda,\,\beta$) algorithm updates are identical to ETD($\lambda$) except for the update to $F_t$:
\begin{equation*}
F_t \leftarrow ~\rho_{t-1}\beta F_{t-1} + 1
\end{equation*}
If $\beta = \gamma$, then the update is identical to ETD($\lambda$). If $\beta = 0$, then the update is identical to Off-policy TD($\lambda$), and there are a spectrum of methods in between. This $\beta$, then, introduces bias to reduce variance in the followon trace for ETD. Hallak et al. showed that $\beta$ can be less than $\gamma$, and ETD($\lambda,\beta$) can still enjoy the convergence properties of ETD($\lambda$), depending on the mismatch between the target and behavior policy. For more similar policies, $\beta$ can be closer to zero and still converge: it can behave like Off-policy TD($\lambda$) and still converge if the setting is almost on-policy. For a greater mismatch, $\beta$ must be nearer to $\gamma$.


\section{Benchmark problems for off-policy, policy evaluation}
\label{sct:TestBeds}

We investigate two simulation problems designed to highlight specific algorithmic properties and fundamental differences between the different algorithms. The first problem is the simplest case of off-policy learning: the objective is to learn the value function for a single target policy from data generated by a different behavior policy. Our second problem highlights the parallel nature of off-policy learning. The agent learns eight independent value functions corresponding to eight unique target policies from the data generated by a random behavior policy. This section describes these problems in detail.

\subsection{Collision problem}
Our first benchmark problem simulates an important real-world use case of off-policy learning: learning about potentially dangerous or destructive situations from partial executions of the target policy. Consider a robot attempting to learn about how likely it is to collide with the wall in the near future if it drove forward from its current location. This prediction can be formulated as a value function and estimated from trajectories of continually driving forward until collision with the wall occurs. It may be more desirable to use a different exploration policy to generate the training data: driving towards the wall, but most of the time retreating before collision occurs. Off-policy training methods can then accurately estimate the probability of termination from the partial executions. Off-policy training allows learning about potentially dangerous situations without needing to frequently encounter these situations.

\begin{figure}[]
      \centering
      \includegraphics[width=0.6\linewidth]{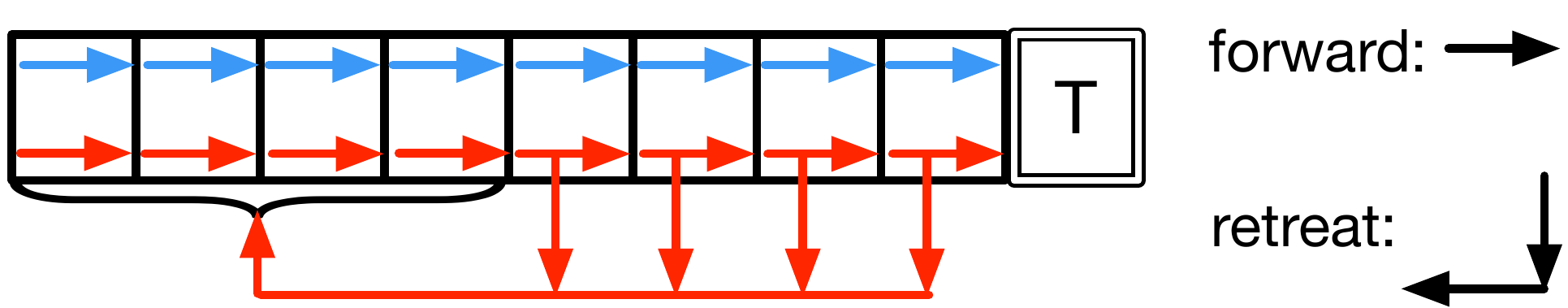}
      \caption{The {\bf Collision} problem benchmark. The target policy (in blue) and the behavior policy (in red) are also shown.}
      \label{fig:ChickenProblemFigure}
\end{figure}

Our first benchmark problem is a simplified version of the collision prediction task described in the previous paragraph. The problem consists of a corridor (one-dimensional grid-world) with eight non-terminal states. The agent starts in one of the first four leftmost states randomly with equal probability ($0.25$ each). There are two actions available in each state: right and retreat. The retreat action causes the agent to abort its forward movement and randomly transition to one of the four leftmost states. The target policy is to always go right. The behavior policy is to go right in the four leftmost states and---in the rightmost states---to go right with $0.5$ probability and retreat  with $0.5$ probability. The discount factor parameter, $\gamma$, is equal to $0.9$ for this task. The reward is zero for all transitions except for taking the right action in the rightmost state, which causes the episode to end (with respect to the behavior policy), and a reward of 1.0 is produced. An eight state version of the Collision problem is shown in Figure \ref{fig:ChickenProblemFigure}.

We used function approximation to solve this problem. At the beginning of every run each state is randomly assigned a unique binary feature vector of length six, where exactly three of the features are one. The feature vectors for each state are held fixed during each run. This feature encoding was chosen because it represents the usual scenario where the values cannot be exactly represented, and there is some undesirable generalization between states.

The Collision problem is of particular interest for comparing methods that use importance sampling. Periodically, the behavior policy will move all the way to the wall and terminate. On each step the importance sampling ratio would be equal to 2.0. Prior correction methods and posterior correction methods product the importance sampling corrections over time, and thus will experience high-magnitude updates when the behavior and target policies align in this way for consecutive time steps. Methods which are sensitive to the magnitude of the importance sampling ratios may require small learning rates to mitigate this variance.

%
%
%

\subsection{Four Rooms Problem}
\begin{figure}[]
      \centering
      \includegraphics[width=0.5\linewidth]{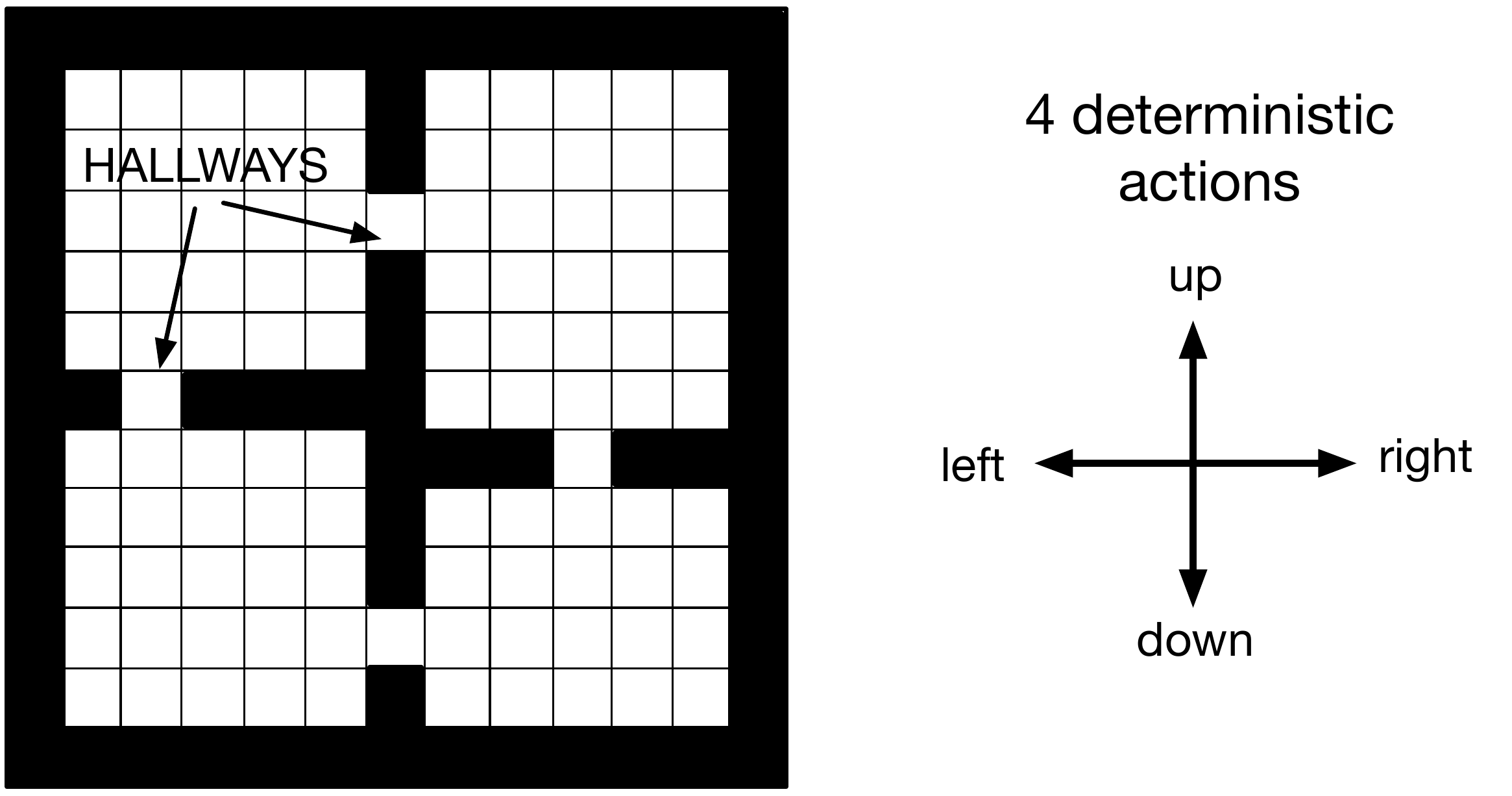}
      \caption{The  world is a grid-world environment with four actions. Arrows point to two of the hallways in the environment.}
      \label{fig:FourRoomsFigure}
\end{figure}

Our second simulation problem helps distinguish different methods when used to train several value functions in parallel. We use a modified version of the Four Rooms Problem described in Sutton et al., (1999). This problem features a grid of states as shown in Figure~\ref{fig:FourRoomsFigure}. Black-colored cells represent walls. There are four deterministic actions available in each state: up, down, left, and right, which transition the agent to the next state in the usual way. If any of these actions take the agent into a wall, then the agent remains in the same state. The only difference between our variant of this problem and the one used by Sutton is that we consider deterministic state transitions given the action. The behavior policy is equiprobable random in all states. Unlike the Collision problem, this problem is continuing. The behavior policy starts in the bottom left corner of the grid and follows the behavior policy thereafter without termination or reset.

There are eight target policies in this task, and they are all fixed. More specifically, two target policies correspond to each room, each of which leads to one of the two hallway states. Each target policy encodes the shortest path to a particular hallway state. In some of the states there is only one action that moves the agent along the shortest path. The reward for each of the eight value functions is one when the target policy reaches the correct hallway state, and zero otherwise. Correspondingly, the discount $\gamma$ is zero when the correct hallway state is reached and 0.9, corresponding to a pesudo or imagined termination in the target policy, but not the behavior (see Sutton et al., 2011). $\gamma$ is zero for all value functions that do not correspond to the current room; once the agent leaves a room, we do not update the weight vector associated with that room anymore.


\begin{figure}[]
      \centering
      \includegraphics[width=0.2\linewidth]{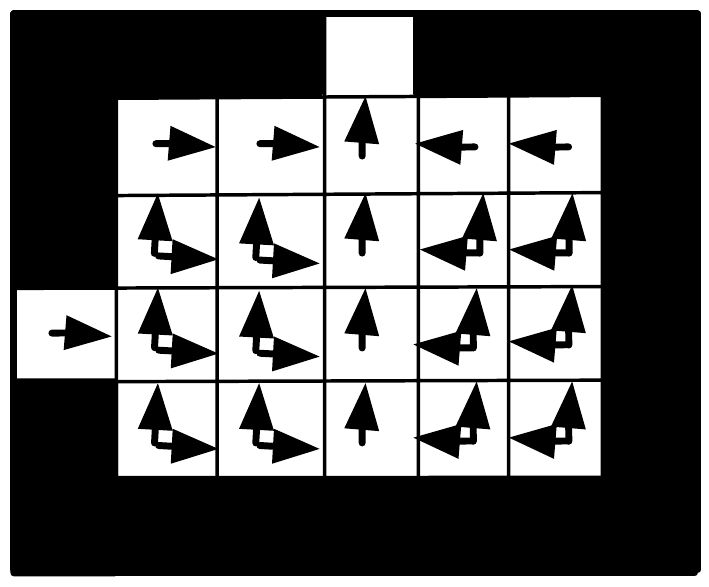}
      \caption{The policy leading to one of the hallways in the bottom right room.}
      \label{fig:FourRoomsOneRoomPolicy}
\end{figure}


We used tile coding (Sutton and Barto, 2018) and linear function approximation to estimate each value function. We used four two by two tilings, producing a coarse representation of the true value function for each policy. When the state representation is nearly perfect, there are little differences in the performance of the methods we investigate in this work. Significant differences only emerge when the representation causes significant aliasing between the states. Tile coding is fast and flexible. It was successfully used with linear and nonlinear function approximation in the past (Sutton, 1996) and it provides the freedom to change the representation easily from a really coarse representation to a very high resolution representation just by changing the number of tilings.

\subsection{Evaluation methodology}
There are many ways to evaluate a learning algorithm, even in the simple policy evaluation case. In order to give multiple perspectives on performance, we use an easily interpretable performance measure---the value error---and present several different visualizations of performance. In the Collision problem, the agent learns a single value function and, thus, the natural performance measure is the root mean-squared value error:
\begin{equation}
{\text{R}}\overline{\text{VE}}(\vecw_t)=\sqrt{\sum_{s\in\S}d_b(s)\left[\hat{v}(s, \vecw_t) - v_\pi(s)  \right]^2}
,
\end{equation}
where $d_b(s)$ is the stationary distribution under the behavior policy, and $v_\pi$ is the true value function---both of which can be estimated from data. $d_b$ was estimated from a batch of ten million temporally contiguous data points collected by following the behavior policy. The true values were computed by executing the target policy once from every state and recording the returns, which provides a representation agnostic estimate of $v_\pi(s)$. 
In the Four Rooms problem, we aggregate the error of each of the eight value functions using a normalized ${\text{R}}\overline{\text{VE}}$ for value function $j$:
\begin{equation*}
{\text{NR}}\overline{\text{VE}}(\vecw^{(j)}_t)=\sqrt{\frac{\sum_{s\in\S}d_b(s)i(s)\left[\hat{v}(s, \vecw_t) - v_\pi(s)  \right]^2}{\sum_{s\in\S}d_b(s)i(s)}}
,
\end{equation*}
where $i(s)$ is the {\em Interest function}, $i: \States \rightarrow [0,1]$, that defines a weighting over states in the error computation. This is only used in the Four Rooms problem. The value error of value function $j$ should only include the errors associated with the states that correspond to its room. Setting $i(s)$ in the error computation ensures that prediction errors from states outside of a room do not contribute to the error computed for value function $j$ and thus will not have an effect on the overall error. We computed a separate ${\text{R}}\overline{\text{VE}}$ for each of the eight value functions, and averaged these to form a single performance measure:
\begin{equation*}
{\text{TR}}\overline{\text{VE}}(\vecw_t)=\frac{1}{8}\sum_{j=1}^8 {\text{NR}}\overline{\text{VE}}(\vecw^{(j)}_t)
.
\end{equation*}
The stationary distribution was estimated from one hundred million samples in the same way as above, and $v_\pi$ was again exactly computed by executing each target policy in every state and recording the returns. Each algorithm was run for 20,000 steps (updating the weights and evaluating the learned value functions on each step) in the Collision problem, and 50,000 steps in the Four Rooms problem.

We performed extensive sweeps over the key parameters of each algorithm. All our results either report the value error with respect to the best performing parameters, or report the performance of each algorithm as a function of one or more of its parameters. Each combination of algorithm and parameter (e.g., TD($\lambda$) with $\alpha =0.1$, and $\lambda=0.9$) was also run for 20,000 steps on the Collision task and 50,000 steps on the Four Rooms problem. Appendicex \ref{app:ParameterSweeps} provide a detailed list of all the parameter values tested. Each experiment was repeated fifty times and the results were averaged.

To determine the best parameter settings, we ranked the performance of each combination using both area under the curve (AUC), and final performance. The AUC is simply the sum total ${\text{R}}\overline{\text{VE}}$ or ${\text{TR}}\overline{\text{VE}}$ on each step of the experiment, divided by the total number of steps. A combination that achieves low AUC typically exhibits fast initial learning, with slightly higher final error at the end of the experiment. We will use the AUC measure to assess the speed of learning---in particular when plotted against time---in a learning curve.
 To determine which combinations achieved the best {\em final performance} we compute the average ${\text{R}}\overline{\text{VE}}$ or ${\text{TR}}\overline{\text{VE}}$ over the last one percent of the steps of the experiment. This measure will be lowest for combinations that achieve low value error consistently for many steps near the end of learning.

For each problem and algorithm combination we present three different plots to provide multiple perspectives on the performance. The first type of graph is the most simple and most common: a learning curve. For each algorithm we plot the value error (${\text{R}}\overline{\text{VE}}$ or ${\text{TR}}\overline{\text{VE}}$) on each time step of the experiment, for the best parameter settings of that algorithm. This results in two graphs. The performance of the best parameter setting according to AUC---which we call {\em early learning performance}---and the best parameter setting according to {\em final performance}; each highlighting different properties of the algorithm. We also include a {\em stepsize plot}, which shows the value error (${\text{R}}\overline{\text{VE}}$ or ${\text{TR}}\overline{\text{VE}}$) obtained by an algorithm for a specific value of the learning rate parameter $\alpha$. All other tunable parameters (e.g., $\lambda$, $\alpha_h$, $\beta$, $\zeta$) are either held to a fixed value or chosen for lowest error, as will be clearly indicated in each figure caption.

The third type of graph is designed to highlight the difficulty in getting each method to perform well in practice, and is new to this work.  A {\em Parameter Sensitivity} plot visualizes the performance of every parameter-algorithm combination tested for a given problem and error measure. The parameter combinations and respective performance are grouped in vertical columns for each algorithm. Each circle denotes the average
value error for a single parameter combination of an algorithm. The circles in each column are randomly offset within the column horizontally,
as many parameter settings may achieve almost identical error. Circles near the bottom of the plot represent low value error, whereas circles arranged in a line in the topmost part of the plot are
parameter combinations that either diverged or exceeded a minimum performance threshold, with the percentage of such parameter combinations given in the graph.

These parameter sensitivity graphs quickly help the reader compare different algorithms along several dimensions. Firstly, the number of circles indicate the number of parameter combinations tested for each algorithm. Algorithms such as ETD(0) only have one tunable parameter $\alpha$. GTD(0) on the other hand has two learning rate parameters and thus more circles on the plot, and thus GTD(0) has more ways to achieve better performance than ETD(0). The groupings of the circles are also informative. A large cluster near the bottom of the plot indicate many parameter settings achieve good performance. Whereas a large percentage of circles above the cutoff indicate that many parameter settings perform poorly. The Parameter Sensitivity plot quickly gives the reader a sense of how difficult it might be to find a good performing set of parameters, if exhaustive parameter search is not practical.

\subsection{Least-squares baselines}
Temporal difference learning methods, when they converge, satisfy the Bellman Equation; a fixed point solution that can be directly computed with Least-squares methods. In fact, TD($\lambda$), Gradient-TD methods, and all methods discussed in Section \ref{sct:Algorithms}
converge to the minimum of the MSPBE, also known as the TD-lambda fixed point. In the case of a fixed basis, we can analytically solve the MSPBE for the weights that satisfy the fixed point equation.
This is called Least-squares Temporal difference learning (LSTD($\lambda$)). The weight vector computed by LSTD($\lambda$) (from a finite batch of training data) represents the weight vector that TD($\lambda$) would converge to with repeated batch presentation of the training data. Different weightings of the MSPBE can be simply incorporated into LSTD($\lambda$). For example, combining LSTD and the emphatic weighting produces the weight-vector that ETD(0) would converge to, given a fixed batch of data. Although, LSTD($\lambda$) can be updated online and incrementally, here we simply use it as a baseline measure, and use the entire batch of training data to compute the weight-vector and compare it against the true weights.
These LSTD($\lambda$) methods are only of interest to us as baselines here, because their computation is quadratic in the number of weights (problematic with large state representations), and they are not applicable to situations where the features are also learned with non-linear transformations, like an Artificial Neural Network.

To compute the LSTD($\lambda$) fixed-point, we first find the value of $\vecw$ for which the fixed-point TD($\lambda$) residual (equation \ref{eq_td_A}) is zero; $\vecw = \Amat^{-1}\vecb$.
The values of $\Amat$ and $\vecb$ can be estimated incrementally using the following update rules:
\begin{align*}
\vece_t &\leftarrow \rho_t\gamma_t(\lambda_t \vece_{t-1} + \vecx_t)\\
\Amat_{t+1} &\leftarrow \Amat_t + \frac{1}{t+1}[\vece_t(\vecx_t - \gamma_{t+1}\vecx_{t+1})^T - \Amat_t]\\
\vecb_{t+1} &\leftarrow \vecb_t + \frac{1}{t+1}[R_{t+1}\vece_t - \vecb_t]
\end{align*}
For each experiment we generate a stream of experience to estimate the values of $\Amat$ and $\vecb$, then compute the weight-vector $\vecw$.
We can similarly find the LSTD($\lambda$) fixed-point with emphatic weighting (denoted as LSETD($\lambda$) in the figures) by simply changing the trace update rule:
\begin{align*}
F_t &\leftarrow \beta \rho_{t-1} F_{t-1} + I_t\\
M_t &\leftarrow \lambda_t  I_t + (1 - \lambda_t) F_t\\
\vece_t &\leftarrow \rho_t(\gamma_t\lambda_t \vece_{t-1} + \vecx_t M_t)
\end{align*}
Finally, we consider the LSTD($\lambda$) fixed-point for Alternative-life TD($\lambda$) (denoted as LSAltTD($\lambda$) in the figures) by changing the LSTD($\lambda$) trace update rule:
\begin{align*}
\vece_t &\leftarrow \left(\gamma_t\lambda_t \vece_{t-1} + \prod^{t-1}_{k=1}\rho_k \vecx_t\right)
\end{align*}
These give unbiased baselines for which we can compare all other methods.

During early learning, the estimate for $\Amat$ may not be well-formed, making the inversion of $\Amat$ unstable.
To avoid this issue---and because our interests lie only in using LSTD($\lambda$) to compute the weight-vector that minimizes the MSBPE on our domains---we choose to only compute $\vecw$ at the end of learning.
We then represent the LSTD($\lambda$) final error as a fixed horizontal line in each learning curve.

\section{Comparing prior and posterior correction methods}
\label{sct:ComparingTheThreeMainFamilies}
We begin our investigation by comparing the two main approaches for achieving stable off-policy updates. Gradient-TD methods and Emphatic-TD methods achieve stability in different ways. Gradient-TD methods achieve off-policy stability with posterior corrections derived from the MSPBE excursion objective function. Emphatic methods, on the other hand, ensure stability in the off-policy case by utilizing both prior and posterior corrections. Our first experiment compares GTD($\lambda$) with ETD($\lambda$).

\begin{figure}[h!]
      \centering
      \includegraphics[width=0.8\linewidth]{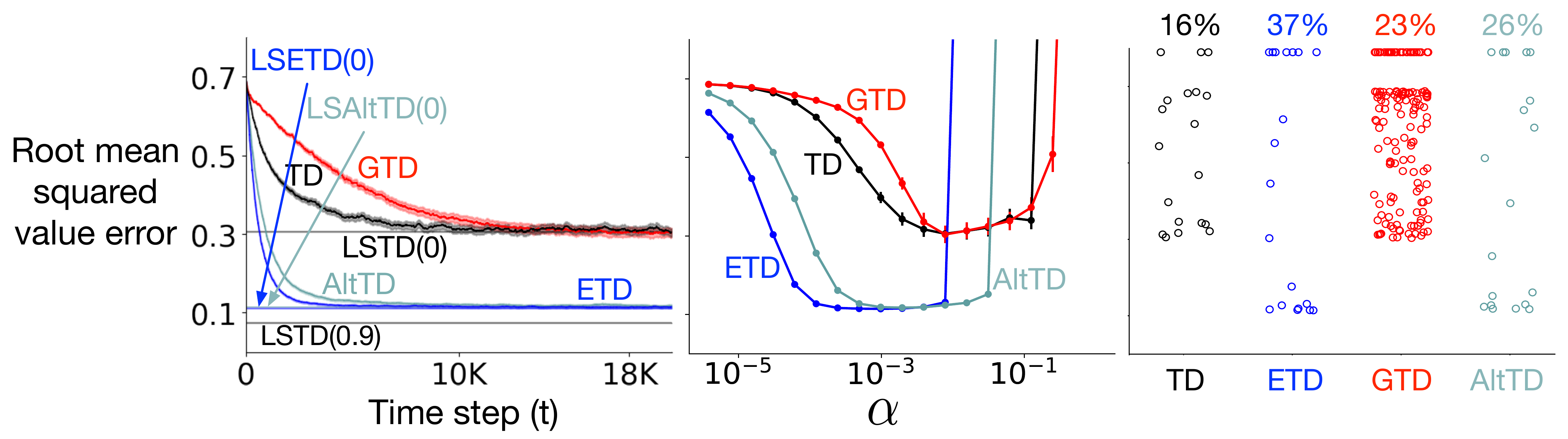}
      \caption{Comparing value error of GTD(0), ETD(0), Off-policy TD(0), and Alternative-life TD(0) on the {\bf Collision} problem. In both the learning curve (left) and stepsize plot (middle) the free parameters are optimized to minimize final performance---the ${\text{R}}\overline{\text{VE}}$ averaged over the last 200 time steps. The parameter sensitivity plot (right) shows the final performance for each parameter configuration tested. ETD(0) exhibits a clear advantage in terms of speed of learning and final performance compared with the other three methods, across all three plots. GTD(0) and Off-policy TD(0) performed similarly but Off-policy TD(0) achieved lower value error with less samples in this domain, when optimizing for final performance.}
      \label{fig:threemain_chicken_final_0}
\end{figure}
\begin{figure}[h!]
      \centering
      \includegraphics[width=0.8\linewidth]{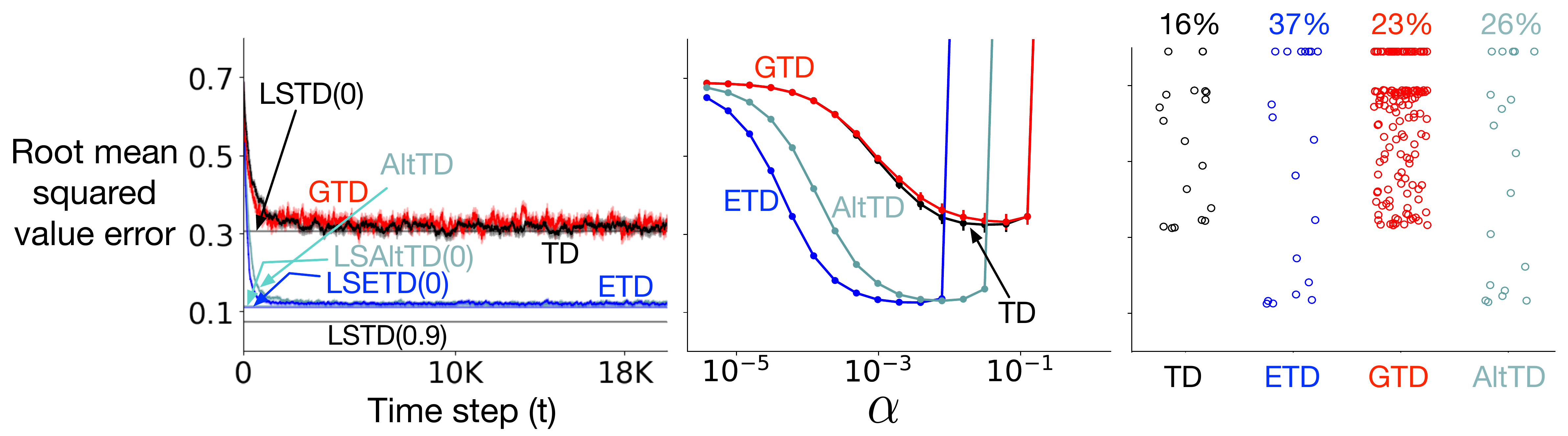}
      \caption{Comparing the learning speed of GTD(0), ETD(0), Off-policy TD(0), and Alternative-life TD(0) on the {\bf Collision} problem. In both the learning curve (left) and stepsize plot (middle) the free parameters are optimized to AUC---the ${\text{R}}\overline{\text{VE}}$ averaged over all 20,000 time steps. Even though the performance of GTD(0) was optimized over many more parameter settings,  if performs almost identically to Off-policy TD(0) when performance is optimized for AUC. ETD(0) again exhibits a clear advantage in terms of speed and final performance. The range of stepsizes for which ETD(0) performs well is slightly smaller than that of Off-policy TD(0) and GTD(0) as shown in the stepsize plot.}
      \label{fig:threemain_chicken_auc_0}
\end{figure}
\begin{figure}[h!]
      \centering
      \includegraphics[width=0.8\linewidth]{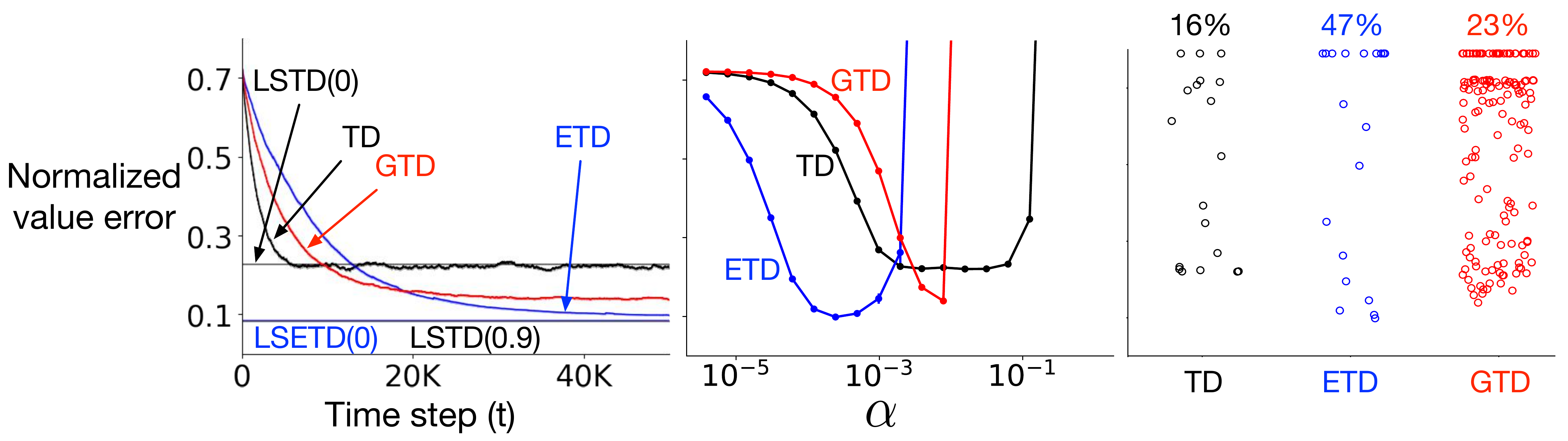}
      \caption{Comparing value error of GTD(0), ETD(0), and Off-policy TD(0) on the {\bf Four Rooms} problem. In both the learning curve (left) and stepsize plot (middle) the free parameters are optimized to minimize final performance---the ${\text{R}}\overline{\text{VE}}$ averaged over the last 500 time steps. ETD(0) exhibits better final performance on this problem, but required more samples than the other algorithms to achieve low ${\text{TR}}\overline{\text{VE}}$. Off-policy TD(0) exhibited the quick early learning performance, but achieved the highest final error. GTD(0) learned more quickly than ETD(0), and larger final error. The stepsize plot (middle) highlights the sensitivity of the GTD(0) algorithm's performance as a function of $\alpha$. One value of $\alpha$ resulted the best performance for GTD(0); however, many combinations of $\alpha$ and $\alpha_h$ perform nearly as well as shown in the parameter sensitivity plot on the right.}
      \label{fig:threemain_four_final_0}
\end{figure}
\begin{figure}[h!]
      \centering
      \includegraphics[width=0.8\linewidth]{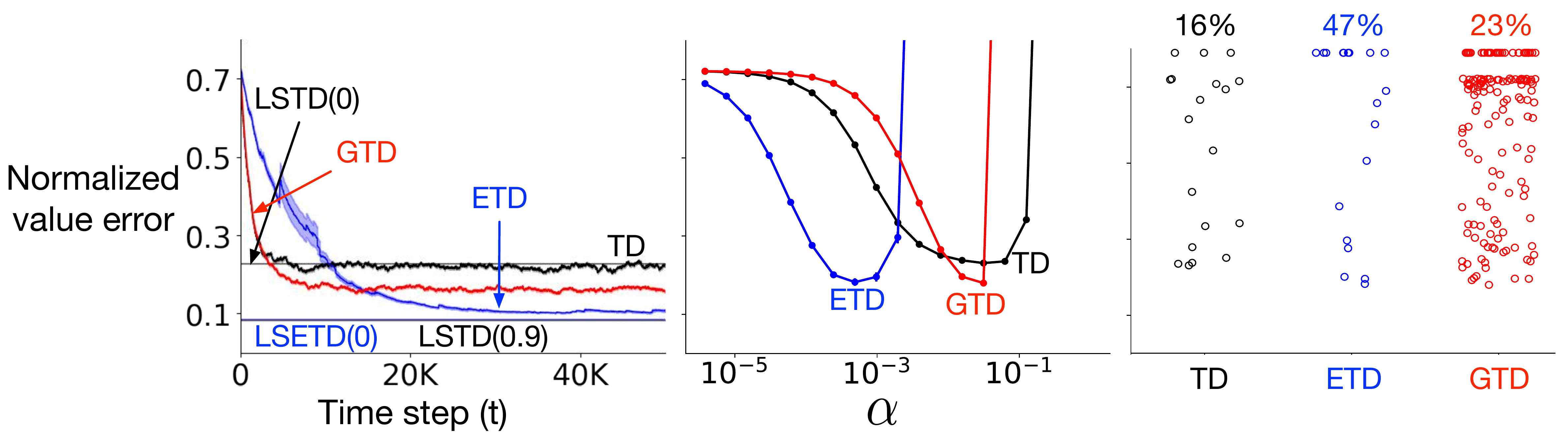}
      \caption{Comparing value error of GTD(0), ETD(0), andy Off-policy TD(0) on the {\bf Four Rooms} problem. In both the learning curve (left) and stepsize plot (middle) the free parameters are optimized to minimize area under the learning curve---the ${\text{R}}\overline{\text{VE}}$ averaged over the last 500 time steps. In both the learning curve (left) and stepsize plot (middle) the free parameters are optimized to AUC---the ${\text{R}}\overline{\text{VE}}$ averaged over all 50,000 time steps. When $\alpha$ is chosen according to the AUC criteria, GTD(0) strikes a good balance between learning speed and final performance. As before, ETD(0) achieves the best final performance, but also learns more slowly than the other algorithms.}
      \label{fig:threemain_four_auc_0}
\end{figure}

The results are grouped by problem and the value of the trace parameter. We begin with the simplest case, where neither GTD($\lambda$) nor ETD($\lambda$) make use of eligibility traces ($\lambda = 0$) in Figures \ref{fig:threemain_chicken_final_0}, \ref{fig:threemain_chicken_auc_0}, \ref{fig:threemain_four_final_0}, and \ref{fig:threemain_four_auc_0}. 
We include Off-policy TD($\lambda$) and Alternative-life TD($\lambda$) as a baseline even though neither are guaranteed to converge in either problem. Alternative-life TD($\lambda$) is only evaluated on the Collision problem as it is not yet clear how it can be extended for solving a continuing task.

%

For $\lambda = 0$, ETD(0) exhibits a clear advantage over the other algorithms in terms of final performance, as shown in Figures \ref{fig:threemain_chicken_final_0} through \ref{fig:threemain_four_auc_0}. Whether optimizing for final performance or AUC, ETD(0) reaches a lower value error. In the Collision problem, for example, the ${\text{R}}\overline{\text{VE}}$ of ETD(0) is almost 0.1, whereas GTD(0) and Off-policy TD(0) achieve a ${\text{R}}\overline{\text{VE}}$ of approximately 0.3. In the Four Rooms problem, the gain is less pronounced, but ETD(0) still statistically significantly outperforming GTD(0) and more clearly outperforming Off-policy TD(0).
This result is particularly interesting, as the ${\text{R}}\overline{\text{VE}}$ is weighted by the behavior policy, but ETD(0) optimizes the MSPBE objective weighted by the emphatic weighting $f$ (see Equation \ref{eq:ETDF}). GTD(0), on the other hand, optimizes the MSPBE weighted by the behavior policy---the same weighting used in the computation of ${\text{R}}\overline{\text{VE}}$---and yet GTD(0) does not achieve lower ${\text{R}}\overline{\text{VE}}$. This result suggests that the emphatic weighting can actually improve performance in terms of the value error. ETD(0) learns more slowly than GTD(0) and TD on the Four Rooms problem, even when $\alpha$ is chosen to optimize AUC (as shown in the learning curve of Figure \ref{fig:threemain_four_auc_0}). In this experiment, ETD(0) performed best with a small stepsize, which results in small updates on steps where the emphatic weighting is not large.

The behavior of each algorithm is also quite different in terms of parameter sensitivity. The best performing stepsize values for ETD(0) are considerably smaller than the best performing stepsize parameter values for GTD(0) and TD(0) in both problems. This is likely due to the fact that the ETD(0) update is the same as the Off-policy TD(0) update multiplied by an emphatic weighting---which can be large. In fact, this suggests the original ETD(0) algorithm could have been designed to account for this magnitude difference, with a normalization factor, and does not necessarily indicate any deficiencies on the part of the algorithm. Otherwise, the sensitivity of both TD methods and ETD(0) to choice of $\alpha$ is quite comparable---both exhibit broad U-shapes indicating that tuning $\alpha$ should not be too much of a challenge. GTD(0) appears to be more sensitive to the choice of $\alpha$ in the Four Rooms problem. This is most clearly illustrated in the parameter sensitivity plot, where many parameter combinations for GTD(0) are clustered points near the top, likely due to performance induced by large values of $\alpha_h$.

\begin{figure}[h!]
      \centering
      \includegraphics[width=0.8\linewidth]{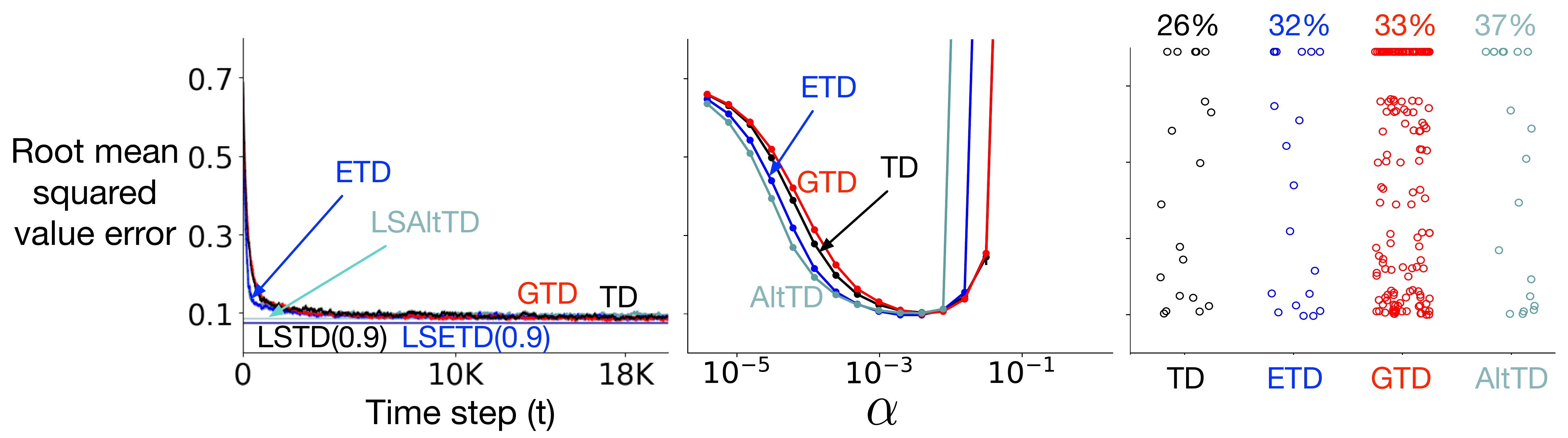}
      \caption{Comparing learning speed of GTD(0.9), ETD(0.9), and Off-policy TD(0.9) on the {\bf Collision} problem, value error over time (optimized for AUC). All methods tend to perform similarly as the value of $\lambda$ increases to 0.9. All methods are faster and have a better final performance when $\lambda=0.9$, compared to when $\lambda=0.0$. Interestingly, the performance of ETD($\lambda$) does not change much when $\lambda$ is increased to 0.9. ETD(0.9) seems to be robust to the value of the trace parameter.}
      \label{fig:threemain_chicken_auc_9}
\end{figure}

\begin{figure}[h!]
      \centering
      \includegraphics[width=0.8\linewidth]{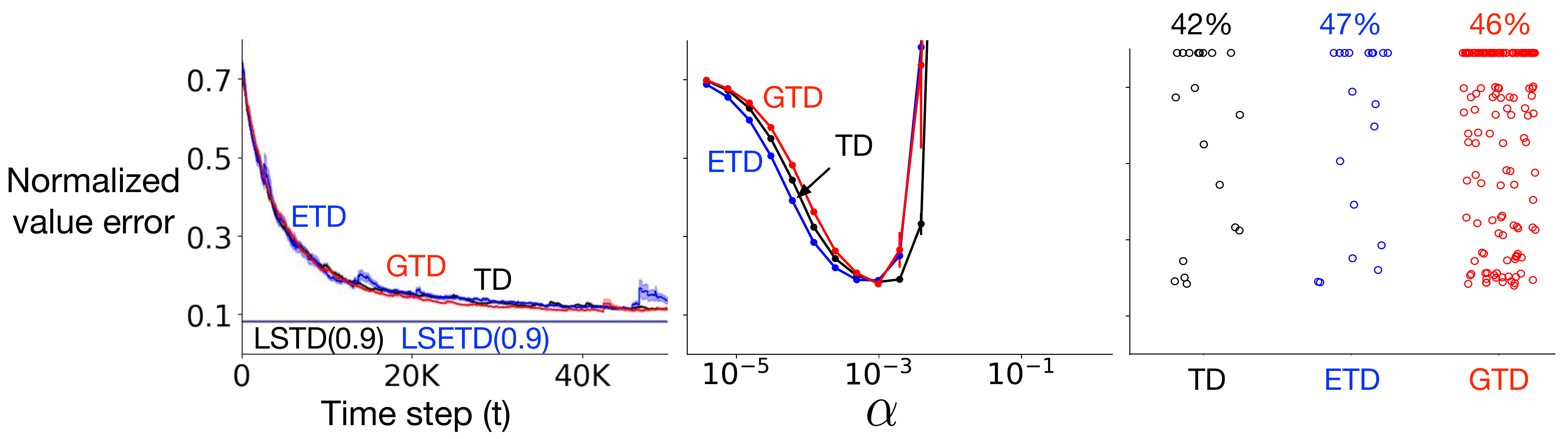}
      \caption{Comparing learning speed of GTD(0.9), ETD(0.9), and Off-policy TD(0.9) on the {\bf Four Rooms} problem, value error over time (optimized for AUC). All methods tend to perform similarly as the value of $\lambda$ increases to 0.9. All methods are faster and have a better final performance when $\lambda=0.9$, compared to when $\lambda=0.0$. Interestingly, the performance of ETD($\lambda$) does not change much when $\lambda$ is increased to 0.9. ETD(0.9) seems to be robust to the value of the trace parameter.}
      \label{fig:threemain_four_auc_9}
\end{figure}

Finally, all four algorithms perform similarly when $\lambda = 0.9$, as shown in Figures \ref{fig:threemain_chicken_auc_9} and \ref{fig:threemain_four_auc_9}. Eligibility traces allow all the algorithms to converge to a lower value error.  ETD($\lambda$) achieves only slightly lower value error compared to ETD(0). The learning speed and final error of both variants of TD($\lambda$) and GTD($\lambda$) improve dramatically with the inclusion of eligibility traces. This suggests that the final performance of ETD($\lambda$) and its speed is not particularly impacted by the choice of $\lambda$, and potentially the emphatic weighting helps to overcome the bias introduced by setting $\lambda = 0$. The results for optimizing the parameters for final performance when $\lambda$ equals 0.9 can be found in the appendix.

The reason ETD($\lambda$) is robust to $\lambda$ is that when $\lambda$ is small, the emphatic weighting plays a larger role, and when lambda is big, the emphatic weighting gets close to 1. In fact, looking at the emphatic weighting, one can see that ETD($\lambda$) has some similarity to TD(1). We show in Appendix \ref{app:EmphaticTDComparedWithTD1} why ETD($\lambda$) has some similarity to TD(1), but that importantly it does still behave differently for different $\lambda$ and better than TD(1).

\section{Accelerating Gradient-TD updates}
\label{sct:ComparingGradientTDMethodsdWithProximalGradientTDMethods}
The GTD($\lambda$) and GTD2($\lambda$) algorithms make use of a secondary set of learned weights; with implications in both theory and practice. The main consequence is that GTD($\lambda$) is not a true gradient descent method---in the sense that the derivation makes a clear approximation to the gradient of the MSPBE---and that $\mathbb{E}_{d_b}[\vecx_t \vecx_t^T]^{-1}\mathbb{E}_{d_b}[\delta_t \vecx_t]$ is nearly stationary and can be estimated with an additional weight vector that is learned more quickly than the primary weights. Characterizing the asymptotic performance of the resultant algorithm requires a two-timescale analysis, requires characterizing other properties of the algorithm---for instance, the finite sample performance---and requires non-standard and potentially novel analysis tools. In addition, GTD($\lambda$) cannot easily make use of recently developed gradient descent accelerations, which may result in improvements in sample efficiency.

The Proximal Gradient-TD methods do not approximate the gradient of the MSPBE, instead formulating the task of minimizing the MSPBE as a saddlepoint problem. Our primary concern here is the practical implications of this new formulation. The proximal methods are identical to GTD($\lambda$) and GTD2($\lambda$), so the key question is how much improvement does the gradient acceleration yield in practice? We compared GTD($\lambda$), GTD2($\lambda$), Proximal GTD($\lambda$), and Proximal GTD2($\lambda$). Prior work has shown these accelerations improve learning in Biard's counter example ---an extreme case where the target and behavior policy are very different---and randomly generated MDPs (Mahadevan et al., 2014; White \& White, 2016). Our experiments attempt to investigate the performance advantage of Proximal-GTD methods in our two simulation problems that were designed to reflect less extreme and more structured problems.\footnote{Although randomly generated tasks are useful to avoid bias in problem design, randomization does not often accurately reflect real-world settings well and should be complemented to problems inspired by real-world problems. See Gent et al. (1997) for a great discussion of the issues with random problems, and other related issues.}

\begin{figure}[h!]
      \centering
      \includegraphics[width=0.8\linewidth]{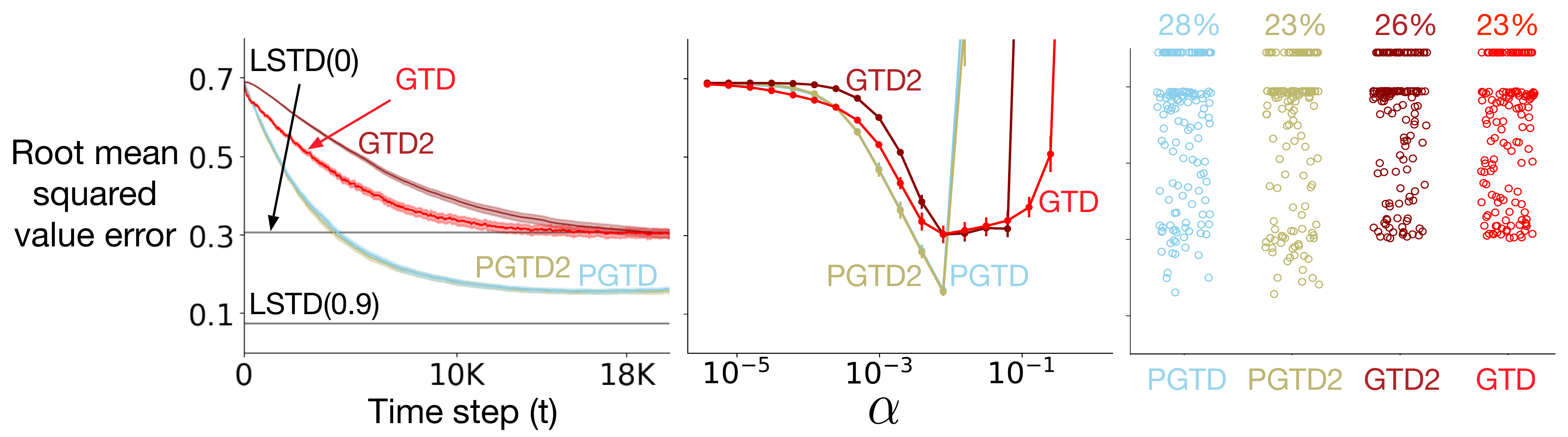}
      \caption{Comparing final performance of GTD(0), GTD2(0), Proximal-GTD(0), and Proximal-GTD2(0) on the {\bf Collision} problem. Parameters are optimized for final performance. The Proximal-GTD methods outperformed their non-proximal counterparts in both speed of learning and final performance as shown on the left. However, the parameter sensitivity plot (right) shows that it is hard to find the combination of first and second stepsizes for which the Proximal Gradient-TD methods get their best performance---few parameter combinations clustered near the bottom. The stepsize plot (middle) paints a similar picture: both Proximal Gradient-TD methods achieve low error for a specific setting of $\alpha$, whereas the Gradient-TD methods are less sensitive to choice of $\alpha$ at the cost of higher error.}
      \label{fig:proximal_chicken_final_0}
\end{figure}

As before, we investigated the performance of each algorithm in our two benchmark problems using two performance measures. The following discussion focuses on the case where the differences between the algorithms were most significant: when $\lambda = 0$, and the tuneable parameters optimized for final performance, in Figures \ref{fig:proximal_chicken_final_0} and \ref{fig:proximal_four_final_0}. The results of the other experiments can be found in Appendix \ref{app:ComparingGradientTDMethodsdWithProximalGradientTDMethods}. 

\begin{figure}[h!]
      \centering
      \includegraphics[width=0.8\linewidth]{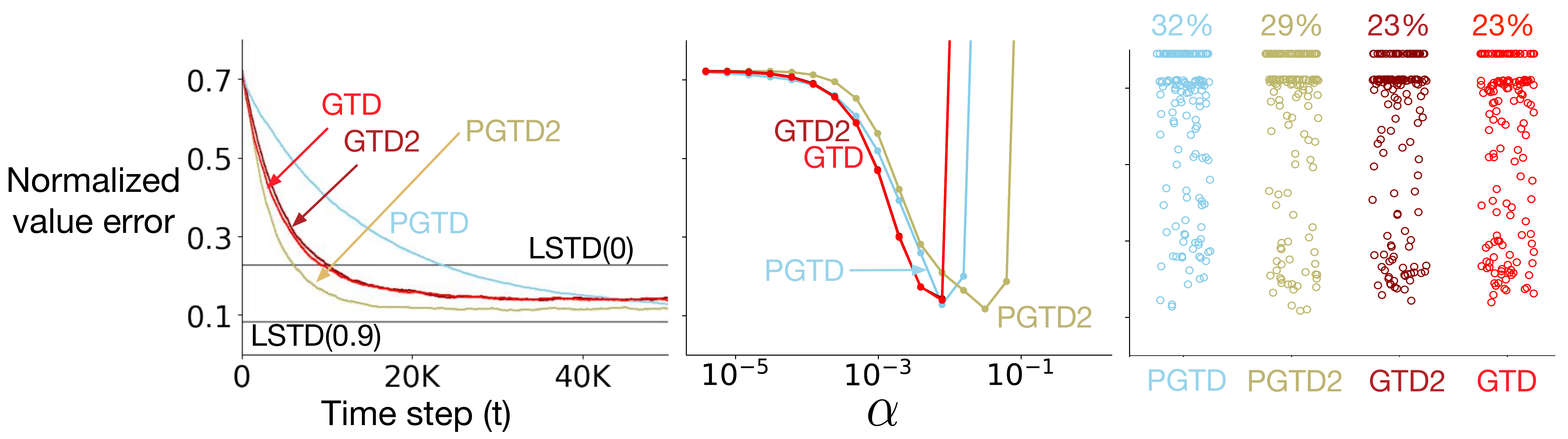}
      \caption{Comparing final performance of GTD(0), GTD2(0), Proximal-GTD(0) and Proximal-GTD2(0) on the {\bf Four Rooms} problem. Parameters are optimized for final performance. All methods performed similarly on this task, except Proximal-GTD(0) was significantly slower, requiring more steps to reach the error achieved by the other methods. Proximal-GTD2(0) on the other hand, was the fastest. The parameter sensitivity plot suggest GTD(0)'s performance is less sensitive to the choice of parameters compared with the other methods, as highlighted by outliers near the bottom of the plot for Proximal-GTD(0), Proximal-GTD2(0), and GTD2(0).}
      \label{fig:proximal_four_final_0}
\end{figure}

The main conclusions are that Proximal Gradient-TD methods achieve significantly lower error in the first problem, but their performance is more sensitive to parameter settings, in particular $\alpha$.
In the Collision problem, Proximal-GTD(0) and Proximal-GTD2(0) perform identically, achieving a final error a bit higher than 0.1, compared to an error of about 0.3 for GTD(0) and GTD2(0). However, looking at the corresponding parameter sensitivity, this lower error is only possible for a particular setting of the $\alpha$. Outside this particular setting, the sensitivity curve is worse for Proximal-GTD(0) and Proximal-GTD2(0), whereas GTD(0) and GTD2(0)'s performance exhibits a wider range of well-performing stepsizes. The parameter sensitivity plot indicates that Proximal-GTD(0) and Proximal-GTD2(0) only manage to get this lower error for a few settings of their key parameters.

Our results might appear in contradiction to prior comparisons of Proximal-GTD methods reported in literature. Originally Proximal-GTD2(0) and Proximal-GTD(0) were specified as using one shared stepsize parameter for updating $\vecw$ and $\vech$. The first empirical comparisons of Proximal-GTD and Gradient-TD methods were conducted on Baird's counter example, where all methods were forced to use a single $\alpha$ value. The experiment showed that Proximal-GTD(0) and Proximal-GTD2(0) reduced the MSPBE over 1000 steps much more quickly than GTD(0) (Mahadevan et al., 2014). Later studies reaffirmed the advantages of proximal methods on Baird's counter-example, when $\vecw$ and $\vech$ where allowed to use different stepsize parameters. However, after 1000 steps the performance of the algorithms converged to the same error level (White \& White, 2016). Results in on-policy Random MDPs (Mahadevan et al., 2014) and an energy domain (Liu et al., 2016) contradict later studies on Random MDPs (White \& White, 2016) that show that conventional Gradient-TD methods perform best in both on and off-policy settings. One possible explanation is that some studies were performed restricting both weight vectors to share a single stepsize parameter. Mahadevan et al. (2014) use the same approach of Dann et al. (2013) to optimize the parameters of the methods. Dann et al. used three independent runs to determine the best performing values of each parameter setting, and thus there is a good chance that the best performing parameters where not actually identified in both works. Finally, our results in the Collision and Four Rooms problems and those of White and White (2016), suggest that it is difficult to tune the parameters of Proximal Tradient-TD methods, whereas prior work only reported learning curves with the best parameters (Mahadevan et al., 2014; Liu et al, 2016). Including stepsize and parameter sensitivity plots helps paint a much fuller picture of each algorithms relative merits and limitations. Nevertheless, our results clearly show the Proximal Gradient-TD methods can learn faster and reach lower error than Gradient-TD methods.

The final performance of the methods with two sets of weights (e.g., GTD(0), HTD(0), and the Mirror-prox methods), is similar to the error achieved by Off-policy TD methods with $\lambda >0$, suggesting that tuning the extra learning rate parameter allows these methods to perform better than the LSTD(0) baseline would suggest possible.  For example, in Figure \ref{fig:threemain_four_auc_0} GTD(0) achieved a final performance that was better than LSTD(0). According to Figure \ref{fig:proximal_four_final_0}, Proximal-GTD(0) performed better than LSTD(0). For connivence, we include the performance of LSTD(0.9) as an additional baseline for any experiment involving methods with two sets of weights.

\section{Improving Gradient-TD with hybrid updates}
\label{sct:ComparingHybridTDMethodsWithTD}

In this section, we attempt to provide new data for the Hybrid TD hypothesis: performing on-policy TD updates whenever the data is generated on-policy will yield better performance compared with Gradient-TD methods. Results from Sutton et al.,\ (2009) suggest that there are situations where conventional TD(0) can outperform GTD(0). This, and a few other observations (e.g., Hackman, 2012), motivated the development of Hybrid TD methods. To our knowledge, there are no empirical results where a Hybrid method exhibits a significant advantage over Gradient-TD.

\begin{figure}[h!]
      \centering
      \includegraphics[width=0.8\linewidth]{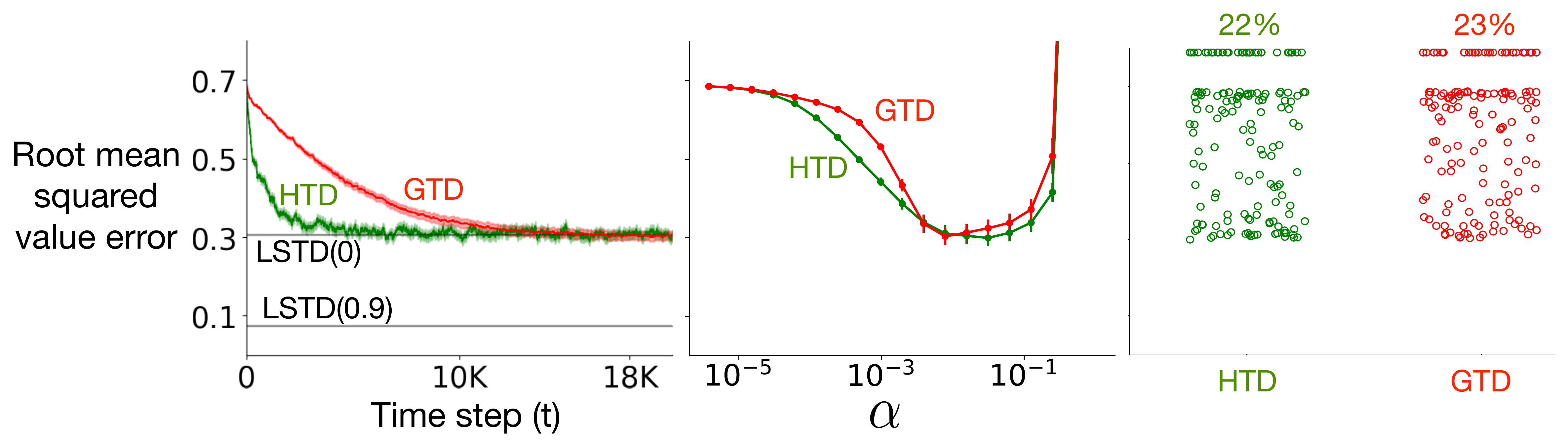}
      \caption{Comparing final performance of GTD(0) and HTD(0) on the {\bf Collision} problem. Free parameters are optimized for final performance. HTD(0) requires fewer samples to achieve the same error as compared with GTD(0). Both methods exhibit similar sensitivity to choice of $\alpha$, with HTD(0) performing slightly better in this regard. The parameter sensitivity plot shows that the sensitivity to parameters is also similar between the two.}
      \label{fig:htd_chicken_final_0}
\end{figure}

\begin{figure}[h!]
      \centering
      \includegraphics[width=0.8\linewidth]{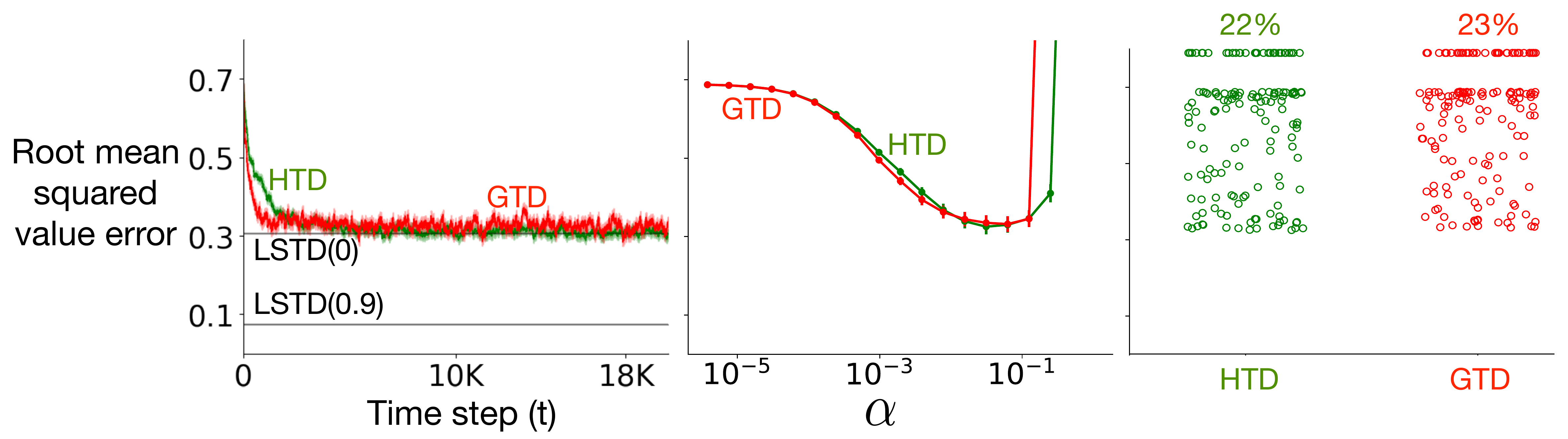}
      \caption{Speed of learning of  GTD(0) and HTD(0) on the {\bf Collision} problem. When $\alpha$ is chosen to minimize AUC the methods perform very similarily. This result combined with figure \ref{fig:htd_chicken_final_0} illustrate the importance of reporting multiple performance measures and optimizing the tunable parameters in different ways to get a more complete assessment of performance.}
      \label{fig:htd_chicken_auc_0}
\end{figure}

\begin{figure}[h!]
      \centering
      \includegraphics[width=0.8\linewidth]{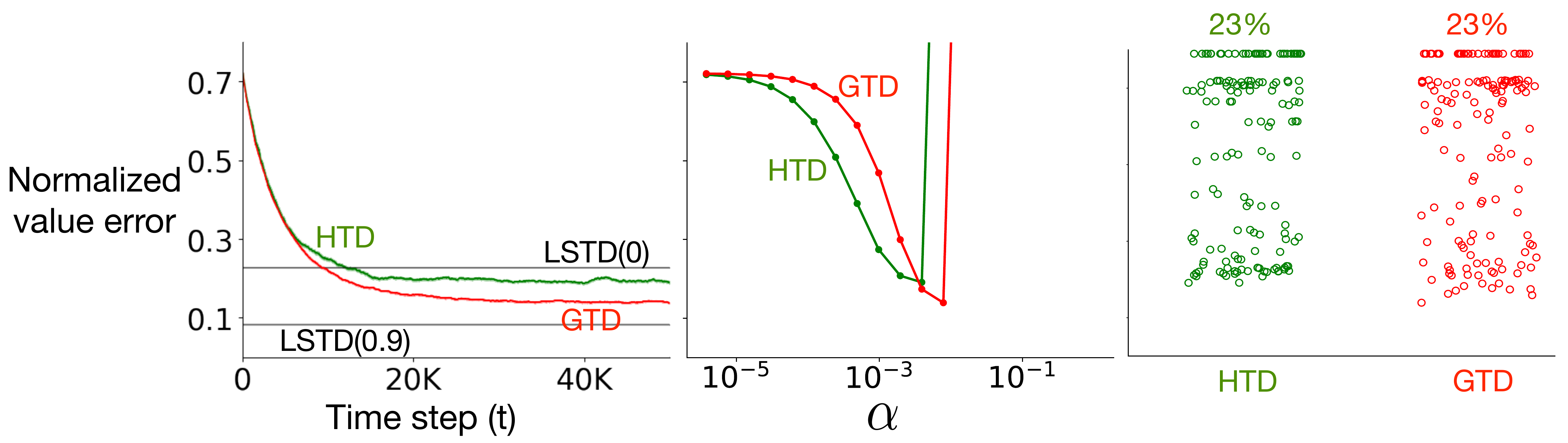}
      \caption{Final performance comparison of GTD(0) and HTD(0) on the {\bf Four Rooms} problem. Free parameters are optimized for final performance. In this problem GTD(0) achieved better final error, and exhibited slightly more sensitivity to the choice of $\alpha$.}
      \label{fig:htd_four_final_0}
\end{figure}

\begin{figure}[h!]
      \centering
      \includegraphics[width=0.65\linewidth]{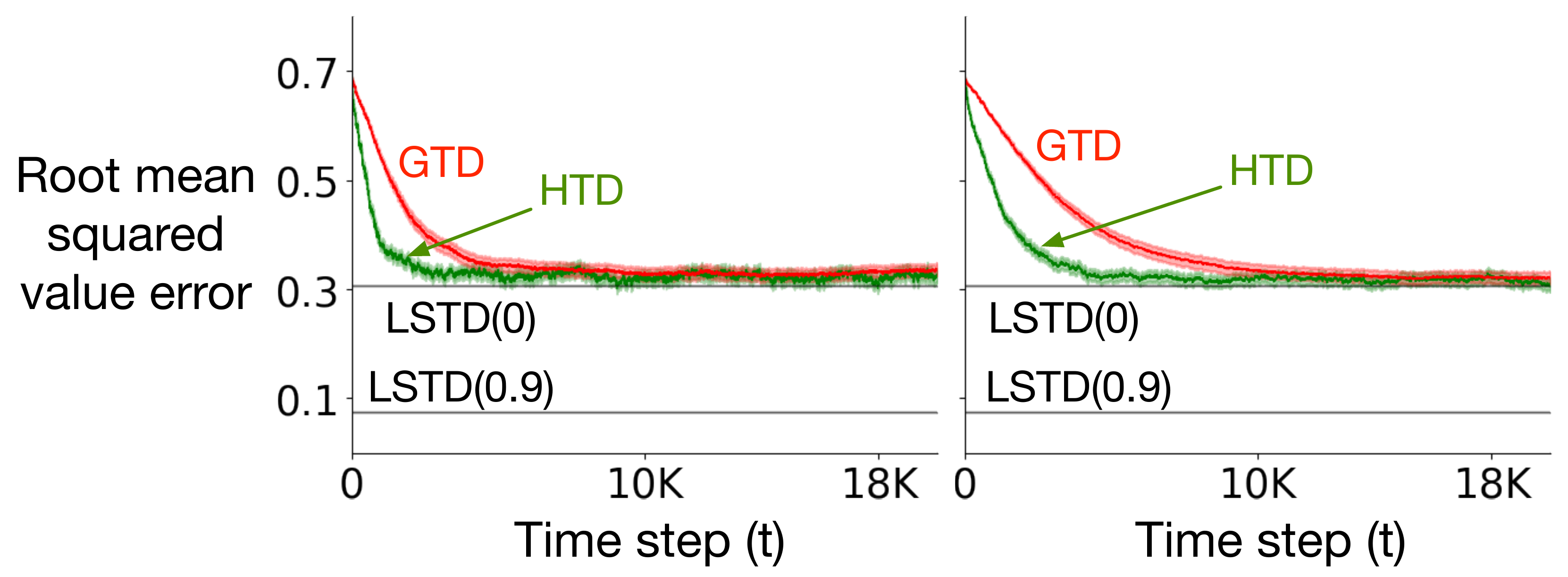}
      \caption{Learning curves comparing GTD(0) and HTD(0) on the \textbf{Collision} problem. The learning curve on the left is plotted using the $\alpha$ values (primary stepsize) that achieved the best AUC for each algorithm, whereas the right plot shows the performance with $\alpha$ values (primary stepsize) that achieved the best final performance. The value of the second step-size parameter was set to $\alpha_h = 2.56 \times \alpha$, to illustrate the performance of the algorithms when we respect the theoretical guidelines for convergence: $\alpha_h > \alpha$. In this setting, HTD(0) requires less samples to achieve the same error as the LSTD(0) baseline, and is significantly faster than GTD(0).}
      \label{fig:htd_chicken_eta}
\end{figure}

As before, we sub-select the most notable results, and relegate the remaining results to Appendix \ref{app:ComparingHybridTDMethodsWithTD}. Figures \ref{fig:htd_chicken_final_0}, \ref{fig:htd_chicken_auc_0}, and \ref{fig:htd_four_final_0} summarize the main results. Overall we conclude that there is not a significant difference between GTD(0) and HTD(0) in either tasks when we optimize all the tunable parameters, agreeing with previous results (White \& White, 2016).

Hybrid methods do exhibit one clear advantage that requires a bit more analysis to uncover. In many problems, either on-policy or problems where the behavior and target policies are very similar, GTD(0) achieves the best performance when $\alpha$ is much larger than $\alpha_h$. This means the gradient correction term has little impact on the primary weight update. In Baird's counter-example---the behavior and target policy are very different---we observe the opposite phenomenon: GTD(0) achieves the best performance when $\alpha_h \gg \alpha$. This is not ideal from a practical point of view. HTD(0) should not have this limitation. When the problem is close to on-policy, the value of $\alpha_h$ does not matter. The results above support this hypothesis. HTD(0) outperforms GTD(0) in the Collision problem because the agent has frequent updates from on-policy transitions. Figure \ref{fig:htd_chicken_eta} plots the value error of HTD(0) and GTD(0) against when $\eta$ is equal to 2.56 (where $\alpha_h = \eta\alpha$) for the Collision problem; here we see HTD(0) has a clear advantage.

\section{Variance reduction for Emphatic-TD} 
\label{sct:ComparingETDAndETDBeta}
Although the analysis was the primary motivation behind the Generalized Emphatic TD($\lambda$) method, here we are interested in its practical benefits. In principle, the $\beta$ parameter of ETD($\lambda$, $\beta$) should enable the algorithm to trade-off bias and variance, and regardless ETD($\lambda$, $\beta$) should have an advantage over ETD($\lambda$) due to the increased flexibility of the extra parameter.

\begin{figure}[h!]
      \centering
      \includegraphics[width=0.8\linewidth]{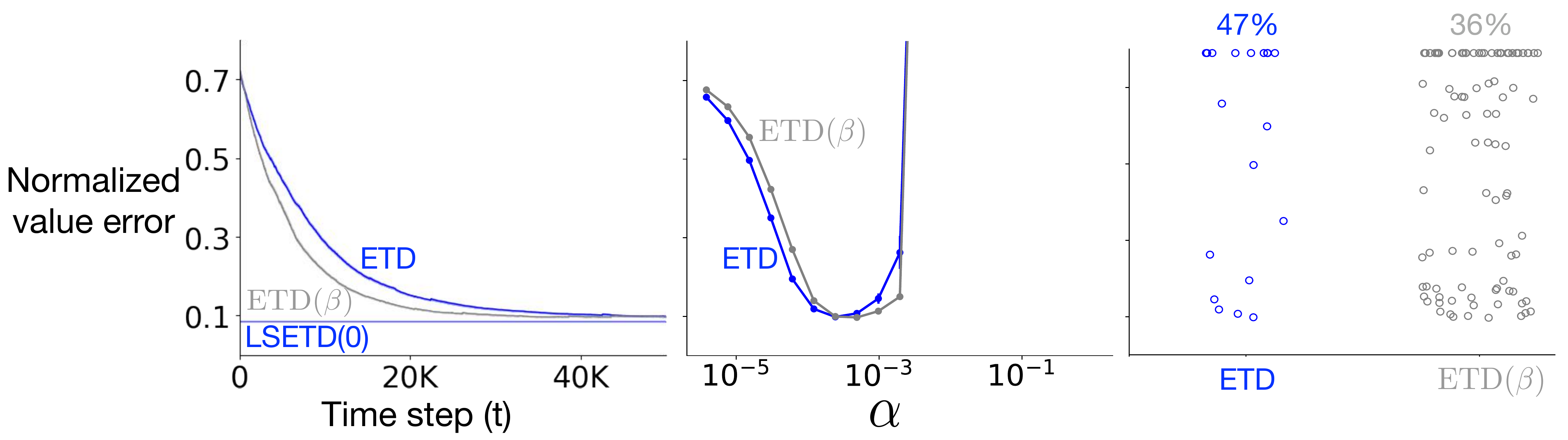}
      \caption{Normalized value error of ETD(0) and ETD(0, $\beta$) on the {\bf Four Rooms} problem. In both the learning curve (left) and stepsize plot (middle) the free parameters are optimized to minimize final performance. Both methods performed similarly on this problem.}
      \label{fig:beta_four_final_0}
\end{figure}

\begin{figure}[h!]
      \centering
      \includegraphics[width=0.8\linewidth]{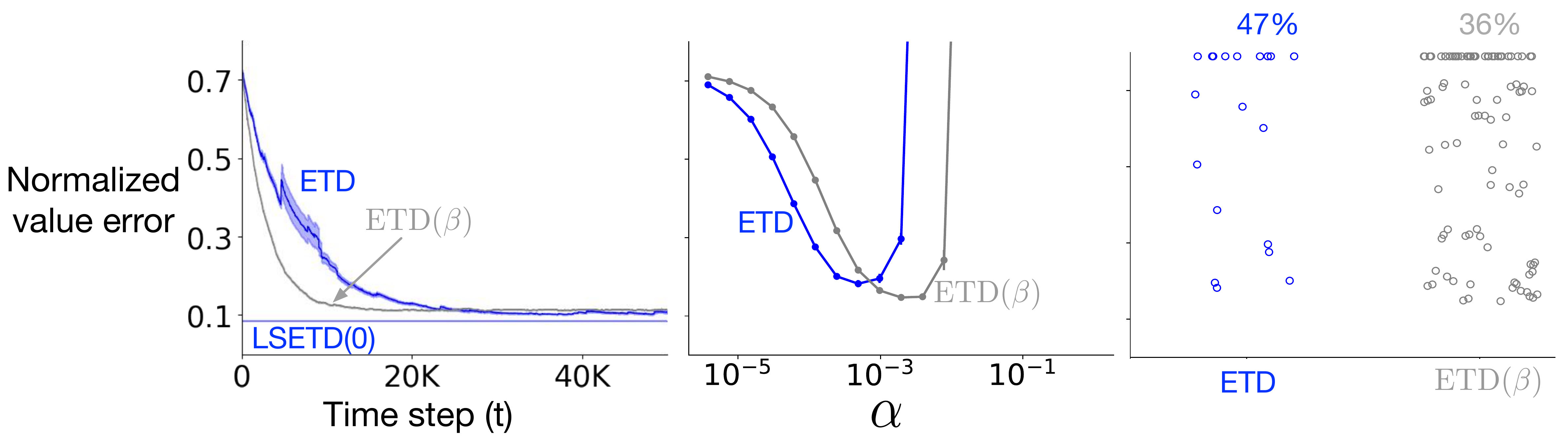}
      \caption{Comparing speed of learning of ETD(0) and ETD(0, $\beta$) on the {\bf Four Rooms} problem. In both the learning curve (left) and stepsize plot (middle) the free parameters are optimized to minimize AUC. Under the AUC criteria, ETD(0, $\beta$) requires less samples than ETD(0) to reach the same error level. Neither method dominates in terms of final error. Although, ETD(0) appears to exhibit more variance, this does not seem to be a common trend if we inspect the results in both problems and both values of $\lambda$. (Figure \ref{fig:beta_four_final_0} and Appendix \ref{app:ComparingETDAndETDBeta}).}
      \label{fig:beta_four_auc_0}
\end{figure}

The clearest differences appear in the Four Rooms problem with $\lambda=0$. The remaining results can be found in Appendix \ref{app:ComparingETDAndETDBeta}. We include the results comparing ETD(0) and ETD(0, $\beta$) with the parameters optimized for AUC in Figure \ref{fig:beta_four_auc_0}, and the results comparing ETD(0) and ETD(0, $\beta$) with the parameters optimized for final performance can be found in Figure \ref{fig:beta_four_final_0}. Both algorithms perform nearly identically in the Collision problem, for both $\lambda$ equal to 0 and 0.9. See Appendix~\ref{app:ComparingETDAndETDBeta} for details.


Overall, ETD($\lambda,\beta)$ and ETD($\lambda$) perform similarly in these problems. ETD($\lambda,\beta)$ appears to learn faster, but both achieve similar final performance in all settings. There are minor differences in the Collision problem when $\lambda = 0$, with ETD(0) exhibiting slightly more sensitivity than ETD(0, $\beta$). However, ETD(0, $\beta$) achieved best performance with $\beta$ equal to 0.8 in the Collision problem. This value is close the value of $\gamma_t$ in this problem (a constant equal to 0.9 in this task), and thus does not introduce bias but should reduce variance compared to ETD(0). It is worth emphasizing that ETD($\lambda,\beta)$ has an additional tunable parameter compared with ETD($\lambda$), and thus optimizing for $\beta$ independently for performance is expected to yield performance improvements. Further experiments are needed to refine these differences, especially in the case where $\beta$ and $\gamma$ change with time.

\section{Reducing variance due to importance sampling}
\label{sct:ComparingTDWithABTDV-traceAndTreeBackup}


All the experiments we have presented focus on methods that use importance sampling, but this approach may be problematic if large importance sampling ratios are common. The Tree Backup($\lambda$), V-trace($\lambda$), and ABTD($\lambda$) algorithms all eliminate the explicit use of importance sampling ratios by varying the amount of bootstrapping in TD updates. In situations where the target and behavior policies differ greatly, we expect these methods to reduce the length of the eligibility traces and exhibit more stable learning when compared with importance sampling based methods like GTD($\lambda$). In the Collision and Four Rooms problems, the two policies are too similar to observe a clear advantage for these action-dependent boostrapping methods.

\begin{figure}[]
      \centering
      \includegraphics[width=0.5\linewidth]{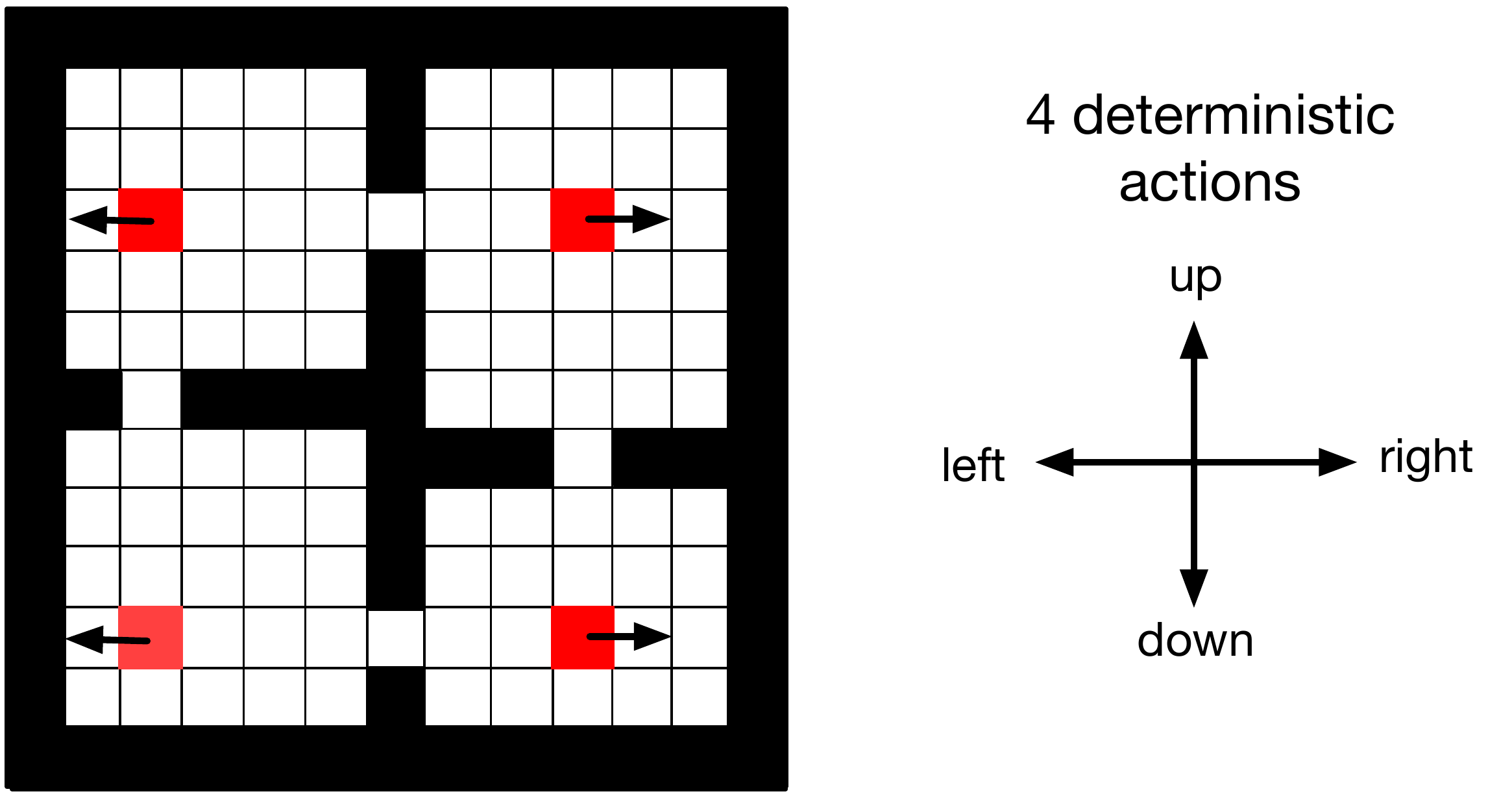}
      \caption{The {\bf High Variance Four Rooms} problem is the same as the Four Rooms problem except for four of the states (shown in red), in which the agent takes the left or the right action (shown as a black arrow) with high probability and the rest of the actions with low probability.}
      \label{fig:HighVarianceFourRoomsFigure}
\end{figure}

To study the potential benefits of these methods, we modified the behavior policy in the Four Rooms problem so that high magnitude importance sampling corrections are encountered occasionally. We selected four states---marked in Figure \ref{fig:HighVarianceFourRoomsFigure}---and modified the behavior policy as follows: the left or right action (as indicated in the figure) are selected with 0.97 probability, while the remaining actions in each of the four states are taken with 0.01 probability. This simple change generates importance sampling ratios as large as 50.0; perhaps not common in practice, but useful for analysis of the algorithms.
The results summarizing the performance of the action-dependent bootstrapping methods in this high variance Four Rooms problem for $\lambda=0$ and $\lambda=0.9$ can be found in Figures \ref{fig:trace_four_final_0} and \ref{fig:trace_four_final_9}, with additional results included in Appendix \ref{app:ComparingTDWithABTDV-traceAndTreeBackup}.

\begin{figure}[h!]
      \centering
      \includegraphics[width=0.8\linewidth]{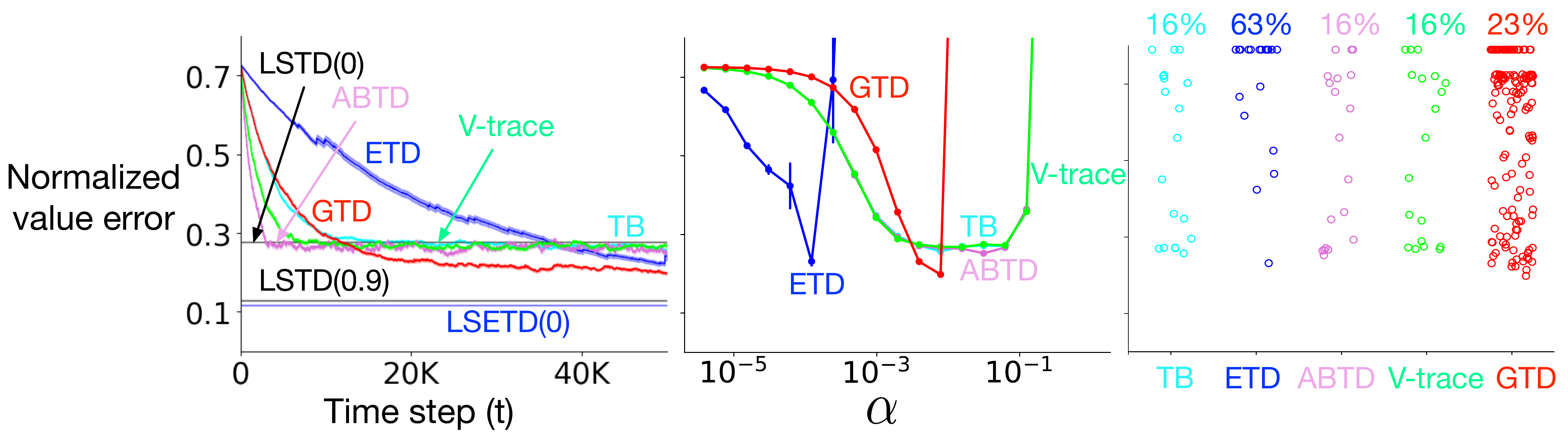}
      \caption{Comparing GTD(0), ETD(0), ABTD(0), Tree Backup(0), and V-trace(0) on the {\bf High Variance Four Rooms} problem. Plots are optimized for final performance. Middle and right graphs show the final performance. GTD(0) and ETD(0) are significantly slower than other methods on this problem; however, they both manage to converge to the best final performance among all methods. It is hard to find ETD(0)'s stepsize for which it converges to its best final performance (see the middle graph and the parameter sensitivity graph on the right). V-trace(0), Tree Backup(0) and ABTD(0), all perform similarly. ABTD(0) is a bit faster than the other two methods.}
      \label{fig:trace_four_final_0}
\end{figure}

\begin{figure}[h!]
      \centering
      \includegraphics[width=0.8\linewidth]{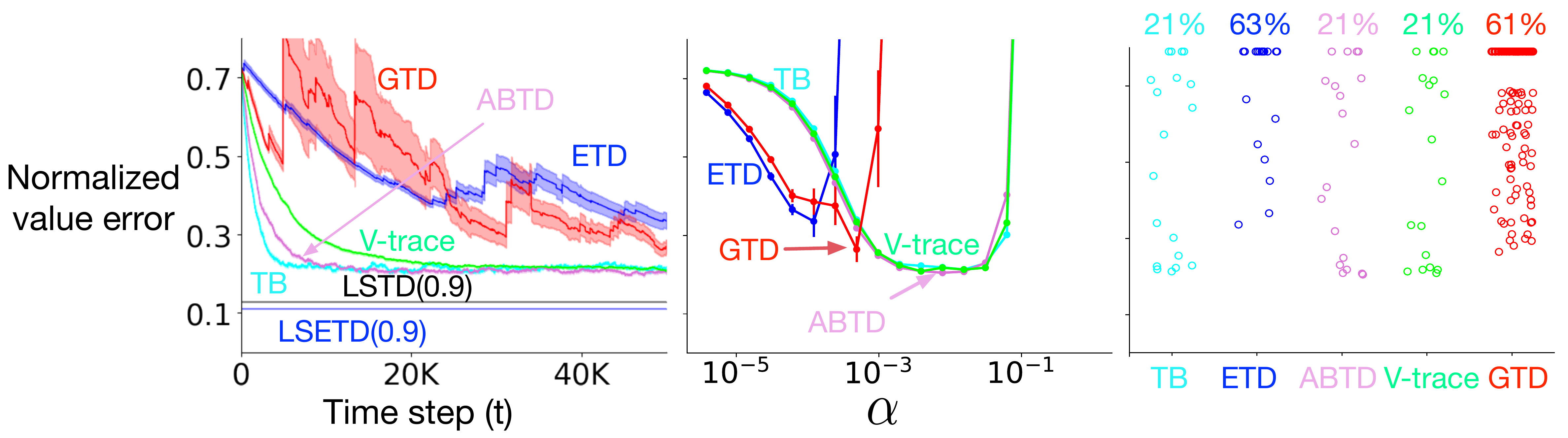}
      \caption{Comparing GTD(0.9), ETD(0.9), ABTD(0.9), Tree Backup(0.9), and V-trace(0.9) on the {\bf High Variance Four Rooms} problem. Plots are optimized for final performance. Middle and right graphs show the final performance. This setting demonstrates issues due to high variance, as GTD(0.9) and ETD(0.9) have much more erratic learning curves. However, V-trace(0.9), ABTD(0.9), and Tree Backup(0.9) manage to have an even better performance compared to the case where $\lambda=0$. Note that ETD(0.9) and GTD(0.9)'s perform more poorly when $\lambda=0.9$ than when $\lambda=0$.}
      \label{fig:trace_four_final_9}
\end{figure}

The overall conclusion from this experiment is that GTD($\lambda$)\footnote{TD($\lambda$) is similarly negatively affected, though its not included in the figures to focus comparison on GTD($\lambda$) and ETD($\lambda$), versus these variance-reduced methods.} is significantly impacted by the large importance sampling ratios, whereas action-dependent bootstrapping methods perform similarly regardless of the value of $\lambda$. When $\lambda$ is equal to zero, the large importance sampling ratios cannot product in the trace and GTD($\lambda$) performs well despite the large magnitude ratios. When $\lambda$ is equal to 0.9, the action-dependent bootstrapping methods are less impacted by variance than GTD(0.9). However, this robustness appears to result in bias. The final error of the action-dependent bootstrapping methods is similar to the error of the LSTD(0) baseline when $\lambda=0$, but there is a large difference between their final error and the error of LSTD(0.9) when $\lambda=0.9$ (see Figures \ref{fig:trace_four_final_0} and \ref{fig:trace_four_final_9}). In fact, these methods effectively set $\lambda$ close to 0, in these high variance situations. Nevertheless, the performance of the action-dependent bootstrapping methods is more reliable than either GTD($\lambda$) or ETD($\lambda$) when used with traces.



\begin{figure}[h!]
      \centering
      \includegraphics[width=0.8\linewidth]{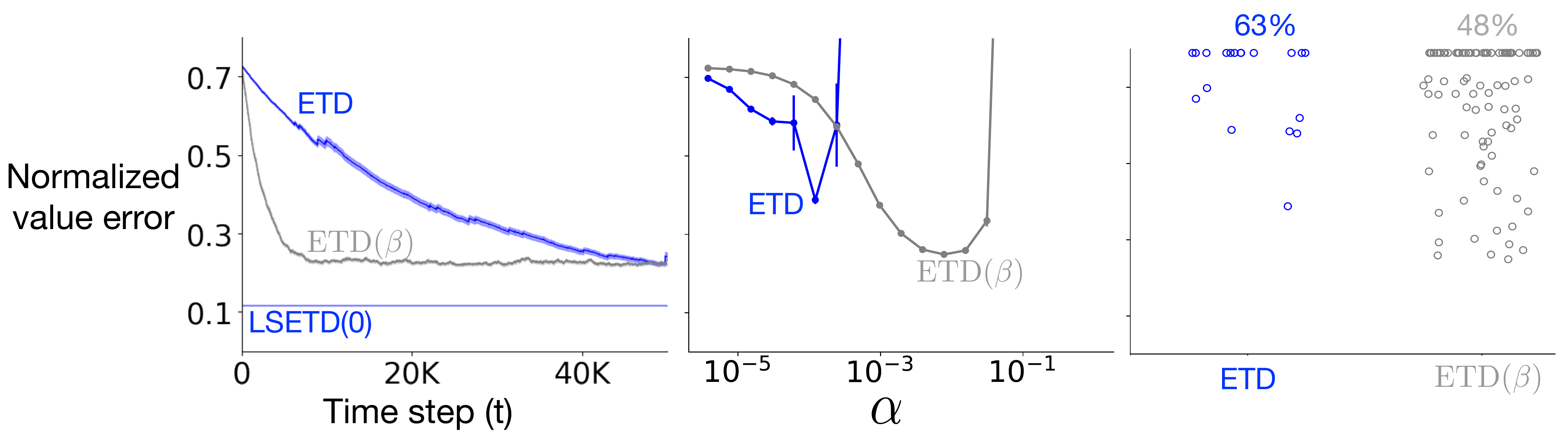}
      \caption{Comparing ETD(0) and ETD(0, $\beta$) on the {\bf High Variance Four Rooms} problem. The key parameters of each method are chosen to optimize AUC. The middle and right graphs plot the AUC divided by number of steps. ETD(0, $\beta$) outperformed ETD(0) in terms of speed of learning, but the methods achieved comparable final performance.}
      \label{fig:ETD_highvariance_four_auc_0}
\end{figure}

\begin{figure}[h!]
      \centering
      \includegraphics[width=0.8\linewidth]{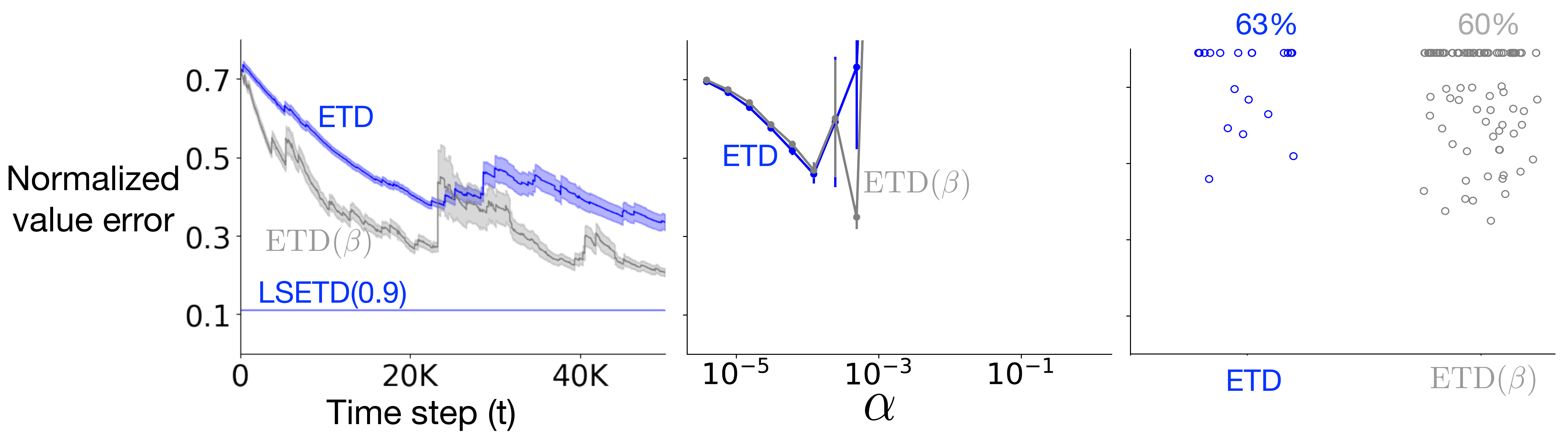}
      \caption{Comparing ETD(0.9) and ETD(0.9, $\beta$) on the {\bf High Variance Four Rooms}.  The key parameters of each method are chosen to optimize AUC. This results show that neither ETD(0.9) nor ETD(0.9, $\beta$) are able to achieve a good performance in the high variance setting when $\lambda=0.9$. }
      \label{fig:ETD_highvariance_four_auc_9}
\end{figure}

As an additional experiment, based on the variability observed for ETD($\lambda$), we investigate if ETD($\lambda, \beta$) is more noticeably different in this high-variance setting, than was observed in Section \ref{sct:ComparingETDAndETDBeta}.
Recall that ETD($\lambda$,$\beta$) algorithm was designed to reduce variance of ETD($\lambda$).
Figures \ref{fig:ETD_highvariance_four_auc_0} and \ref{fig:ETD_highvariance_four_auc_9} summarize the main results of the experiment for $\lambda$ equal to zero and 0.9. In the experiment without eligibility traces, ETD($\lambda$,$\beta$) learns faster, but converges to a biased solution---far from the error of the LSETD($\lambda$) baseline. ETD($\lambda$,$\beta$) also exhibits robustness to choice of the step-size parameter value, and more generally many parameter settings generate good performance. However, with longer traces ($\lambda=0.9$), both methods exhibit high variance, slow learning, and sensitivity to parameter choice.

%

\section{How to best incorporate importance sampling}
\label{sct:ImportanceSamplingPlacement}

The importance sampling ratio is often used in off-policy learning to correct for the differences that might exist between the behavior and target policies. However, it is used in various ways to correct for the differences between the policies. Specifically, there exist two approaches in the literature that we examine here. In this section, we assume that the discount factor, $\gamma$, and the trace parameter, $\lambda$, are both constants for simplicity of exposition. All of our results, however, extend to varying $\lambda$ and $\gamma$.


First we rewrite the earlier Off-policy TD($\lambda$), in an equivalent form, that will make it more straightforward to see that the $\rho$ could have been used slightly differently.
Recall the updates mentioned earlier, as given in the original work on Off-policy TD($\lambda$) (Precup, Sutton \& Dasgupta, 2001):
\begin{align*}
\vecw_{t+1} & \leftarrow \vecw_t + \alpha \delta_t \vecz_t \\
\vecz_t^\rho & \leftarrow \rho_t\left( \gamma \lambda \vecz_{t-1} + \vecx_{t}  \right)  \textnormal{\quad with } \vecz_{-1}^\rho = \bf{0}\\
\delta_t & \defeq R_{t+1} + \gamma \vecw_{t}^{\tr}\vecx_{t+1} - \vecw_{t}^{\tr}\vecx_{t}
\end{align*}
These update rules have at times been written in a different way for Off-policy TD($\lambda$) (e.g., Yu, 2015; van Hasselt, Mahmood \& Sutton, 2014):
\begin{align*}
\vecw_{t+1} & \leftarrow \vecw_t + \alpha \rho_t \delta_t \vecz_t^\prime\\
\vecz_t^\prime & \leftarrow \rho_{t-1} \gamma \lambda \vecz_{t-1}^\prime  + \vecx_{t} \textnormal{\quad with } \vecz_{-1}^\prime = \bf{0}
\end{align*}
These updates are actually equivalent, as we show in Appendix \ref{app_importancesampling}, with
$\dt\zt$ equal to the product of $\rho_t\dt\ztp$ on each step given that $\vecz_{-1} = \vecz_{-1}^\prime=\bf{0}$.

We can specify a different off-policy update form that is only equivalent in expectation.
There are a few papers (e.g., Geist \& Scherrer, 2014) that specify the Off-policy TD($\lambda$) update as follows:
\begin{align*}
\vecw_{t+1} & \leftarrow \vecw_t + \alpha \delta_t^{'} \vecz_t^{'}\\
\vecz_t^{'} & \leftarrow \rho_{t-1} \gamma \lambda \vecz_{t-1}^{'}  + \vecx_{t} \textnormal{\quad with } \vecz_{-1}^{'} = \bf{0}\\
\delta_t^{'} & \defeq \rho_t \left( R_{t+1} + \gamma \vecw_{t}^{\tr}\vecx_{t+1}  \right) - \vecw_{t}^{\tr}\vecx_{t}
\end{align*}
The key difference is that the importance sampling ratio only corrects the first part of the TD-error, rather than the whole TD-error.
This correction is nonetheless unbiased, because the importance sampling ratio corrects the distribution over action and next state, not the current state. We formally show this is unbiased in Appendix \ref{app_importancesampling}.

The major difference between these forms is how the TD-error ($\delta_t$) is corrected. Should one correct all three terms in the TD-error or should one just correct the first two?
This choice is not restricted to TD($\lambda$) but can be applied to all twelve methods mentioned before. It is not clear what difference each decision might make. An empirical study on both forms of these methods can help us understand how this choice can impact the performance.

\begin{figure}[h!]
      \centering
      \includegraphics[width=0.7\linewidth]{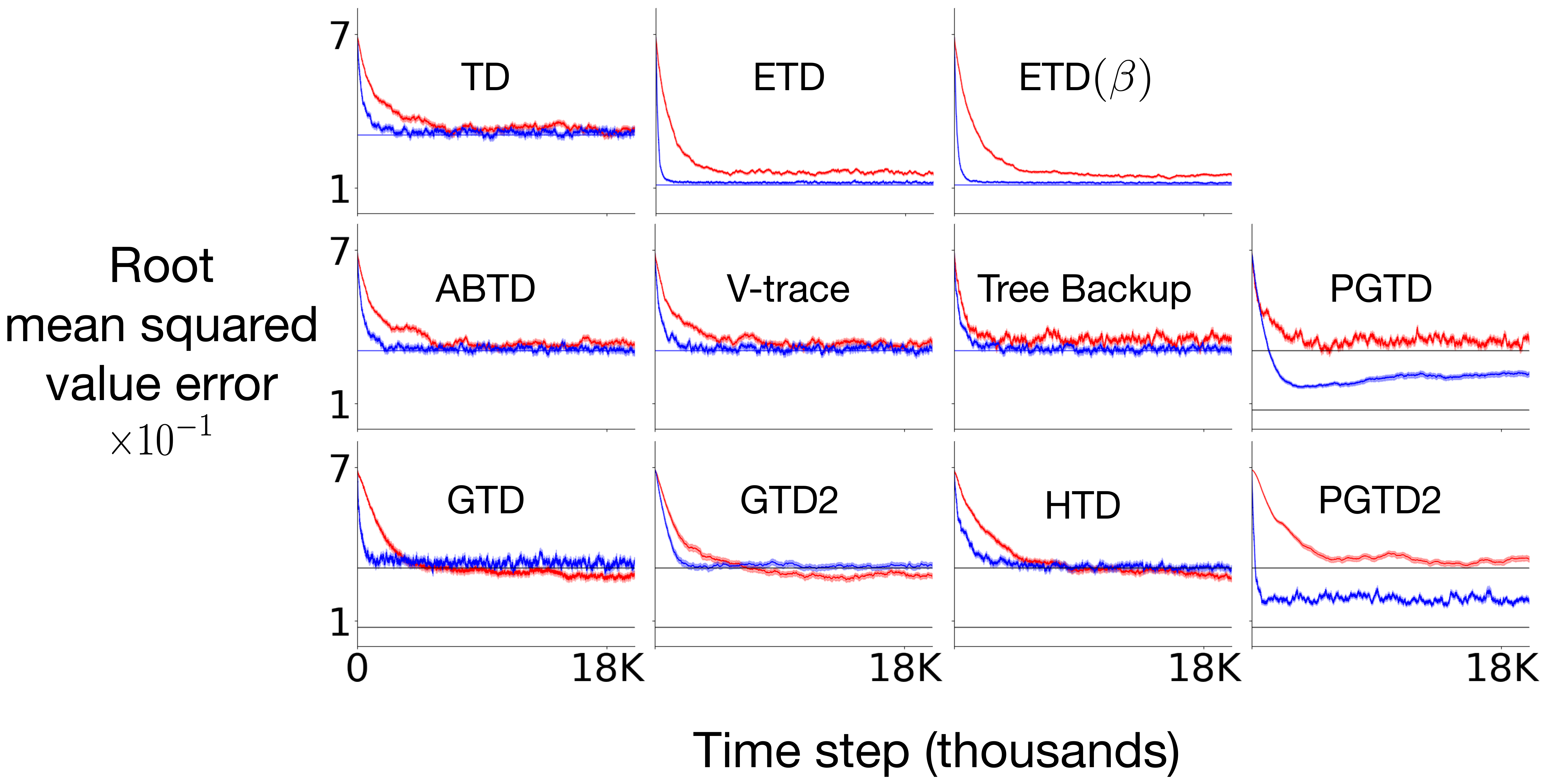}
      \caption{Different $\rho$ placements for the {\bf Collision} problem when $\lambda=0$. The curves are optimized for the area under the curve. The figures compare how $\rho$ placement can affect the performance of each method. Blue is when the whole TD-error term is corrected and red is when $v(S_t$) is not corrected. Corresponding least-squares baseline (TD(0) or ETD(0)) is shown as a solid black line.}
      \label{fig:8StateChickenDifferentRhoLearningCurveBestFinalPerformance}
\end{figure}

\begin{figure}[h!]
      \centering
      \includegraphics[width=0.7\linewidth]{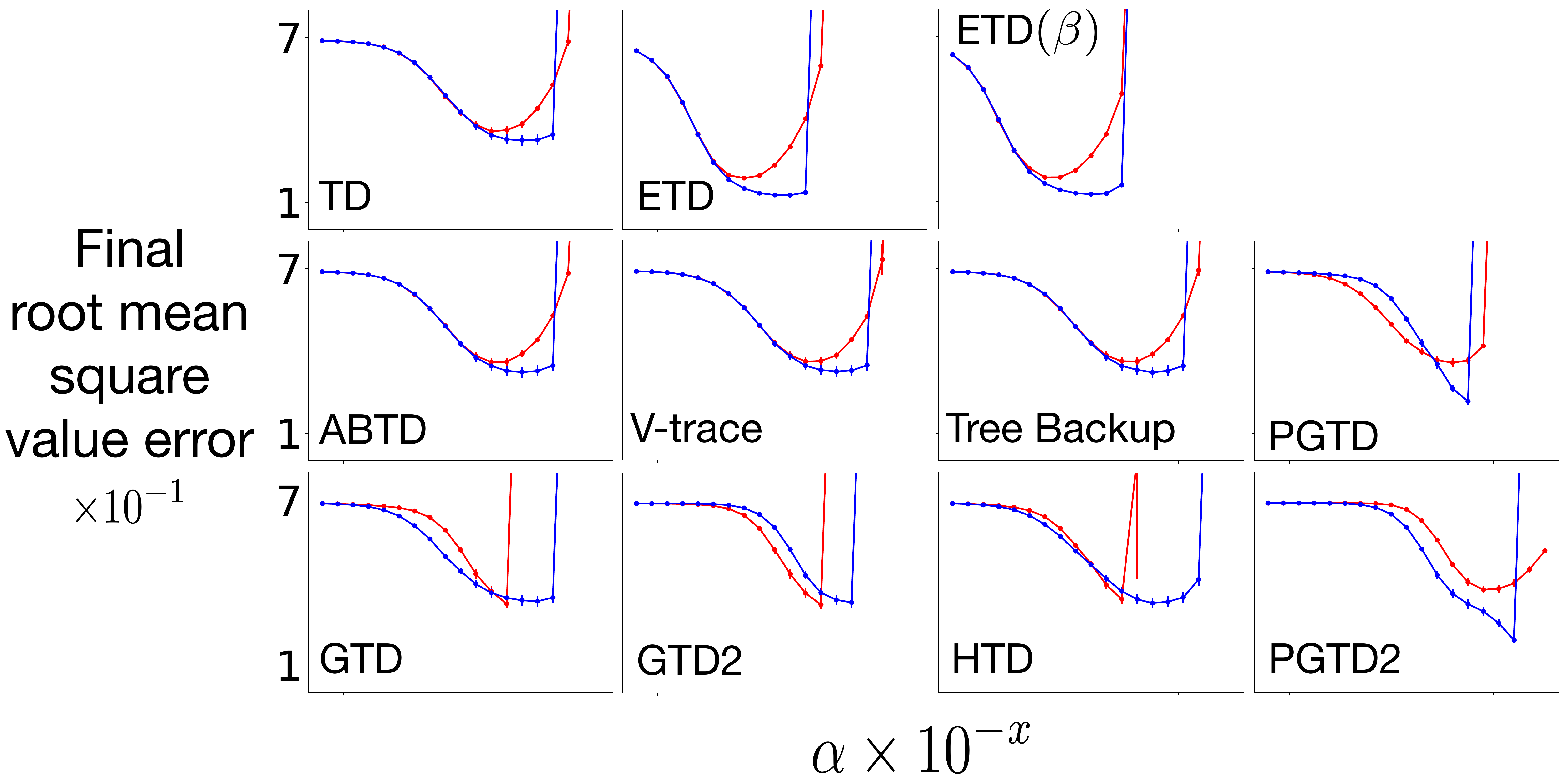}
 \caption{{\bf Collision} problem when $\lambda=0$. The curves are optimized for the area under the curve. The figures compare how $\rho$ placement can affect the performance of each method. Blue is when the whole TD-error term is corrected and red is when $v(S_t$) is not corrected.}
\label{fig:8StateChickenDifferentRhoParameterStudyBestFinalPerformance}
\end{figure}

We studied the differences between all methods when $\vecw_{t}^{\tr}\vecx_{t}$ is corrected by $\rho_t$ and when it is not. We applied all twelve methods in both forms to the Collision problem and measured the performance. Figure \ref{fig:8StateChickenDifferentRhoLearningCurveBestFinalPerformance} shows the learning curve when the whole TD-error term is corrected and when it is corrected partially. As the figure shows, all methods had more variance over different runs and the final performance was also worse for most of the methods when $\vecw_{t}^{\tr}\vecx_{t}$ was not corrected. With the exception of GTD(0), GTD2(0), and HTD(0), other methods had a better final performance when the importance sampling ratio was used to correct the whole TD-error term. For the aforementioned methods, one still should prefer to correct for the whole TD-error term because the gain in the performance when not correcting for the whole TD-error is marginal, but the range of stepsize for which these methods converge shrinks significantly when the whole term is not corrected. Figure \ref{fig:8StateChickenDifferentRhoParameterStudyBestFinalPerformance}, shows the parameter study for different methods when $\lambda=0$.

\section{Conclusions}


In the introduction, we set out to answer a variety of questions. We found that in the two major families, Emphatic-TD has better asymptotic performance than Gradient-TD, particularly for smaller $\lambda$. We found that ETD($\lambda$)'s prior corrections did not seem to result in faster initial learning, though ETD($\lambda$) is significantly less sensitive to the meta-parameter $\lambda$ in our domains. We found that ETD($\lambda$) has some similarities to TD(1), but it outperforms TD(1) in high-variance settings. In the high variance setting, ETD($\lambda$) also seems to be significantly slower than other methods (including but not limited to Gradient-TD).

Hybrid-TD methods and Gradient-TD methods perform similarly suggesting that the hypothesis that Hybrid-TD methods learn faster than Gradient-TD methods does not seem to be true. This is consistent with the findings from (White \& White, 2016). However, HTD($\lambda$) seems to exhibit a clear advantage when $\alpha_h > \alpha$, a setting which might be more useful in practice because it mitigates the risk of divergence under off-policy sampling.

Proximal Gradient-TD methods often outperform their Gradient-TD variants in terms of final performance. This superiority, however, is limited to a very particular parameter setting and comes at the cost of performing twice the number of updates. Proximal-GTD($\lambda$) is often slower than GTD($\lambda$), GTD2($\lambda$), and Proximal-GTD2($\lambda$).

Methods that do not make an explicit use of the importance sampling ratio in some cases perform worse than Gradient-TD methods in terms of final performance when $\lambda=0$. The advantage of these methods can be most clearly seen in tasks with high variance in the importance sampling ratio and long eligibility traces ($\lambda>0$). These methods are fast and converge to a good final performance in such tasks by (aggressively) decreasing the implicit $\lambda$. Further, neither Tree Backup($\lambda$) or V-trace($\lambda$) dominate one another. In some cases, Tree Backup($\lambda$) shows an advantage, in others, V-trace($\lambda$) does. ABTD($\zeta$), however, will sometimes outperform V-trace($\lambda$) and Tree Backup($\lambda$). Even when it does not, it still performs comparably to the best performing method.

ETD($\lambda\,, \beta$) does not seem to provide a clear advantage over ETD($\lambda$). ETD($\lambda\, , \beta$) performs best when its extra parameter is close to $\gamma$, in which case it reduces to ETD($\lambda$). In problems where high variance is expected, ETD($\lambda\, , \beta$) provides a slight advantage over ETD($\lambda$) in terms of parameter sensitivity and learning speed; however, in such tasks, methods that do not explicitly use an importance sampling ratio (V-trace($\lambda$), Tree Backup($\lambda$), and ABTD($\zeta$)) have a clear advantage to ETD($\lambda\, , \beta$).

Finally, the simpler way to add importance sampling corrections (correcting the whole TD-error) almost always performs the same or better than the approach that only corrects part of the TD-error and thus it should be preferred.

\subsubsection*{Acknowledgments} 
The authors thank Banafsheh Rafiee for insights and discussions contributing to the results presented in this paper. The authors gratefully acknowledge funding from Alberta Innovates--Technology Futures, the Natural Sciences and Engineering Research Council of Canada, and Google DeepMind.
\clearpage

\section*{References}

\begin{list}{}{%
\setlength{\topsep}{0pt}%
\setlength{\leftmargin}{0.2in}%
\setlength{\listparindent}{-0.2in}%
\setlength{\itemindent}{-0.2in}%
\setlength{\parsep}{\parskip}%
}%
\item[]
Baird, L. C. (1995). Residual algorithms: Reinforcement learning with function approx- imation. In \textit{Proceedings of the 12th International Conference on Machine Learning}, pp. 30–37. Morgan Kaufmann, San Francisco. Important modifications and errata added to the online version on November 22, 1995.

\item[]
Bhatnagar, S., Precup, D., Silver, D., Sutton, R.S., Maei, H.R., Szepesv\'ari, C., (2009). Convergent temporal-difference learning with arbitrary smooth function approximation. In \textit{Advances in Neural Information Processing Systems 23 (NIPS 2010)}, pp. 1204-1212.

\item[]
Bertsekas, D. P. (2012). \textit{Dynamic Programming and Optimal Control, Volume 2: Approximate Dynamic Programming,} fourth edition. Athena Scientific, Belmont, MA.

\item[]
Dann, C., Neumann, G., Peters, J. (2014). Policy evaluation with temporal differences: A survey and comparison. \textit{Journal of Machine Learning Research,} 15:809–883.

\item[]
Du, S. S., Chen, J., Li, L., Xiao, L., and Zhou, D. (2017) Stochastic variance reduction methods for policy evaluation. \textit{In International Conference on Machine Learning.}

\item[]
Espeholt, L., Soyer, H., Munos, R., Simonyan, K., Mnih, V., Ward, T., Doron, Y., Firoiu, V., Harley, T., Dunning, I. and Legg, S. (2018) IMPALA: Scalable distributed Deep-RL with importance weighted actor-learner architectures. \textit{In International Conference on Machine Learning.}

\item[]
Geist, M., Scherrer, B. (2014). Off-policy learning with eligibility traces: A survey. \textit{Journal
of Machine Learning Research 15}:289–333.

\item[]
Gent, I. P., MacIntyre, E., Prosser, P., Walsh, T. (1997). The scaling of search cost. In \textit{Proceedings of the Fourteenth AAAI Conference on Artificial Intelligence (AAAI-16)}, pp. 315-320.

\item[]
Hallak, As., Tamar, A., Munos, R., Mannor, S. (2016). Generalized emphatic temporal difference learning: Bias-variance analysis. In \textit{Proceedings of the Thirtieth AAAI Conference on Artificial Intelligence (AAAI-16)}, pp. 1631–1637. AAAI Press, Menlo Park, CA.

\item[]
Hackman, L. (2012). \textit{Faster Gradient-TD Algorithms}. M.Sc. thesis, University of Alberta, Edmonton.

\item[]
Henderson, P., Islam, R., Bachman, P., Pineau, J., Precup, D., Meger, D. (2017). Deep reinforcement learning that matters. ArXiv:1709.06560.

\item[]
Jaderberg, M., Mnih, V., Czarnecki, W. M., Schaul, T., Leibo, J. Z., Silver, D., Kavukcuoglu, K. (2016). Reinforcement learning with unsupervised auxiliary tasks. ArXiv:1611.05397.

\item[]
Juditsky, A., Nemirovski, A., Tauvel, C. (2011). Solving variational inequalities with stochastic mirror-prox algorithm. \textit{Stochastic Systems, 1(1)}, 17-58.

\item[]
Kolter, J. Z. (2011). The fixed points of Off-policy TD. In \textit{Advances in Neural Information Processing Systems 24 (NIPS 2011)}, pp. 2169–2177. Curran Associates, Inc.

\item[]
Lillicrap, T. P., Hunt, J. J., Pritzel, A., Heess, N., Erez, T., Tassa, Y., Silver, D., Wierstra, D. (2015). Continuous control with deep reinforcement learning. ArXiv:1509.02971.

\item[]
Littman, M. L., Sutton, R. S., Singh (2002). Predictive representations of state. In \textit{Advances in Neural Information Processing Systems 14 (NIPS 2001)}, pp. 1555-1561. MIT Press, Cambridge, MA.

\item[]
Liu B, Liu J, Ghavamzadeh M, Mahadevan S, Petrik M (2015). Finite-Sample Analysis of Proximal Gradient TD Algorithms. In \textit{Proceedings of the 31st International Conference on Uncertainty in Artificial Intelligence (UAI-2015)}, pp. 504-513. AUAI Press Corvallis, Oregon.

\item[]
Liu B, Liu J, Ghavamzadeh M, Mahadevan S, Petrik M (2016). Proximal Gradient Temporal Difference Learning Algorithms. In \textit{Proceedings of the 25th International Conference on Artificial Intelligence (IJCAI-16)}, pp. 4195-4199. AAAI Press.

\item[]
Mahadevan, S., Liu, B., Thomas, P., Dabney, W., Giguere, S., Jacek, N., Gemp, I., Liu, J. (2014). Proximal reinforcement learning: A new theory of sequential decision making in primal-dual spaces. ArXiv:1405.6757.

\item[]
Mahmood, A. R., Yu, H., Sutton, R. S. (2017). Multi-step off-policy learning without importance sampling ratios. ArXiv:1702.03006.

\item[]
Maei, H. R. (2011). \textit{Gradient temporal-difference learning algorithms}. Ph.D. thesis, University of Alberta, Edmonton.

\item[]
Maei, H. R., Sutton, R. S. (2010). GQ($\lambda$): A general gradient algorithm for temporal-difference prediction learning with eligibility traces. \textit{In Proceedings of the 3rd Conference on Artificial General Intelligence,} pp. 91–96.

\item[]
Mnih, V., Kavukcuoglu, K., Silver, D., Rusu, A. A., Veness, J., Bellemare, M. G., Graves, A., Riedmiller, M., Fidjeland, A. K., Ostrovski, G., Petersen, S., Beattie, C., Sadik, A., Antonoglou, I., King, H., Kumaran, D., Wierstra, D., Legg, S., Hassabis, D. (2015). Human level control through deep reinforcement learning. \textit{Nature, 518}(7540):529–533.

\item[]
Munos, R., Stepleton, T., Harutyunyan, A., Bellemare, M. (2016). Safe and efficient off-policy reinforcement learning. In \textit{Advances in Neural Information Processing Systems 29 (NIPS 2016)}, pp. 1046–1054. Curran Associates, Inc.

\item[]
Precup, D., (2000). Eligibility traces for off-policy policy evaluation. \textit{Computer Science Department Faculty Publication Series}, pp.80.

\item[]
Precup, D., Sutton, R. S., Dasgupta, S. (2001). Off-policy temporal-difference learning with function approximation. In \textit{Proceedings of the 18th International Conference on Machine Learning}, pp. 417–424.

\item[]
Precup, D., Sutton, R. S., Singh, S. (2000). Eligibility traces for off-policy policy evaluation. In \textit{Proceedings of the 17th International Conference on Machine Learning}, pp. 759–766. Morgan Kaufmann.

\item[]
Ring, M. B. (in preparation). Representing knowledge as forecasts (and state as knowledge).

\item[]
Sutton, R. S. (1988). Learning to predict by the methods of temporal differences. \textit{Machine Learning 3}:9–44, erratum p. 377.
also

\item[]
Sutton, R. S. (1996). Generalization in reinforcement learning: Successful examples using sparse coarse coding. In \textit{Advances in Neural Information Processing Systems 8 (NIPS 1995),} pp. 1038–1044. MIT Press, Cambridge, MA.

\item[]
Sutton, R. S., Barto, A. G. (2018). \textit{Reinforcement Learning: An Introduction,} second edition. MIT press.

\item[]
Sutton, R. S., Maei, H. R., Precup, D., Bhatnagar, S., Silver, D., Szepesv\'ari, Cs., Wiewiora, E. (2009). Fast gradient-descent methods for temporal-difference learning with linear function approximation. In \textit{Proceedings of the 26th International Conference on Machine Learning}, pp. 993–1000, ACM.

\item[]
Sutton, R. S., Mahmood, A. R., Precup, D., van Hasselt, H. (2014). A new Q($\lambda$) with in- terim forward view and Monte Carlo equivalence. In \textit{Proceedings of the 31st International Conference on Machine Learning.} JMLR W\&CP 32(2).

\item[]
Sutton, R. S., Mahmood, A. R., White, M. (2016). An emphatic approach to the problem of off-policy temporal- difference learning. \textit{Journal of Machine Learning Research, 17}(73):1–29.

\item[]
Sutton, R. S., Modayil, J., Delp, M., Degris, T., Pilarski, P. M., White, A., Precup, D. (2011). Horde: A scalable real-time architecture for learning knowledge from unsuper- vised sensorimotor interaction. In \textit{Proceedings of the 10th International Conference on Autonomous Agents and Multiagent Systems}, pp. 761–768, Taipei, Taiwan.

\item[]
Sutton, R. S., Precup, D., Singh, S. (1999). Between MDPs and semi-MDPs: A framework for temporal abstraction in reinforcement learning. \textit{Artificial Intelligence, 112}(1-2):181–211.

\item[]
Tanner, B., Sutton, R. S., (2005). TD($\lambda$) networks: temporal-difference networks with eligibility traces. In \textit{Proceedings of the 22nd international conference on Machine learning,} pp. 888-895.

\item[]
Touati, A., Bacon, P. L., Precup, D., Vincent, P. (2017). Convergent tree-backup and retrace with function approximation. ArXiv:1705.09322.

\item[]
van Hasselt H, Mahmood A. R., Sutton R. S. (2014).  Off-policy TD($\lambda$) with a true online equivalence. In \textit{Proceedings of the 30th Conference on Uncertainty in Artificial Intelligence (UAI-2014)}, AUAI Press Corvallis, Oregon.

\item[]
White, A. (2015). \textit{Developing a Predictive Approach to Knowledge.} Ph.D. thesis, University of
Alberta, Edmonton.

\item[]
White, M.. (2017). Unifying Task Specification in Reinforcement Learning. In \textit{Proceedings of the 34th International Conference on Machine Learning.} in PMLR 70:3742-3750.

\item[]
Wang, Z., Bapst, V., Heess, N., Mnih, V., Munos, R., Kavukcuoglu, K., de Freitas, N. (2016). Sample efficient actor-critic with experience replay. ArXiv:1611.01224.

\item[]
Yu, H. (2015). On convergence of emphatic temporal-difference learning. In \textit{Proceedings of the 28th Annual Conference on Learning Theory, JMLR W\&CP 40}. Also ArXiv:1506.02582.

\item[]
Yu, H. (2016). Weak convergence properties of constrained emphatic temporal-difference learning with constant
and slowly diminishing stepsize. \textit{Journal of Machine Learning Research, 17}(220):1–58.

\item[]
Yu, H. (2017). On Convergence of some Gradient-based Temporal-Differences Algorithms for Off-Policy Learning. Arxiv:1712.09652.

\end{list}

\appendix

\newpage

\section{Parameter sweeps}
\label{app:ParameterSweeps}

We swept over different parameters for different methods to find the parameter setting that minimizes the area under the learning curve/final performance. The parameter sweeps were also used for creating the step size plots as well as the parameter sensitivity plots.

\begin{table}[h!]
\centering
\begin{tabular}{|c|c|c|c|c|}
\hline
\multicolumn{2}{|c|}{Algorithms} & $\alpha_h$ or $\beta$ & $\lambda$ or $\zeta$ & $\alpha$ \\ \hline
\multirow{2}{*}{\begin{tabular}[c]{@{}c@{}}prior and\\ posterior\\ TD($\lambda$)\end{tabular}} & \begin{tabular}[c]{@{}c@{}}Off-policy \\ TD($\lambda$)\end{tabular} & --- & \multirow{12}{*}{\begin{tabular}[c]{@{}c@{}}0 \\ and\\  0.9\end{tabular}} & \multirow{12}{*}{\begin{tabular}[c]{@{}c@{}}$2^{-18},2^{-17}$\\, ... ,   $2^0$\end{tabular}} \\ \cline{2-3}
 & \begin{tabular}[c]{@{}c@{}}Alternative life\\ TD($\lambda$)\end{tabular} & --- &  &  \\ \cline{1-3}
\multirow{5}{*}{\begin{tabular}[c]{@{}c@{}}Gradient-TD\\ methods\end{tabular}} & GTD($\lambda$) & \multirow{5}{*}{\begin{tabular}[c]{@{}c@{}} $\alpha_h=0.01\times2^0,$\\ $0.01 \times 2^2,\,0.01 \times 2^4,\,$\\$0.01 \times 2^6,\, 0.01 \times 2^8$\\ $,\, ... \,0.01 \times 2^{14}$\end{tabular}} &  &  \\ \cline{2-2}
 & GTD2($\lambda$) &  &  &  \\ \cline{2-2}
 & HTD($\lambda$) &  &  &  \\ \cline{2-2}
 & Proximal GTD($\lambda$) &  &  &  \\ \cline{2-2}
 & Proxima GTD2($\lambda$) &  &  &  \\ \cline{1-3}
\multirow{2}{*}{\begin{tabular}[c]{@{}c@{}}Emphatic-TD\\ methods\end{tabular}} & ETD($\lambda$) & --- &  &  \\ \cline{2-3}
 & ETD($\lambda,\beta$) & \begin{tabular}[c]{@{}c@{}}$\beta$ = 0.0, 0.2, \\ 0.4, 0.6, 0.8, 1.0\end{tabular} &  &  \\ \cline{1-3}
\multirow{3}{*}{\begin{tabular}[c]{@{}c@{}}Variable-$\lambda$\\ methods\end{tabular}} & Tree Backup($\lambda$) & --- &  &  \\ \cline{2-3}
 & V-trace($\lambda$) & --- &  &  \\ \cline{2-3}
 & ABTD($\zeta$) & --- &  &  \\ \hline
\end{tabular}
\caption[Table caption text]{Parameter settings for different algorithms.}
\end{table}

\section{Pseudocode}
\label{app:PseudoCodes}

In this section we list the pseudo codes for all algorithms empirically studied in the text.
This list provides a concise reference for the update rules for each algorithm.
Each algorithm is grouped with its major family of algorithms: Gradient-TD methods, Proximal Gradient-TD methods, Emphatic-TD methods, and Variable-$\lambda$ TD methods.
We include this list as a single point of reference for each algorithm discussed in the main body of the text.
\newline

\noindent\textbf{TD($\lambda$):}
\begin{align*}
\delta_t \defeq&\,\, R_{t+1} + \gamma_{t+1} \vecw_{t}^{\tr}\vecx_{t+1} - \vecw_{t}^{\tr}\vecx_{t}\\
\vecz^\rho_t \leftarrow& ~\rho_t(\gamma_t \lambda_t \vecz^\rho_{t-1}  +\vecx_t) \textnormal{\qquad with } \vecz_{-1}^\rho = \bf{0}\nonumber\\
\vecw_{t+1} \leftarrow& ~ \vecw_t +\alpha \delta_t \vecz_t^{\rho}
\end{align*}

\noindent\textbf{GTD($\lambda$):}
\begin{align*}
\delta_t \defeq&\,\, R_{t+1} + \gamma_{t+1}\vecw_t^{\tr}\vecx_{t+1} - \vecw_t^{\tr}\vecx_{t}\nonumber\\
\vecz^\rho_t \leftarrow& ~\rho_t(\gamma_t \lambda_t \vecz^\rho_{t-1}  +\vecx_t) \textnormal{\qquad with } \vecz_{-1}^\rho = \bf{0}\nonumber\\
\vech_{t+1} \leftarrow& ~\vech_t + \alpha_h\biggl[\delta_t\vecz^\rho_t - (\vech_t^{\tr}\vecx_t)\vecx_{t}  \biggr] \nonumber\\\
\vecw_{t+1} \leftarrow& ~\vecw_t + \alpha \delta_t \vecz^\rho_t - \alpha \gamma_{t+1} (1-\lambda_{t+1})(\vech_t^{\tr}\vecz^\rho_t)\vecx_{t+1}
\end{align*}

\noindent\textbf{GTD2($\lambda$):}
\begin{align*}
\delta_t \defeq&\,\, R_{t+1} + \gamma_{t+1}\vecw_t^{\tr}\vecx_{t+1} - \vecw_t^{\tr}\vecx_{t}\nonumber\\
\vecz^\rho_t \leftarrow& ~\rho_t(\gamma_t \lambda_t \vecz^\rho_{t-1}  +\vecx_t) \textnormal{\qquad with } \vecz_{-1}^\rho = \bf{0}\nonumber\\
\vech_{t+1} \leftarrow& ~\vech_t + \alpha_h\biggl[\delta_t\vecz^\rho_t - (\vech_t^{\tr}\vecx_t)\vecx_{t}  \biggr] \nonumber\\
\vecw_{t+1} \leftarrow& ~\vecw_t +\alpha (\vech_t^{\tr}\vecx_t)\vecx_t - \alpha \gamma_{t+1} (1 - \lambda_{t+1}) (\vech_t^{\tr}\vecz^\rho_t)\vecx_{t+1}
\end{align*}

\noindent\textbf{HTD($\lambda$):}
\begin{align*}
\delta_t \defeq&\,\, R_{t+1} + \gamma_{t+1}\vecw_t^{\tr}\vecx_{t+1} - \vecw_t^{\tr}\vecx_{t}\nonumber\\
\vecz^\rho_t \leftarrow& ~\rho_t(\gamma \lambda \vecz^\rho_{t-1}  +\vecx_t) \textnormal{\qquad with } \vecz_{-1}^\rho = \bf{0}\nonumber\\
\vecz_t \leftarrow& ~\gamma \lambda \vecz_{t-1}  +\vecx_t\nonumber\\
\vech_{t+1} \leftarrow& ~\vech_t + \alpha_h\biggl[\delta_t\vecz^\rho_t - (\vecx_t - \gamma_{t+1}\vecx_{t+1})(\vech_t^{\tr}\vecz_t)  \biggr] \nonumber\\\
\vecw_{t+1} \leftarrow& ~\vecw_t + \alpha\biggl[\delta_t\vecz^\rho_t + (\vecx_t - \gamma_{t+1}\vecx_{t+1})(\vecz^\rho_t -\vecz_t)^{\tr}\vech_t  \biggr]
\end{align*}

\noindent\textbf{Proximal GTD($\lambda$):}
\begin{align*}
\delta_t \defeq&\,\, R_{t+1} + \gamma_{t+1}\vecw_t^{\tr}\vecx_{t+1} - \vecw_t^{\tr}\vecx_{t}\nonumber\\
\vecz^\rho_t \leftarrow& ~\rho_t(\gamma_t \lambda_t \vecz^\rho_{t-1}  +\vecx_t)\textnormal{\quad with } \vecz_{-1}^\rho = \bf{0}\nonumber\\
\vech_{t+\frac{1}{2}} \leftarrow& ~\vech_{t} + \alpha_\vech \biggl[\delta_{t} \vecz_t^{\rho} - (\vech_{t}^{\tr}\vecx_t)\vecx_t\biggr]\\
\vecw_{t+\frac{1}{2}} \leftarrow& ~\vecw_t +\alpha \delta_{t} \vecz_t^{\rho} -\alpha \gamma_{t+1} (1 - \lambda_{t+1}) (\vech_{t}\vecz_{t}^{\rho} ) \vecx_{t+1} \\
\delta_{t+\frac{1}{2}} \defeq&\,\, R_{t+1} + \gamma_{t+1} \vecw_{t+\frac{1}{2}}^{\tr}\vecx_{t+1} - \vecw_{t+\frac{1}{2}}^{\tr}\vecx_{t}\\
\vech_{t+1} \leftarrow& ~\vech_{t} + \alpha_\vech \biggl[\delta_{t+\frac{1}{2}} \vecz_t^\rho - (\vech_{t+\frac{1}{2}}^{\tr}\vecx_t)\vecx_t\biggr]\\
\vecw_{t+1} \leftarrow& ~\vecw_t +\alpha \delta_{t+\frac{1}{2}} \vecz_t^\rho - \alpha \gamma_{t+1} (1 - \lambda_{t+1}) (\vech_{t+\frac{1}{2}}^{\tr}\vecz_t^\rho) \vecx_{t+1}
\end{align*}

\noindent\textbf{Proximal GTD2($\lambda$):}
\begin{align*}
\delta_t \defeq&\,\, R_{t+1} + \gamma_{t+1}\vecw_t^{\tr}\vecx_{t+1} - \vecw_t^{\tr}\vecx_{t}\nonumber\\
\vecz^\rho_t \leftarrow& ~\rho_t(\gamma_t \lambda_t \vecz^\rho_{t-1}  +\vecx_t)\textnormal{\quad with } \vecz_{-1}^\rho = \bf{0}\nonumber\\
\vech_{t+\frac{1}{2}} \leftarrow& ~\vech_{t} + \alpha_\vech \biggl[\delta_{t} \vecz_t^{\rho} - (\vech_{t}^{\tr}\vecx_t)\vecx_t\biggr]\\
\vecw_{t+\frac{1}{2}} \leftarrow& ~\vecw_t +\alpha (\vech_t^{\tr}\vecx_t)\vecx_t -\alpha \gamma_{t+1} (1 - \lambda_{t+1}) (\vech_t^{\tr}\vecz_{t}^{\rho} ) \vecx_{t+1}\\
\delta_{t+\frac{1}{2}} \defeq&\,\, R_{t+1} + \gamma_{t+1} \vecw_{t+\frac{1}{2}}^{\tr}\vecx_{t+1} - \vecw_{t+\frac{1}{2}}^{\tr}\vecx_{t}\\
\vech_{t+1} \leftarrow& ~\vech_{t} + \alpha_\vech \biggl[\delta_{t+\frac{1}{2}} \vecz_t^\rho - (\vech_{t+\frac{1}{2}}^{\tr}\vecx_t)\vecx_t\biggr]\\
\vecw_{t+1} \leftarrow& ~\vecw_t +\alpha (\vech_{t+\frac{1}{2}}^{\tr}\vecx_t)\vecx_t -\alpha  \gamma_{t+1} (1 - \lambda_{t+1}) (\vech_{t+\frac{1}{2}}^{\tr}\vecz_{t}^{\rho} ) \vecx_{t+1}
\end{align*}

\noindent\textbf{ETD($\lambda$):}
\begin{align*}
\delta_t \defeq&\,\, R_{t+1} + \gamma_{t+1}\vecw_t^{\tr}\vecx_{t+1} - \vecw_t^{\tr}\vecx_{t}\nonumber\\
F_t \leftarrow& ~\rho_{t-1}\gamma_t F_{t-1} + I_t \textnormal{\quad with } F_{0} = I_0\\
M_{t}  ~\defeq&\,\, \lambda_t I_t + (1 - \lambda_t)F_t\\
\vecz_t^{\rho} \leftarrow& ~\rho_t \left(\gamma_t \lambda \vecz_{t-1}^{\rho} + M_t \vecx_{t}\right)\textnormal{\quad with } \vecz_{-1}^{\rho} = \bf{0}\\
\vecw_{t+1} \leftarrow& ~ \vecw_t +\alpha \delta_t \vecz_t^{\rho}
\end{align*}

\noindent\textbf{ETD($\lambda\, ,\beta$):}
\begin{align*}
\delta_t \defeq&\,\, R_{t+1} + \gamma_{t+1}\vecw_t^{\tr}\vecx_{t+1} - \vecw_t^{\tr}\vecx_{t}\nonumber\\
F_t \leftarrow& ~\rho_{t-1}\beta F_{t-1} + I_t \textnormal{\quad with } F_{0} = I_0\\
M_{t}  \defeq&\,\, \lambda_t I_t + (1 - \lambda_t)F_t\\
\vecz_t^{\rho} \leftarrow& ~\rho_t \left(\gamma_t \lambda \vecz_{t-1}^{\rho} + M_t \vecx_{t}\right)\textnormal{\quad with } \vecz_{-1}^{\rho} = \bf{0}\\
\vecw_{t+1} \leftarrow& ~ \vecw_t +\alpha \delta_t \vecz_t^{\rho}
\end{align*}

\noindent\textbf{Tree Backup for prediction($\lambda$):}
\begin{align*}
\delta_t^{\rho} \defeq&\,\, \rho_t \biggl( R_{t+1} + \gamma_{t+1}\vecw_t^{\tr}\vecx_{t+1} - \vecw_t^{\tr}\vecx_{t}\biggr)\nonumber\\
\vecz_t \leftarrow& ~\gamma_t \lambda_t \pi_{t-1} \vecz_{t-1} + \vecx_{t}\textnormal{\quad with } \vecz_{-1} = \bf{0}\\
\vecw_{t+1} \leftarrow& ~\vecw_t +\alpha \delta_t^{\rho} \vecz_t
\end{align*}

\noindent\textbf{V-trace($\lambda$):}
\begin{align*}
\delta_t \defeq&\,\, R_{t+1} + \gamma_{t+1}\vecw_t^{\tr}\vecx_{t+1} - \vecw_t^{\tr}\vecx_{t}\nonumber\\
\vecz_t \leftarrow& ~\textnormal{max}(\rho_t, 1) \left(\gamma_t \lambda \vecz_{t-1} + \vecx_{t}\right)\textnormal{\quad with } \vecz_{-1} = \bf{0}\\
\vecw_{t+1} \leftarrow& ~\vecw_t +\alpha \delta_t \vecz_t^{\rho}
\end{align*}

\noindent\textbf{ABTD($\zeta$):}
\begin{align*}
\delta_t^{\rho} \defeq&\,\, \rho_t \biggl( R_{t+1} + \gamma_{t+1}\vecw_t^{\tr}\vecx_{t+1} - \vecw_t^{\tr}\vecx_{t}\biggr)\nonumber\\
\vecz_t \leftarrow& ~\gamma_t \nu_{t-1} \pi_{t-1} \vecz_{t-1} + \vecx_{t}\textnormal{\quad with } \vecz_{-1} = \bf{0}\\
\vecw_{t+1} \leftarrow& ~ \vecw_t +\alpha \delta_t^{\rho} \vecz_t
\end{align*}

\section{Derivations for Importance Sampling Placement}\label{app_importancesampling}
We first show that the placement of the importance sampling correction term, $\rho$, is equivalent between two conventions found in the literature.
The original work on Off-policy TD($\lambda$) (Precup, Sutton, \& Dasgupta, 2001) used:

\begin{align*}
\vecw_{t+1} & \leftarrow \vecw_t + \alpha \delta_t \vecz_t \\
\vecz_t^{\rho} & \leftarrow \rho_t\left( \gamma \lambda \vecz_{t-1}^{\rho} + \vecx_{t}  \right)  \textnormal{\quad with } \vecz_{-1}^{\rho} = \bf{0}\\
\delta_t & \defeq R_{t+1} + \gamma \vecw_{t}^{\tr}\vecx_{t+1} - \vecw_{t}^{\tr}\vecx_{t}
\end{align*}
Other works have used a different placement of $\rho$ (e.g., Yu 2015, van Hasselt, Mahmood \& Sutton, 2014):
\begin{align*}
\vecw_{t+1} & \leftarrow \vecw_t + \alpha \rho_t \delta_t \vecz_t^\prime\\
\vecz_t^\prime & \leftarrow \rho_{t-1} \gamma \lambda \vecz_{t-1}^\prime  + \vecx_{t} \textnormal{\quad with } \vecz_{-1}^\prime = \bf{0}
\end{align*}
We show below that, given $\vecz_{-1} = \vecz'_{-1} = 0$, these updates are equivalent.

We first show that $\dt\zt$ is equal to the product $\rho_t\dt\ztp$ on each step, given that $\vecz_{-1}=\vecz_{-1}^\prime=\bf{0}$.
\begin{align}
\rho_t\dt\ztp &= \rho_t\dt \left[ \rho_{t-1} \gamma\lambda \vecz_{t-1}^\prime + \vecx_t \right] \nonumber \\
&=\rho_t\dt \left[ \rho_{t-1} \gamma \lambda ( \rho_{t-2}\gamma \lambda \vecz_{t-2}^\prime + \vecx_{t-1}) + \vecx_t \right] \nonumber\\
&=\rho_t\dt \left[ \left( \gamma \lambda \right)^2 \rho_{t-1} \rho_{t-2} \vecz_{t-2}^\prime + \gamma \lambda \rho_{t-1} \vecx_{t-1} + \vecx_t \right] \nonumber\\
&=\rho_t\dt \left[ \left( \gamma \lambda \right)^2 \rho_{t-1} \rho_{t-2}       \left( \rho_{t-3} \gamma \lambda \vecz_{t-3}^\prime + \vecx_{t-2} \right)      + \gamma \lambda \rho_{t-1} \vecx_{t-1} + \vecx_t \right] \nonumber\\
&=\rho_t\dt \left[ \left( \gamma \lambda \right)^3 \rho_{t-1} \rho_{t-2} \rho_{t-3} \vecz_{t-3}^\prime + \left( \gamma \lambda \right)^2 \rho_{t-1} \rho_{t-2} \vecx_{t-2} + \gamma \lambda \rho_{t-1} \vecx_{t-1} + \vecx_{t} \right] \nonumber\\
&\vdots \nonumber\\
&=\rho_t\dt \left( \sum_{i=0}^t{\left( \gamma \lambda \right)^i \vecx_{t-i} \prod_{k=1}^i{\rho_{t-k}}} \right) + \rho_t\dt \left( \left( \gamma \lambda \right)^{t+1} \prod_{k=1}^{t+1}{\rho_{t-k}} \right)\vecz_{-1}^\prime \label{eq:RhoPrimeZPrimeResults}
\intertext{assuming that $\vecz_{-1}^\prime=\bf{0}$:}
\rho_t\dt\ztp &=\rho_t\dt \left( \sum_{i=0}^t{\left( \gamma \lambda \right)^i \prod_{k=1}^i{\rho_{t-k}}} \right). \nonumber
\intertext{On the other hand we have:}
\dt\zt^{\rho} &= \dt \left[ \rho_t \left( \gamma \lambda \vecz_{t-1}^\rho + \vecx_t \right) \right] \nonumber\\
&= \rho_t\dt \left[ \gamma \lambda \rho_{t-1} \left( \gamma \lambda \vecz_{t-2}^\rho+\vecx_{t-1} \right)  + \vecx_{t} \right] \nonumber\\
&= \rho_t\dt \left[ \left( \gamma \lambda \right)^2 \rho_{t-1} \vecz_{t-2}^\rho + \gamma \lambda \rho_{t-1} \vecx_{t-1} + \vecx_{t} \right] \nonumber\\
&= \rho_t\dt \left[ \left( \gamma \lambda \right)^2 \rho_{t-1} \left( \rho_{t-2} \left( \gamma \lambda \vecz_{t-3}^{\rho} + \vecx_{t-2} \right) \right) + \gamma \lambda \rho_{t-1} \vecx_{t-1} + \vecx_{t} \right] \nonumber\\
&= \rho_t\dt \left[ \left( \gamma \lambda \right)^3 \rho_{t-1} \rho_{t-2} \vecz_{t-3}^{\rho} + \left( \gamma \lambda \right)^2 \rho_{t-1} \rho_{t-2} \vecx_{t-2} + \gamma \lambda \rho_{t-1} \vecx_{t-1} + \vecx_{t} \right] \nonumber\\
&\vdots \nonumber\\
&=\rho_t\dt \left( \sum_{i=0}^t{\left( \gamma \lambda \right)^i \vecx_{t-i} \prod_{k=1}^i{\rho_{t-k}}} \right) + \rho_t\dt \left( \left( \gamma \lambda \right)^{t+1} \prod_{k=1}^{t}{\rho_{t-k}} \right)\vecz_{-1}^{\rho} \nonumber
\intertext{which is equal to \eqref{eq:RhoPrimeZPrimeResults} given that $\vecz_{-1}^\rho=\bf{0}$.\nonumber}
\end{align}

Next we show that the third update---which only corrects a part of the TD-error---is unbiased:
\begin{align*}
\vecw_{t+1} & \leftarrow \vecw_t + \alpha \delta_t^{'} \vecz_t^{'}\\
\vecz_t^{'} & \leftarrow \rho_{t-1} \gamma \lambda \vecz_{t-1}^{'}  + \vecx_{t} \textnormal{\quad with } \vecz_{-1}^{'} = \bf{0}\\
\delta_t^{'} & \defeq \rho_t \left( R_{t+1} + \gamma \vecw_{t}^{\tr}\vecx_{t+1}  \right) - \vecw_{t}^{\tr}\vecx_{t}
\end{align*}
These update rules are also valid because they are equal to the previous ones in expectation. That is, $\CEb{\dt\zt}{\H_t}=\CEb{\rho_t\dt\ztp}{\H_t}=\CEb{\dtp\ztpp}{\H_t}$ where $\H_t$ is the history up to time $t$ ($\H_t=\{S_0, A_0, S_1, A_1, \hdots, S_{t-1}, A_{t-1}, S_t\}$). It is immediate from what we showed above that $\CEb{\dt\zt}{\H_t}=\CEb{\rho_t\dt\ztp}{\H_t}$. Here we show that $\CEb{\rho_t\dt\ztp}{\H_t}=\CEb{\dtp\ztpp}{\H_t}$:

\begin{align}
\CEb{\rho_t\dt\ztp}{\H_t} &= \ztp\CEb{\rho_t\dt}{\H_t} \qquad\qquad\text{since all of $\ztp$ is part of the history and known} \nonumber \\
&=\ztp\CEb{\rho_{t} \left( R_{t+1} + \gamma \vecw_{t}^{\tr}\vecx_{t+1} - \vecw_{t}^{\tr}\vecx_{t} \right)}{\H_t} \nonumber\\
&=\ztp\CEb{\rho_{t} \left( R_{t+1} + \gamma \vecw_{t}^{\tr}\vecx_{t+1} \right)}{\H_t} -\CEb{\rho_t \left( \vecw_{t}^{\tr}\vecx_{t}\right)}{\H_t}. \label{eq:deltaPrimezPrime}
\intertext{On the other hand:}
\CEb{\dtp\ztp}{\H_t} &= \ztp\CEb{\dtp}{\H_t} \nonumber \\
&=\ztp\CEb{\rho_{t} \left( R_{t+1} + \gamma \vecw_{t}^{\tr}\vecx_{t+1} \right) - \vecw_{t}^{\tr}\vecx_{t}}{\H_t} \nonumber\\
&=\ztp\CEb{\rho_{t} \left( R_{t+1} + \gamma \vecw_{t}^{\tr}\vecx_{t+1} \right)}{\H_t}-\CEb{\vecw_{t}^{\tr}\vecx_{t}}{\H_t} \label{eq:deltaDoublePrimezPrime}
\intertext{which is equal to \eqref{eq:deltaPrimezPrime} if:}
\CEb{\vecw_{t}^{\tr}\vecx_{t}}{\H_t}&=\CEb{\rho_t \left( \vecw_{t}^{\tr}\vecx_{t}\right)}{\H_t}. \nonumber
\intertext{This is true since $\CEb{\vecw_{t}^{\tr}\vecx_{t}}{\H_t}=\vecw_{t}^{\tr}\vecx_{t}$ and:} \nonumber
\CEb{\rho_t \left( \vecw_{t}^{\tr}\vecx_{t}\right)}{\H_t}&=\vecw_{t}^{\tr}\vecx_{t} \underbrace{\CEb{\rho_t}{\H_t}}_{=1}=\vecw_{t}^{\tr}\vecx_{t}.
\end{align}

\section[dummyName]{Derivations for TB($\lambda$), ABTD($\zeta$), and V-trace($\lambda$)}\label{app_abq}

These three algorithms can all be derived in a similar way because they are all variants of GTD($\lambda$) or TD($\lambda$) with a particular choice of $\lambda$ at each step.
Unlike other off-policy methods, ABQ($\zeta$), Retrace($\lambda$) and Tree Backup($\lambda$) do not use importance sampling ratios. Rather, they specify the trace function $\lambda$ to remove explicit use of importance sampling ratios, but still enable convergence.
To understand how this is possible, we derive the TD($\lambda$) version of ABQ($\zeta$), which we call ABTD($\zeta$). Using this, we provide the extensions for Retrace($\lambda$) and Tree Backup($\lambda$) to estimate value functions.

Consider the generalized $\lambda$-return, for a $\lambda$ based on the state and action---as in ABQ($\zeta$)---or the entire transition (White, 2017). Let $\lambda_{t+1} = \lambda(S_{t}, A_{t}, S_{t+1})$ be defined based on the transition $(S_{t}, A_{t}, S_{t+1})$, corresponding to how rewards and discounts are defined based on the transition, $R_{t+1} = r(S_{t}, A_{t}, S_{t+1})$ and $\gamma_{t+1} = \gamma(S_{t}, A_{t}, S_{t+1})$. Then, given a value function $\vhat$, the $\lambda$-return $G_t^\lambda$ for generalized $\gamma$ and $\lambda$ is defined recursively as
\begin{equation}
\Glambda_t \defeq \rho_t \left(R_{t+1} + \gamma_{t+1} \left[ (1-\lambda_{t+1}) \vhat(S_{t+1}) + \lambda_{t+1} \Glambda_{t+1} \right] \right)
\end{equation}
Similarly to ABQ($\zeta$) (Mahmood et al., 2017, Equation 7), this $\lambda$-return can be written using TD-errors
\begin{equation}
\delta_t \defeq R_{t+1} + \gamma_{t+1} \vhat(S_{t+1}) - \vhat(S_t)
\end{equation}
as
\begin{align*}
\Glambda_t
&= \rho_t \left(R_{t+1} + \gamma_{t+1} \vhat(S_{t+1}) - \gamma_{t+1} \lambda_{t+1} \vhat(S_{t+1}) + \gamma_{t+1} \lambda_{t+1} \Glambda_{t+1} \right)\\
&= \rho_t \left( \delta_t + \vhat(S_{t}) + \gamma_{t+1} \lambda_{t+1} \left[\Glambda_{t+1} - \vhat(S_{t+1})\right] \right)\\
&= \rho_t \delta_t + \rho_t\vhat(S_{t}) + \rho_t \gamma_{t+1} \lambda_{t+1} \left(\rho_{t+1} \delta_{t+1} + \rho_{t+1} \gamma_{t+2} \lambda_{t+2} \left[\Glambda_{t+2} - \vhat(S_{t+2})\right] \right) \\
&= \rho_t \sum_{n=t}^\infty (\rho_{t+1} \lambda_{t+1} \gamma_{t+1})^n \delta_t + \rho_t \vhat(S_{t})
\end{align*}
where
\begin{equation}
\left(\rho_{t+1} \lambda_{t+1} \gamma_{t+1} \right)^n \defeq \prod_{i=t+1}^n \rho_i \lambda_i \gamma_i
.
\end{equation}

This return differs from the return used by ABQ($\zeta$), because it corresponds to the return from a state, rather than the return from a state and action. In ABQ($\zeta$), the goal is to estimate the action-value for a given state and action. For ABTD($\zeta$), the goal is to estimate the value for a given state. For the return from a state $S_t$, we need to correct the distribution over actions $A_t$ with importance sampling ratio $\rho_t$. For ABQ($\zeta$), the correction with $\rho_t$ is not necessary, and importance sampling corrections only need to be computed for future states and actions, with $\rho_{t+1}$ onward. For ABTD($\zeta$), therefore, unlike ABQ($\zeta$), not all importance sampling ratios can be avoided. We can, however, still set $\lambda$ in a similar way to ABQ($\zeta$) to mitigate the variance effects of importance sampling, resulting in the below ABTD($\zeta$) algorithm.

The trace function in ABTD($\zeta$) is set to ensure $\rho_t \lambda_{t+1}$ is well-behaved.
For some constant $\psi > 0$, let
\begin{align}
\lambda(S_{t}, A_{t}, S_{t+1}) &= \nu(\psi, S_t, A_t) b(S_t, A_t)\\
&\text{where} \ \ \ \  \nu(\psi, S_t, A_t) \defeq \min \left(\psi, \frac{1}{\max(b(S_t, A_t), \pi(S_t, A_t)}\right) \nonumber
.
\end{align}
In the $\lambda$-return, then
\begin{equation*}
\rho_t \lambda_{t+1} =  \frac{\pi(S_t, A_t)}{b(S_t,A_t)} \nu(\psi, S_t, A_t) b(S_t, A_t) = \nu(\psi, S_t, A_t) \pi(S_t, A_t)
.
\end{equation*}
This removes the importance sampling ratios from the eligibility trace.

The resulting ABTD($\zeta$) algorithm can be written as the standard GTD($\lambda$) algorithm, for a particular setting of $\lambda$.
However, it is more numerically stable to explicitly write the updates as a function of $\nu$ and $\pi$.
The GTD($\lambda$) algorithm, with this $\lambda$, is called ABTD($\zeta$), with updates
\begin{align*}
\nu_{t-1} &= \min \left(\psi, (\max(b_{t-1}, \pi_{t-1})^{-1}\right)\\
\vecz_t & = \gamma_t \nu_{t-1} \pi_{t-1} \vecz_{t-1} + \vecx_{t}\\
\delta_t & = R_{t+1} + \gamma_{t+1} \vecw_{t}^{\tr}\vecx_{t+1} - \vecw_{t}^{\tr}\vecx_{t}\\
\vecw_{t+1} & = \vecw_t +\alpha \rho_t \left( \delta_t \vecz_t - \gamma_{t+1} \left( 1 - \nu_t b_t \right) \left(\vecz_t^{\tr} \vech_t \right) \vecx_{t+1} \right)\\
\vech_{t+1} & = \vech_t +\beta \left( \rho_t \delta_t \vecz_t - \left( \vech_t^{\tr} \vecx_t \right) \vecx_t \right)
\end{align*}
This reduces to a TD algorithm, that uses this $\lambda$, if we set $\vech_t = \zerovec$, because this causes the correction term $ \gamma_{t+1} \rho_t \left( 1 - \nu_t b_t \right) \left(\vecz_t^{\tr} \vech_t \right) \vecx_{t+1}$ to be omitted.


Finally, we can adapt Retrace($\lambda$) and Tree Backup($\lambda$) for policy evaluation.
Mahmood, Yu, \& Sutton (2017) showed that Retrace($\lambda$) can be specified with a particular setting of $\nu_t$ (in their Equation 36). We can similarly obtain Retrace($\lambda$) for prediction with the setting
\begin{align*}
\nu_{t-1} &= \psi \min \left(\frac{1}{\pi_{t-1}}, \frac{1}{b_{t-1}}\right)
\end{align*}
For Tree Backup($\lambda$) for prediction, the setting for $\nu_t$ is any constant value in $[0,1]$ (see Algorithm 2 (Precup et al., 2000)).
The original Tree Backup($\lambda$) algorithm was derived only for the tabular setting, but the selection of $\lambda_t$---or correspondingly of $\nu_t$---more generally defines the returns, and so extends to a function approximation setting.

In this paper, we used a simplified version of ABTD($\zeta$) algorithm where $\vech_t$ is equal to 0 for all $t$. This algorithm does not have a gradient correction and thus is not guaranteed to converge. We use this non-convergent version of the method as it's really close to that of Tree Backup($\lambda$) and V-trace($\lambda$). Our ABTD($\zeta$) method is fully described by the following equations:

\begin{align*}
\delta_t^{\rho} ~\defeq~& \rho_t \biggl( R_{t+1} + \gamma_{t+1}\vecw_t^{\tr}\vecx_{t+1} - \vecw_t^{\tr}\vecx_{t}\biggr)\nonumber\\
\vecz_t \leftarrow& ~\gamma_t \nu_{t-1} \pi_{t-1} \vecz_{t-1} + \vecx_{t}\textnormal{\quad with } \vecz_{-1} = \bf{0}\\
\vecw_{t+1} \leftarrow& ~ \vecw_t +\alpha \delta_t^{\rho} \vecz_t
\end{align*}

This simplified version of the ABTD($\zeta$) algorithm is closely related to Retrace($\lambda$) and also Tree Backup($\lambda$). These three methods can all be looked at as a specific way of choosing the value of $\lambda_t$ for the original TD($\lambda$) algorithm. V-trace($\lambda$) update rules are the same as ABTD($\zeta$) with the difference:

\begin{align*}
\vecz_t \leftarrow& ~\textnormal{max}(\rho_t,1) \gamma_t \lambda_t \pi_{t-1} \vecz_{t-1} + \vecx_{t}\\
\end{align*}

\noindent Tree Backup($\lambda$) is very similar to ABTD($\zeta$) but it has a different eligibility trace update:

\begin{align*}
\vecz_t \leftarrow& ~\gamma_t \lambda_t \pi_{t-1} \vecz_{t-1} + \vecx_{t}\\
\end{align*}

Another way to look at the three variance reduction methods is to assume that the term $\rho_{t-1}$ is actually always present in the eligibility trace update but it is not explicit. For Tree Backup($\lambda$), $\rho_{t-1}\lambda_t = \pi_{t-1} \lambda$ (See equation \ref{eq:TBTrace}). Meaning that:
\begin{align*}
\pi_{t-1}\lambda = \frac{\pi_{t-1}}{b_{t-1}}\lambda_t,
\intertext{which means}
\lambda_t = b_{t-1}\lambda,
\intertext{which is actually how we set $\lambda_t$ for Tree Backup($\lambda$).}
\end{align*}
The same assumption (implicit existence of $\rho_{t-1}$) in the eligibility trace update for ABTD($\zeta$) (See equation \ref{eq:ABTDTrace}) gives:
\begin{align*}
\nu_{t-1}\pi_{t-1} = \rho_{t-1}\lambda_t,
\intertext{which leads to}
\lambda_t = \nu_{t-1}b_{t-1},
\intertext{which is actually equal to the value of $\lambda_t$ that we set before.}
\end{align*}
Assumption of implicit existence of $\rho_{t-1}$ in equation \ref{eq:V-traceTrace} (for V-trace($\lambda$)) gives:
\begin{align*}
\min \left(\bar{c},\rho_{t-1}\right) \lambda = \rho_{t-1}\lambda_t,
\end{align*}
\noindent which leads to
\begin{align*}
\lambda_t &= \frac{\min\left(\bar{c},\rho_{t-1}\right)}{\rho_{t-1}}\lambda\\
&=\min\left({\bar{c}, \rho_{t-1}}\right) \frac{\lambda}{\rho_{t-1}}\\
&=\min\left(\frac{\bar{c}}{\rho_{t-1}}, 1\right)\lambda\\
&=\min\left(\frac{\bar{c}b{t-1}}{\pi_{t-1}}, 1\right)\lambda\\
&=\min\left(\frac{\bar{c}}{\pi_{t-1}}, \frac{1}{b{t-1}}\right)\lambda b{t-1},
\end{align*}

\textbf{An alternative but incorrect extension of ABQ($\zeta$) to ABTD($\zeta$)}\label{app_abq}

The ABQ($\zeta$) algorithm specifies $\lambda$ to ensure that $\rho_t \lambda_t$ is well-behaved, whereas we specified $\lambda$ so that $\rho_t \lambda_{t+1}$ is well-behaved.
This difference arises from the fact that for action-values, the immediate reward and next state are not re-weighted with $\rho_t$. Consequently, the $\lambda$-return of a policy from a given state and action is
\begin{equation}
R_{t+1} + \gamma_{t+1} \left[ (1-\lambda_{t+1}) \vhat(S_{t+1}) + \rho_{t+1} \lambda_{t+1} \Glambda_{t+1} \right]
.
\end{equation}
To mitigate variance in ABQ($\zeta$) when learning action-values, therefore, $\lambda_{t+1}$ should be set to ensure that $\rho_{t+1} \lambda_{t+1}$ is well-behaved. For ABTD($\zeta$), however, $\lambda_{t+1}$ should be set to mitigate variance from $\rho_t$ rather than from $\rho_{t+1}$.

To see why more explicitly, the central idea of these methods is to avoid importance sampling altogether: this choice ensures that the eligibility trace does not include importance sampling ratios. The eligibility trace $\vecz^a_t$ in TD when learning
action values is
\begin{equation}
\vecz^a_t = \rho_{t} \lambda_{t} \gamma_t \vecz^a_{t-1} + \vecx^a_t
\end{equation}
for state-action features $\vecx^a_t$.
For $\rho_t \lambda_t = \nu_t \pi_t$, this trace reduces to $\vecz^a_t = \nu_{t} \pi_{t} \gamma_t \vecz^a_{t-1} + \vecx^a_t$ (Equation 18, Mahmood et al., 2017).
For ABTD($\zeta$), one could in fact also choose to set $\lambda_{t}$ so that $\rho_t \lambda_t = \nu_t \pi_t$ instead of $\rho_t \lambda_{t+1} = \nu_t \pi_t$. However, this would result in eligibility traces that still contain importance sampling ratios.
The eligibility trace in TD when learning state-values is
\begin{equation}
\vecz_t = \rho_{t-1} \lambda_{t} \gamma_t \vecz_{t-1} + \vecx_t
\end{equation}
Setting $\rho_t \lambda_{t} = \nu_t \pi_t$ would result in the update
$\vecz_t = \rho_{t-1} \nu_t \frac{\pi_t}{\rho_t} \gamma_t \vecz_{t-1} + \vecx_t$,
which does not remove important sampling ratios from the eligibility trace. Rather,
the corresponding update for policy evaluation requires $\rho_{t-1} \lambda_{t} = \nu_{t-1} \pi_{t-1}$,{}
giving the above ABTD($\zeta$) in Section \ref{subsct:ABTDDerivation}.

\section[dummyTitle]{The role of $\lambda$ in Emphatic TD($\lambda$)}
\label{app:EmphaticTDComparedWithTD1}
The results in Section \ref{sct:ComparingTheThreeMainFamilies}, suggest that Emphatic TD($\lambda$)'s performance is basically the same for both $\lambda$ equal zero or 0.9, whereas most other methods such as Gradient TD($\lambda$) perform significantly differently with and without eligibility traces. In fact, in both the Collision problem and the Four Rooms problem, ETD($\lambda$) performs nearly identically to Off-policy TD(1) (not shown). If we consider the update equations of ETD($\lambda$) (see Section \ref{sec:etd}), we can see that the follow-on trace $F_t$ stores historical information about prior values of $\gamma_t$ and $\rho_t$ even when $\lambda=0$, similar to an eligibility trace $\vece_t$. This raises an interesting question: how is Emphatic TD($\lambda$) different from Off-policy TD(1) and is there any reason to prefer one or the other. Recall that Off-policy TD(1) is equivalent to a Monte Carlo Gradient Descent algorithm and thus converges under function approximation and off-policy sampling just like ETD($\lambda$).

\begin{figure}[H]
      \centering
      \includegraphics[width=0.7\linewidth]{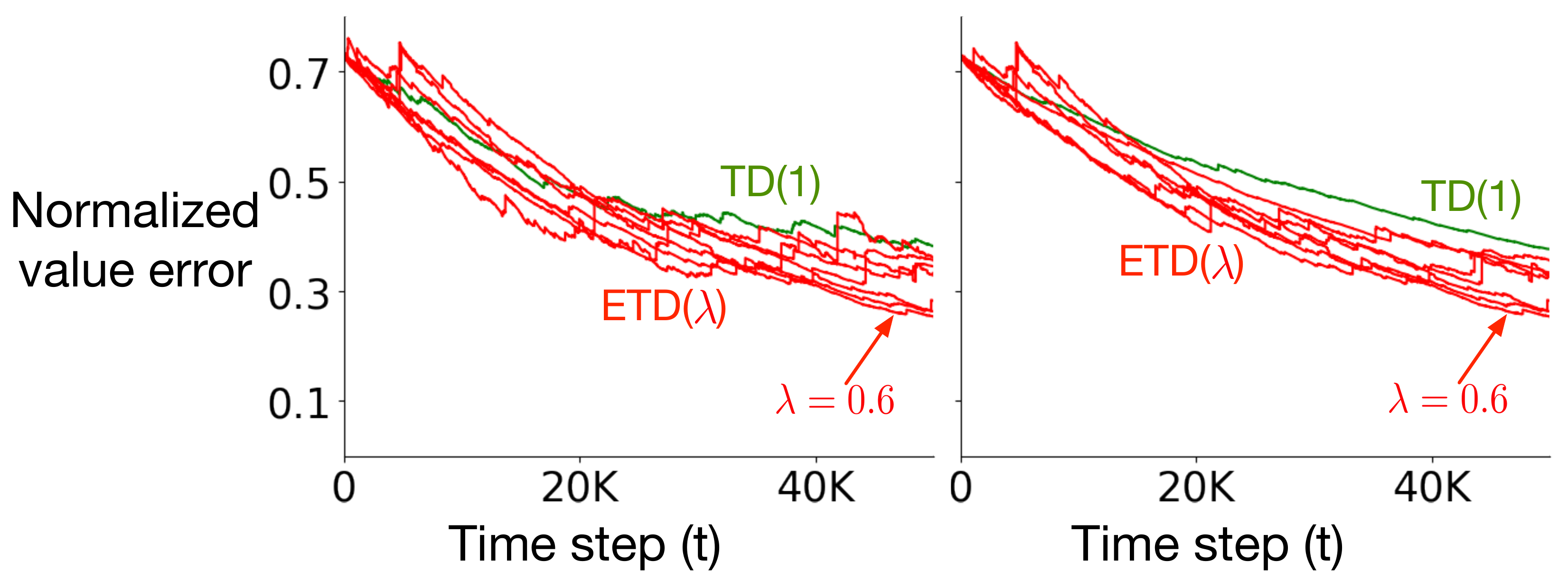}
      \caption{ETD($\lambda$) (red) is compared with Off-policy TD(1) (green) on the \textbf{High Variance Four Rooms} problem. The performance of ETD($\lambda$) is shown with $\lambda=\{0.0\,, 0.2\,, 0.4\,, 0.6\,, 0.8\,, 0.9\,, 0.95\,, 1.0\}$. For each value of $\lambda$, the learning curve is shown for the value of step-size that minimizes either the area under the learning curve (left) or final performance (right). The results are averaged over 200 independent runs. ETD($\lambda$) outperforms Off-policy TD(1) in this problem.}
      \label{appfig:ETDLambdaVsTD1}
\end{figure}

To illustrate the difference between ETD($\lambda$) and Off-policy TD(1) we compared the performance of ETD($\lambda$) with several different values of $\lambda$ on the High-variance Four Rooms problem in Figure \ref{appfig:ETDLambdaVsTD1}. These results clearly indicate two things: (1) different values of $\lambda$ do impact both the speed of learning and the final performance of ETD($\lambda$), and (2) ETD($\lambda$) can outperform Off-policy TD. The results are statistically significant (error bars excluded for clarity).

We still only have rudimentary understanding of the ETD($\lambda$) algorithm's performance with respect to $\lambda$, and relationship to Off-policy TD(1) in both theory and practice. In Figure~\ref{appfig:ETDLambdaVsTD1} we provide the first empirical results highlighting that these methods are indeed different, but further exploration and analysis is needed.

\section{Comparing prior and posterior correction methods}
\label{app:ComparingTheThreeMainFamilies}
In this section---and following sections---we present additional results beyond those presented in the main text.
For every comparison, we investigated algorithms across two metrics and with/without eligibility traces.
Further, all comparisons were done on both benchmark environments.

In the main text, we discussed the results that are most informative; those that show the most profound difference in methods, or those that empirically demonstrate some phenomenon we discuss theoretically.
In the corresponding appendices, we show all other obtained results; including those that show negligible differences between algorithms.

\begin{figure}[H]
      \centering
      \includegraphics[width=0.8\linewidth]{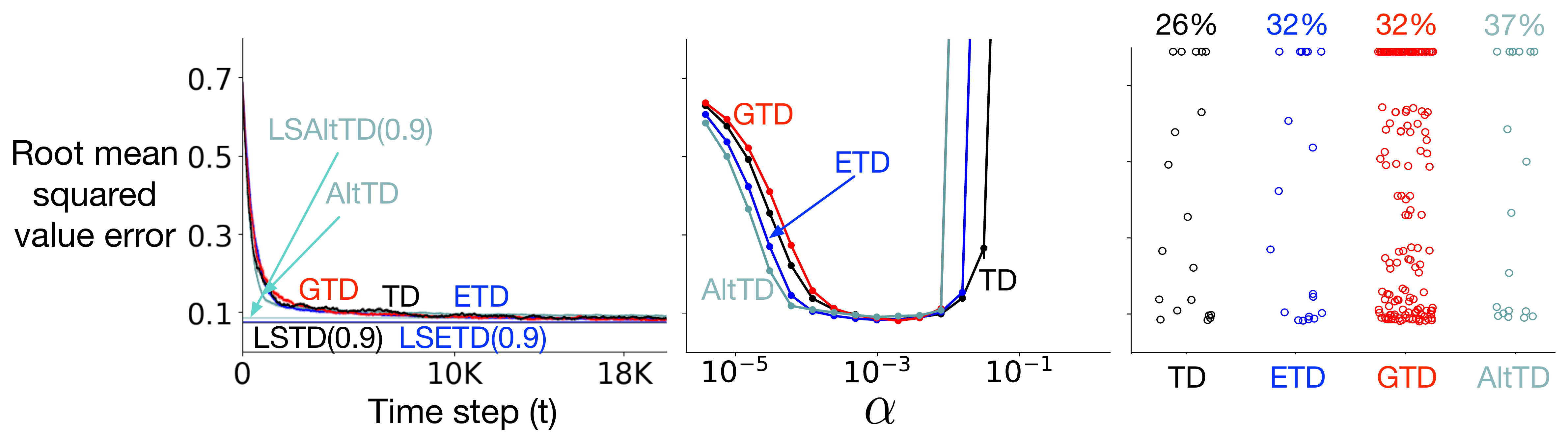}
      \caption{Comparing TD(0.9), GTD(0.9), ETD(0.9), and Alternative-life TD(0.9) on the {\bf Collision} problem. The free parameters in both the learning curve (left) and the stepsize plot (middle) are optimized for final performance. All methods performed similarly when high values of trace were used, though ETD(0.9) and TD(0.9) maintain less sensitivity to choice of $\alpha$ than GTD(0.9). Fewer parameter settings for TD(0.9) diverged than for the other methods and alternative life TD(0.9) had a much more bimodal density of well-performing parameter settings.}
      \label{fig:threemain_chicken_final_9}
\end{figure}

\begin{figure}[H]
      \centering
      \includegraphics[width=0.8\linewidth]{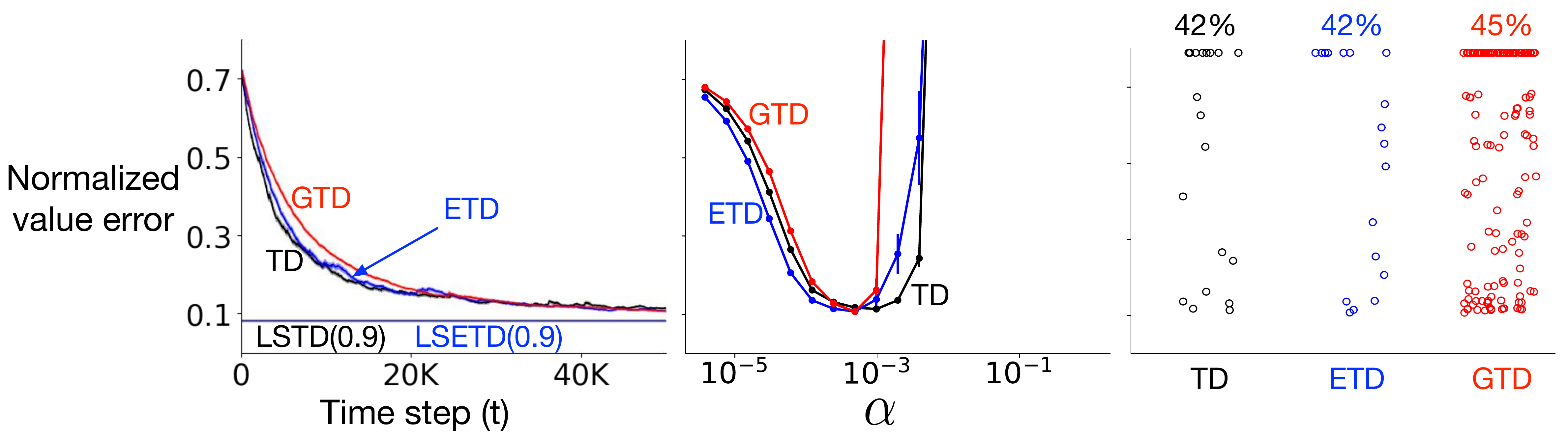}
      \caption{Comparing TD(0.9), GTD(0.9), and ETD(0.9) on the {\bf Four Rooms} problem. The free parameters in both the learning curve (left) and the stepsize plot (middle) are optimized for final performance. All methods performed similarly when high values of trace were used, though GTD(0.9) has a much more narrow stepsize sensitivity curve.}
      \label{fig:threemain_four_final_9}
\end{figure}

In Section \ref{sct:ComparingTheThreeMainFamilies} we compared prior and posterior correction methods mostly without use of eligibility traces.
Here we show additional results when traces are used ($\lambda=0.9$).
Figures \ref{fig:threemain_chicken_final_9} and \ref{fig:threemain_four_final_9} show comparisons between Off-policy TD($\lambda$), Gradient TD($\lambda$), and Emphatic TD($\lambda$) (Alternative-life TD($\lambda$) is also included in Figure~\ref{fig:threemain_chicken_final_9}).
When the trace parameter is $\lambda=0.9$, we notice little difference between the methods.
Notably, GTD(0.9) is more sensitive to choice of stepsize than TD(0.9) and ETD(0.9).

\section{Comparing accelerated Gradient-TD updates}
\label{app:ComparingGradientTDMethodsdWithProximalGradientTDMethods}

In Section \ref{sct:ComparingGradientTDMethodsdWithProximalGradientTDMethods} we showed the results of Proximal Gradient-TD methods compared with that of Gradient-TD methods.
Here we provide additional results that compare these methods on our two benchmark problems.

\begin{figure}[H]
      \centering{}
      \includegraphics[width=0.8\linewidth]{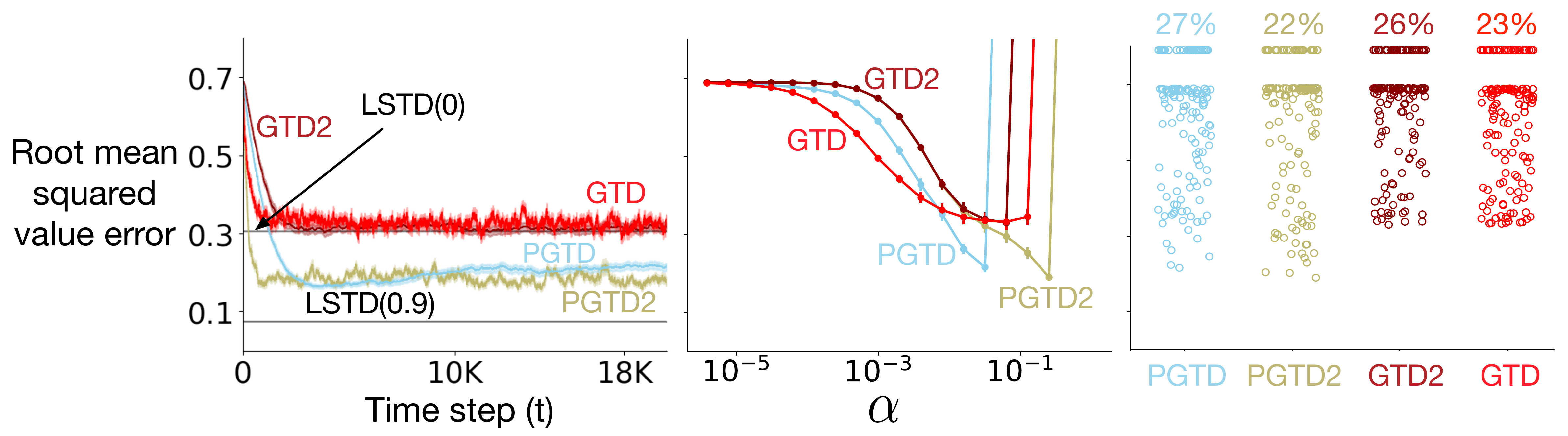}
      \caption{Comparing the learning speed of GTD(0), GTD2(0), Proximal-GTD(0), and Proximal-GTD2(0) on the {\bf Collision} problem. The free parameters for the learning curve (left) and stepsize sensitivity plot (middle) are optimized for AUC. The proximal methods outperformed GTD(0) and GTD2(0). The middle graph, however, shows that the proximal methods only outperform GTD(0) and GTD2(0) for a very specific range of stepsizes.}
      \label{appfig:CollisionAUCProximalVsGradientLambda0}
\end{figure}

\begin{figure}[H]
      \centering
      \includegraphics[width=0.8\linewidth]{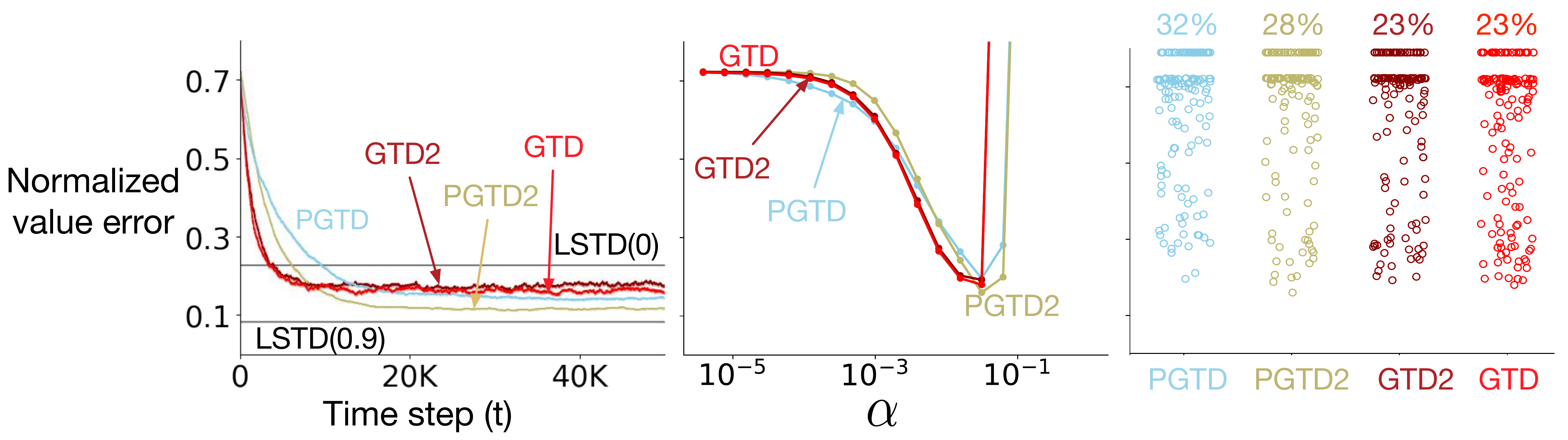}
      \caption{Comparing the learning speed of GTD(0), GTD2(0), Proximal-GTD(0), and Proximal-GTD2(0) on the {\bf Four Rooms} problem. The free parameters for the learning curve (left) and stepsize sensitivity plot (middle) are optimized for AUC. GTD(0) and GTD2(0) showed faster learning than Proximal-GTD(0) and Proximal-GTD2(0), however the proximal methods converged to a slightly lower error. }
      \label{appfig:FourRoomsAUCProximalVsGradientLambda0}
\end{figure}

When optimizing for AUC (when $\lambda=0$), in the collision problem all methods were similar in terms of learning speed (see Figure\ref{appfig:CollisionAUCProximalVsGradientLambda0}). In the Four Rooms problem, however, Gradient-TD methods learned slightly faster than Proximal Gradient-TD methods (see Figure~\ref{appfig:FourRoomsAUCProximalVsGradientLambda0}).

\begin{figure}[H]
      \centering
      \includegraphics[width=0.8\linewidth]{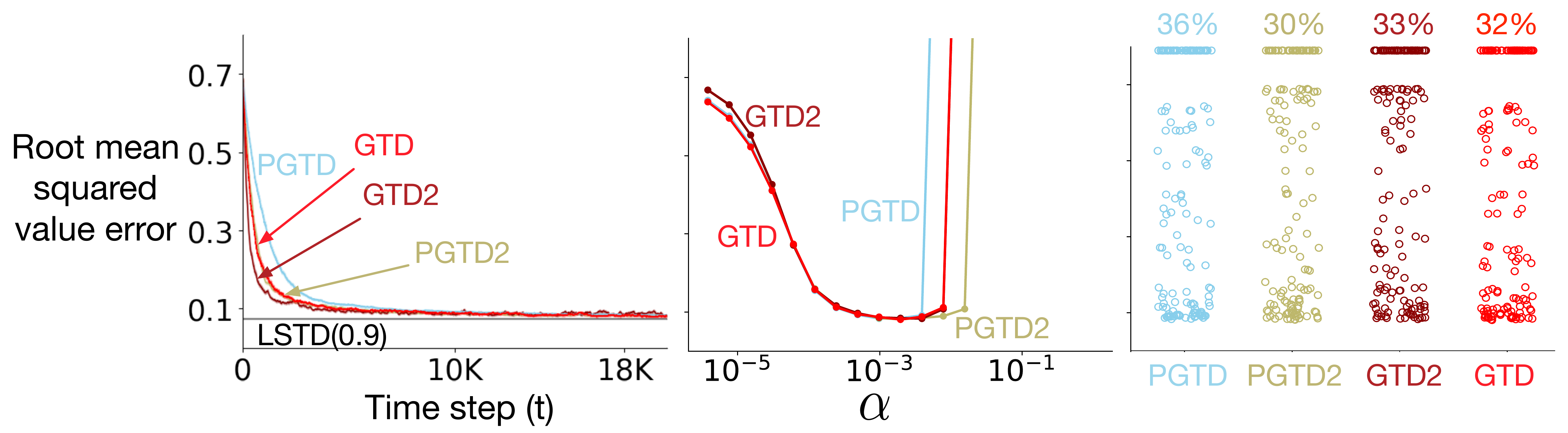}
      \caption{Comparing the final performance of GTD(0.9), GTD2(0.9), Proximal-GTD(0.9), and Proximal-GTD2(0.9) on the {\bf Collision} problem. The free parameters for the learning curve (left) and stepsize sensitivity plot (middle) are optimized for final performance. Again, for $\lambda=0.9$, all methods performed similarly in terms of speed and final performance.}
      \label{appfig:ProximalVsGradientLambda9FinalPerfCollision}
\end{figure}

\begin{figure}[H]
      \centering
      \includegraphics[width=0.8\linewidth]{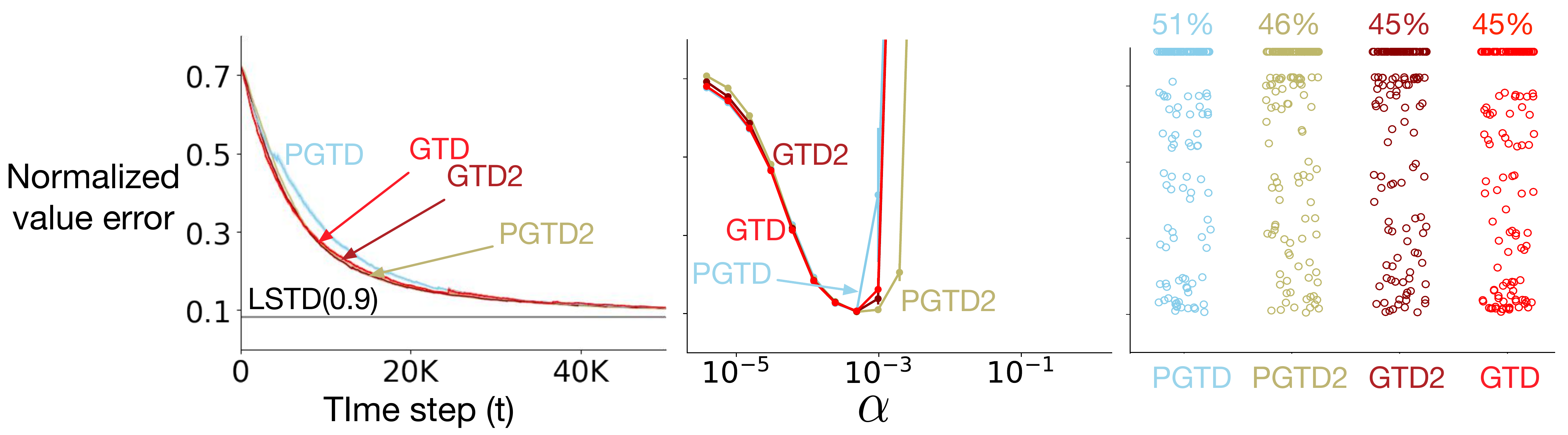}
      \caption{Comparing the final performance of GTD(0.9), GTD2(0.9), Proximal-GTD(0.9), and Proximal-GTD2(0.9) on the {\bf Four Rooms} problem. The free parameters for the learning curve (left) and stepsize sensitivity plot (middle) are optimized for final performance. Again, for $\lambda=0.9$, all methods performed similarly in terms of speed and final performance.}
      \label{appfig:ProximalVsGradientLambda9FinalPerfFourRooms}
\end{figure}

\begin{figure}[H]
      \centering
      \includegraphics[width=0.8\linewidth]{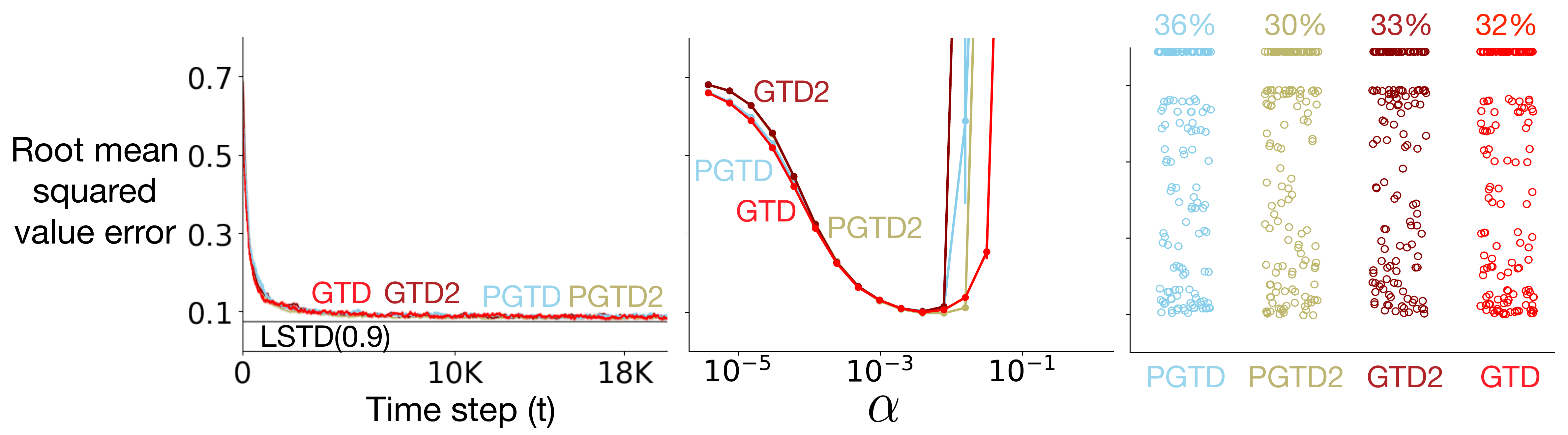}
      \caption{Comparing the learning speed of GTD(0.9), GTD2(0.9), Proximal-GTD(0.9), and Proximal-GTD2(0.9) on the {\bf Collision} problem. The free parameters for the learning curve (left) and stepsize sensitivity plot (middle) are optimized for AUC. All methods performed similarly in terms of speed and final performance.}
      \label{appfig:ProximalVsGradientLambda9AUCCollision}
\end{figure}

\begin{figure}[H]
      \centering
      \includegraphics[width=0.8\linewidth]{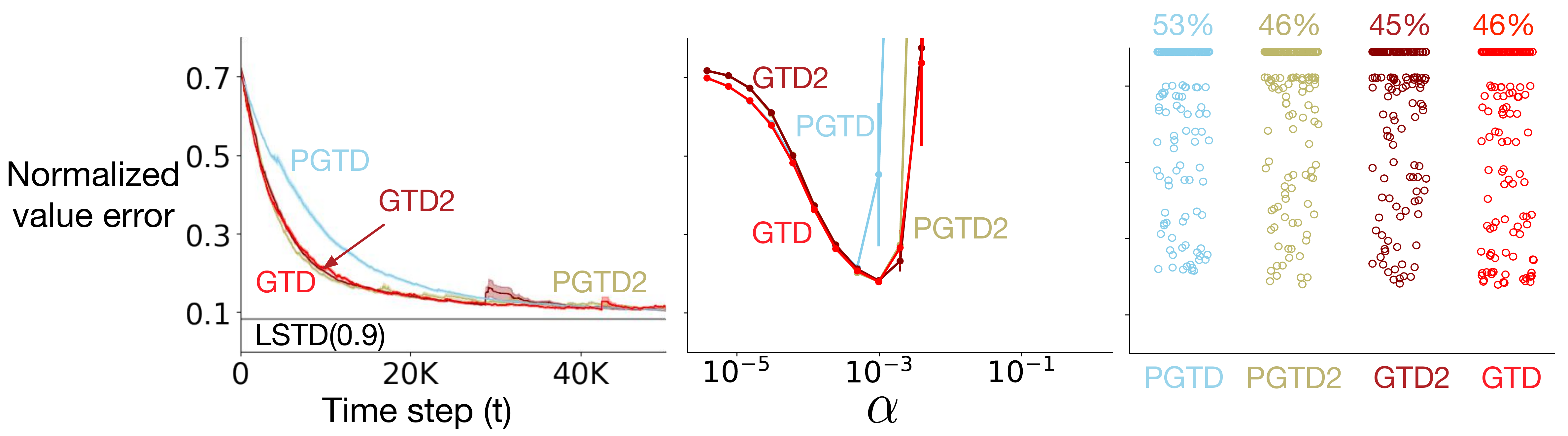}
      \caption{Comparing the learning speed of GTD(0.9), GTD2(0.9), Proximal-GTD(0.9), and Proximal-GTD2(0.9) on {\bf Four Rooms} problem. The free parameters for the learning curve (left) and stepsize sensitivity plot (middle) are optimized for AUC. Interestingly, Proximal-GTD(0.9) showed both slower learning and increased sensitivity to stepsize in this domain.}
      \label{appfig:ProximalVsGradientLambda9AUCFourRooms}
\end{figure}

When $\lambda=0.9$, all methods performed very similarly. However, Proximal-GTD(0.9) was slightly slower than the three other methods (GTD(0.9), GTD2(0.9), and Proximal-GTD2(0.9)). See Figures~\ref{appfig:ProximalVsGradientLambda9FinalPerfCollision}, \ref{appfig:ProximalVsGradientLambda9FinalPerfFourRooms}, \ref{appfig:ProximalVsGradientLambda9AUCCollision}, and \ref{appfig:ProximalVsGradientLambda9AUCFourRooms}. Proximal-GTD($\lambda$) showed more sensitivity to choice of learning rate than Proximal-GTD2($\lambda$), and had a larger percentage of parameter settings that diverged overall in all cases.

\section{Comparing Gradient-TD to hybrid updates}
\label{app:ComparingHybridTDMethodsWithTD}

In Section \ref{sct:ComparingHybridTDMethodsWithTD}, we investigated the effects of performing on-policy TD($\lambda$) updates whenever data is generated on-policy through using hybrid temporal difference learning.
Here we show additional results comparing Gradient-TD methods and Hybrid TD methods on our two benchmark domains.
We analyze the performance of each method with free parameters optimized for either final performance or area under the learning curve.

\begin{figure}[H]
      \centering
      \includegraphics[width=0.8\linewidth]{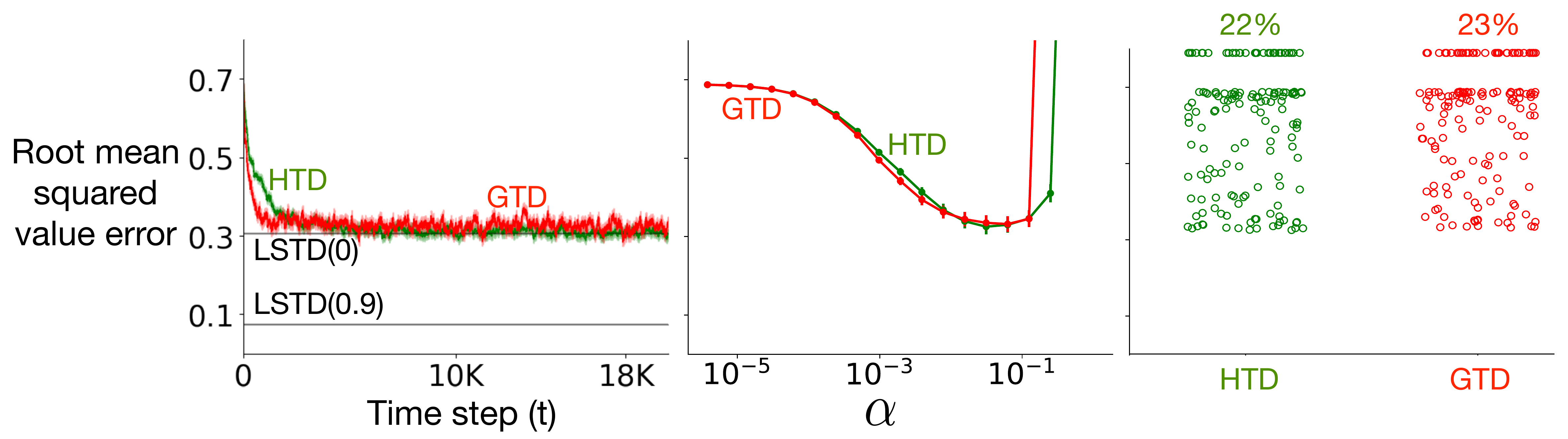}
      \caption{Comparing the learning speed of GTD(0) and HTD(0) on the {\bf Collision} problem. Free parameters are optimized for AUC. Both methods performed similarly and the parameter sensitivity plot (right) shows a similar distribution between methods.}
      \label{appfig:HTDCollisionLambda0AUC}
\end{figure}

\begin{figure}[H]
      \centering
      \includegraphics[width=0.8\linewidth]{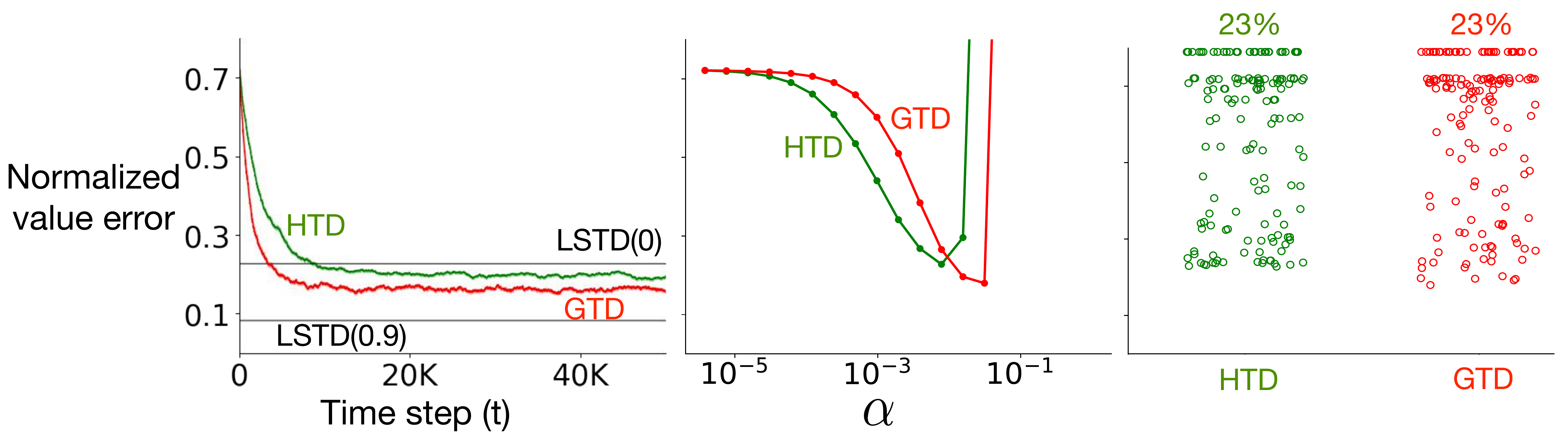}
      \caption{Comparing the learning speed of GTD(0) and HTD(0) on the {\bf Four Rooms} problem. Free parameters are optimized for AUC. GTD(0) outperformed HTD(0) in terms of speed and final performance; however, the right graph shows that GTD(0) outperforms HTD(0) only for specific parameter settings. For a majority of parameter settings, both methods have similar performance.}
      \label{appfig:HTDFourRoomsLambda0AUC}
\end{figure}

In the case where bootstrapping was complete($\lambda=0$ -- Figures~\ref{appfig:HTDCollisionLambda0AUC} and \ref{appfig:HTDFourRoomsLambda0AUC}) the methods performed very similarly. HTD(0) had a slightly larger range of stepsizes for which it converged. GTD(0) was, however, slightly faster than HTD(0) on both tasks. The percentage of parameter settings for which the methods diverged were similar.

\begin{figure}[H]
      \centering
      \includegraphics[width=0.8\linewidth]{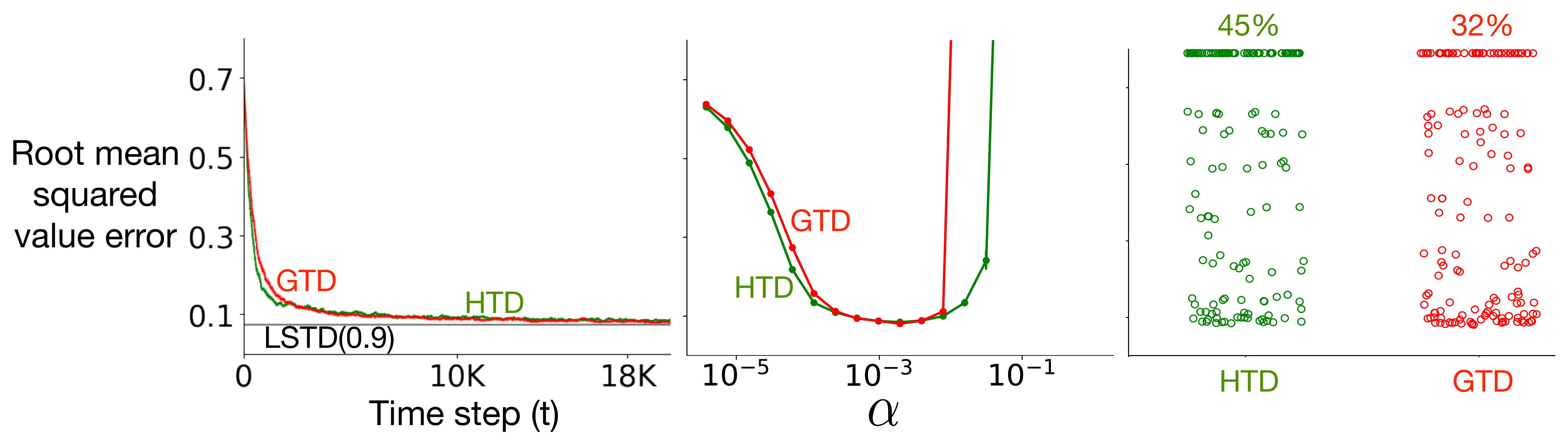}
      \caption{Comparing GTD(0.9) and HTD(0.9) on the {\bf Collision} problem. Free parameters are optimized for final performance. Both methods converge at approximately the same rate, and to the same final performance. HTD(0.9) has slightly less sensitivity to stepsize than GTD(0.9).}
      \label{appfig:htd_finalperf_9}
\end{figure}

\begin{figure}[H]
      \centering
      \includegraphics[width=0.8\linewidth]{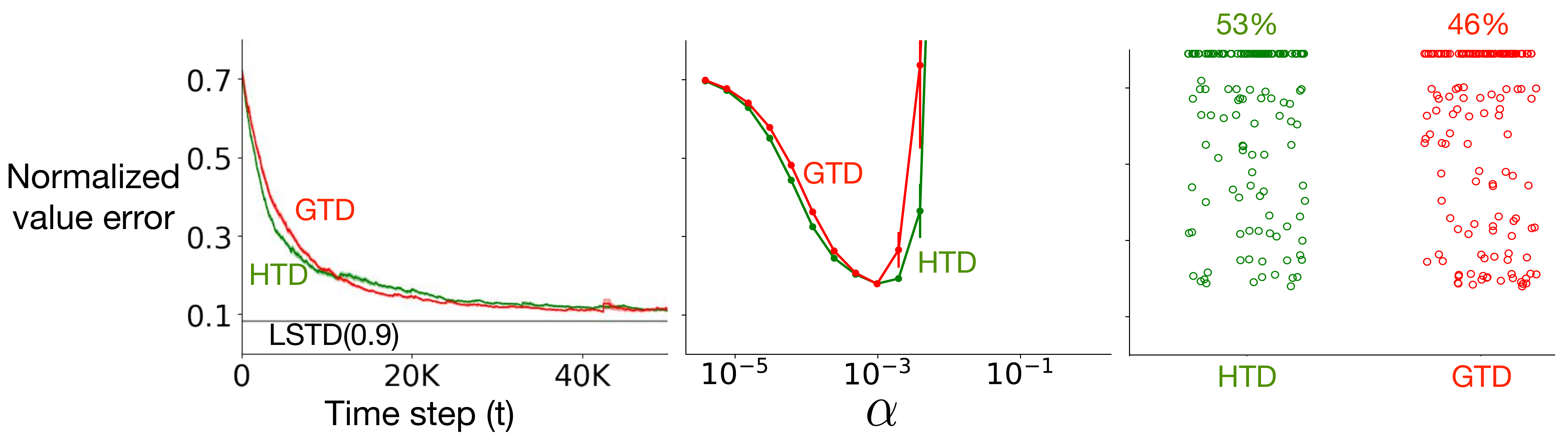}
      \caption{Comparing GTD(0.9) and HTD(0.9) on the {\bf Four Rooms} problem. Free parameters are optimized for area under the learning curve. Methods were similar in terms of speed and final performance. HTD(0.9) demonstrates a higher sensitivity to free parameter settings in the parameter sensitivity graph (right).}
      \label{appfig:htd_auc_9}
\end{figure}

In the case where traces are used ($\lambda = 0.9$ -- Figure \ref{appfig:htd_finalperf_9} through \ref{appfig:htd_auc_9}), HTD(0.9) has a slightly wider sensitivity curve for stepsize than GTD(0.9) implying less sensitivity to choice of $\alpha$; however, HTD(0.9) has a higher percentage of parameter settings that diverge overall.
This implies that once the stepsize is fixed, HTD(0.9) has a higher sensitivity to other free parameters than GTD(0.9).
Both methods consistently converge to the same final error under these settings.

\begin{figure}[H]
      \centering
      \includegraphics[width=0.8\linewidth]{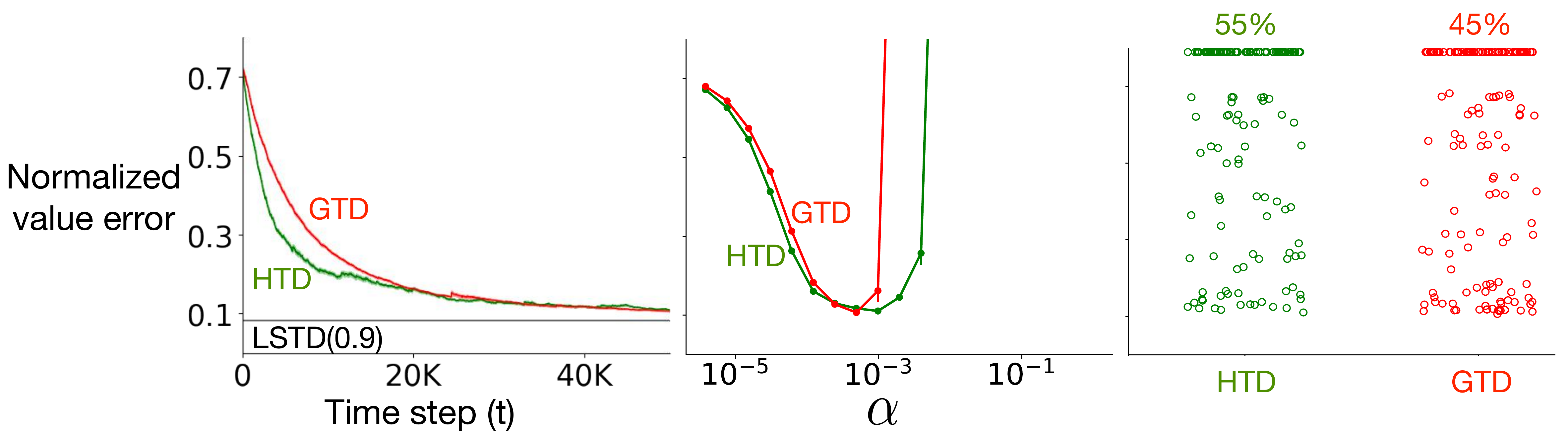}
      \caption{Comparing GTD(0.9) and HTD(0.9) on the {\bf Four Rooms} problem. Free parameters are optimized for final performance. HTD(0.9) converges to final performance a bit faster than GTD(0.9).}
      \label{fig:}
\end{figure}

\begin{figure}[H]
      \centering
      \includegraphics[width=0.8\linewidth]{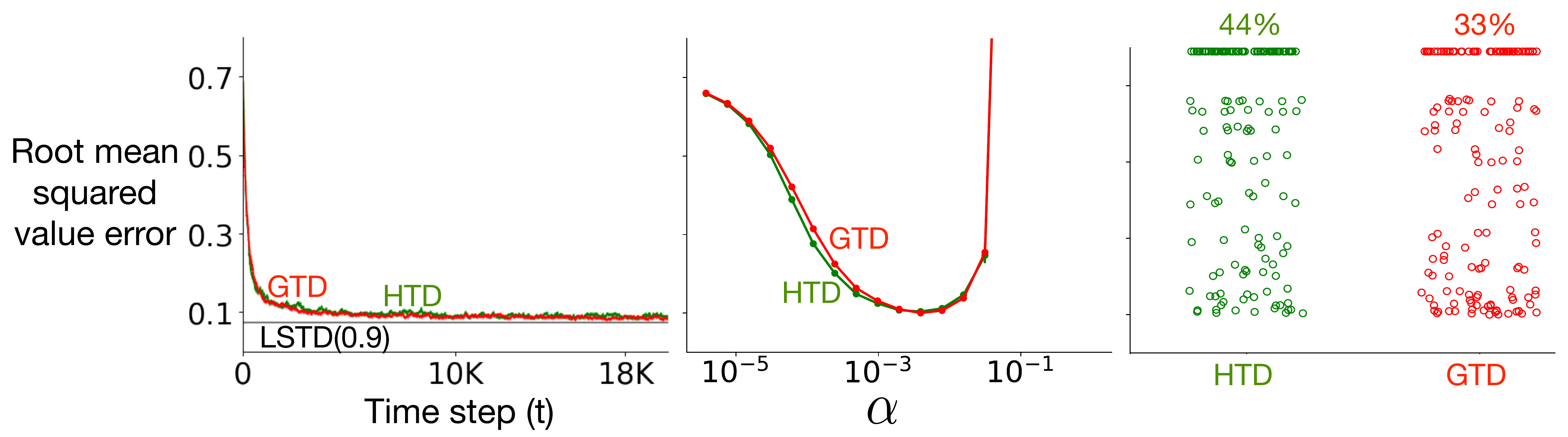}
      \caption{Comparing GTD(0.9) and HTD(0.9) on the {\bf Collision} problem. Free parameters are optimized for area under the learning curve. Methods were similar in terms of speed and final performance. HTD(0.9) demonstrates a higher sensitivity to free parameter settings in the parameter sensitivity graph (right).}
      \label{fig:}
\end{figure}

\section{Investigating variance reduction for emphatic updates}
\label{app:ComparingETDAndETDBeta}

In Section \ref{sct:ComparingETDAndETDBeta} we discussed the ETD($\lambda, \beta$), a method proposed to reduce the variance of the ETD($\lambda$) method.
ETD($\lambda, \beta$) introduces a mechanism for balancing between bias and variance.
We swept over values of $\beta$ and compared the difference between ETD($\lambda$) and ETD($\lambda, \beta$).
Specifically we swept over the following values of $\beta$: \{0.0, 0.2, 0.4, 0.6, 0.8, 1.0\}. To compare ETD($\lambda$) and ETD($\lambda, \beta$) methods, we assumed that ETD($\lambda, \beta$) cannot have its $\beta$ equal to the value of $\gamma$, in which case ETD($\lambda, \beta$) reduces to ETD($\lambda$).
As in other experiments, we run both methods on both benchmark domains, optimizing for either area under the learning curve or final performance, and with $\lambda=\{0, 0.9\}$.

Both ETD($\lambda$) and ETD($\lambda, \beta$) performed similarly across all domains and values of $\lambda$.
The most notable difference between these two algorithms is the percentage of parameter settings that diverge.
Because ETD($\lambda, \beta$) has an additional tunable parameter to reduce variance from the followon trace, there is a smaller percentage of parameter settings that diverge than for ETD($\lambda$).

\begin{figure}[H]
      \centering
      \includegraphics[width=0.8\linewidth]{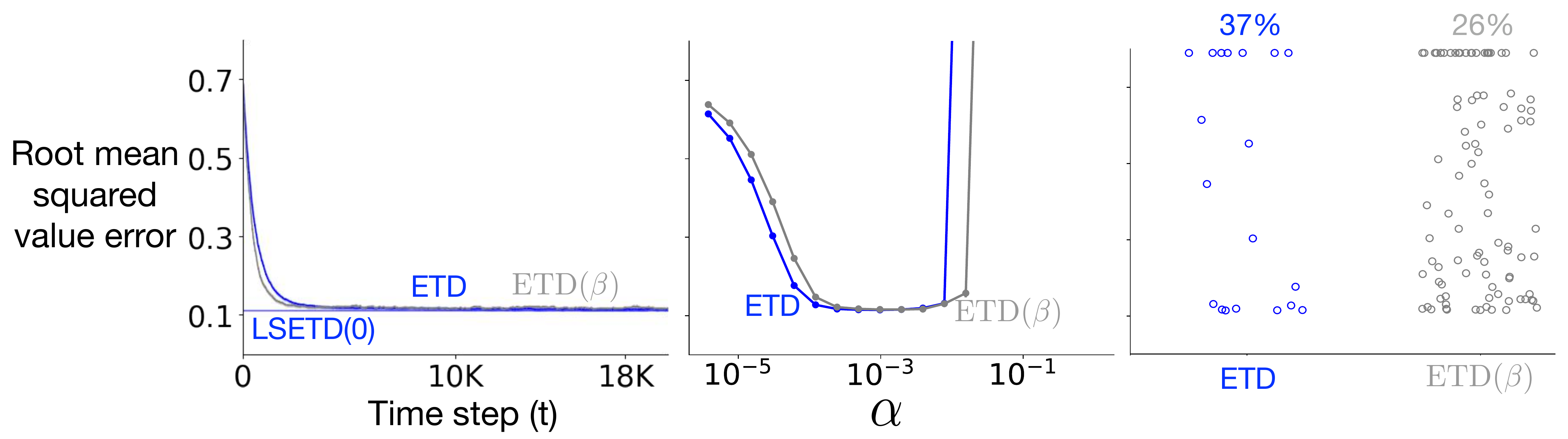}
      \caption{Comparing ETD(0) and ETD(0, $\beta$) on the {\bf Collision} problem. Free parameters are optimized for final performance. BMethods were similar in terms of speed and final performance. Notably, ETD(0) had a much larger percentage of parameter settings that diverged in this domain as compared to ETD(0, $\beta$).}
      \label{fig:}
\end{figure}

\begin{figure}[H]
      \centering
      \includegraphics[width=0.8\linewidth]{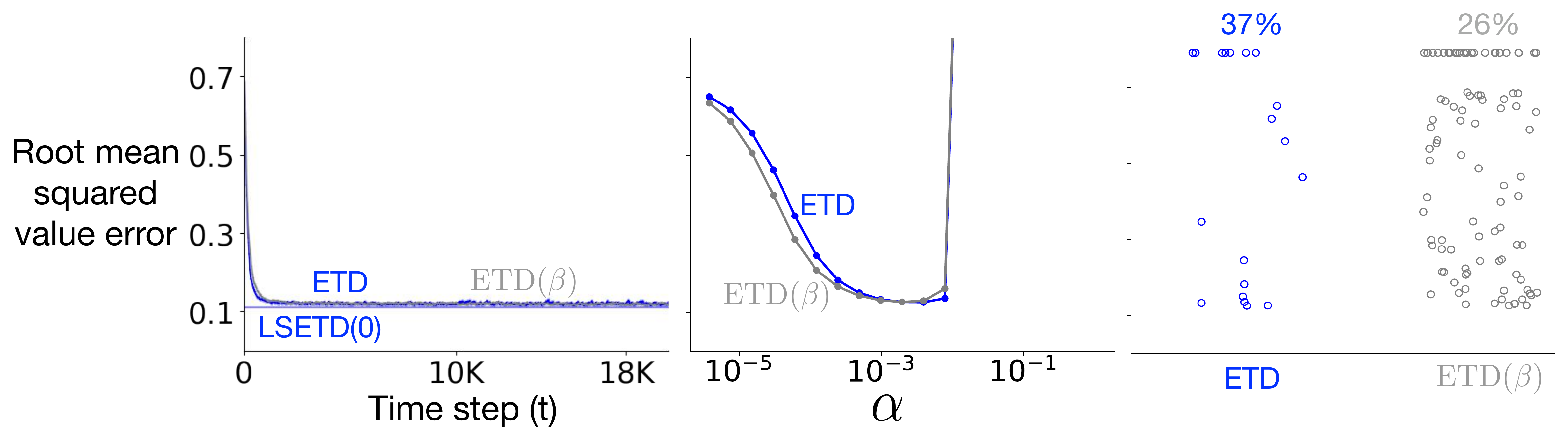}
      \caption{Comparing ETD(0) and ETD(0, $\beta$) on the {\bf Collision} problem. Free parameters are optimized for area under the learning curve. Methods were similar in terms of speed and final performance. Notably, ETD(0) had a much larger percentage of parameter settings that diverged in this domain as compared to ETD(0, $\beta$).}
      \label{fig:}
\end{figure}

\begin{figure}[H]
      \centering
      \includegraphics[width=0.8\linewidth]{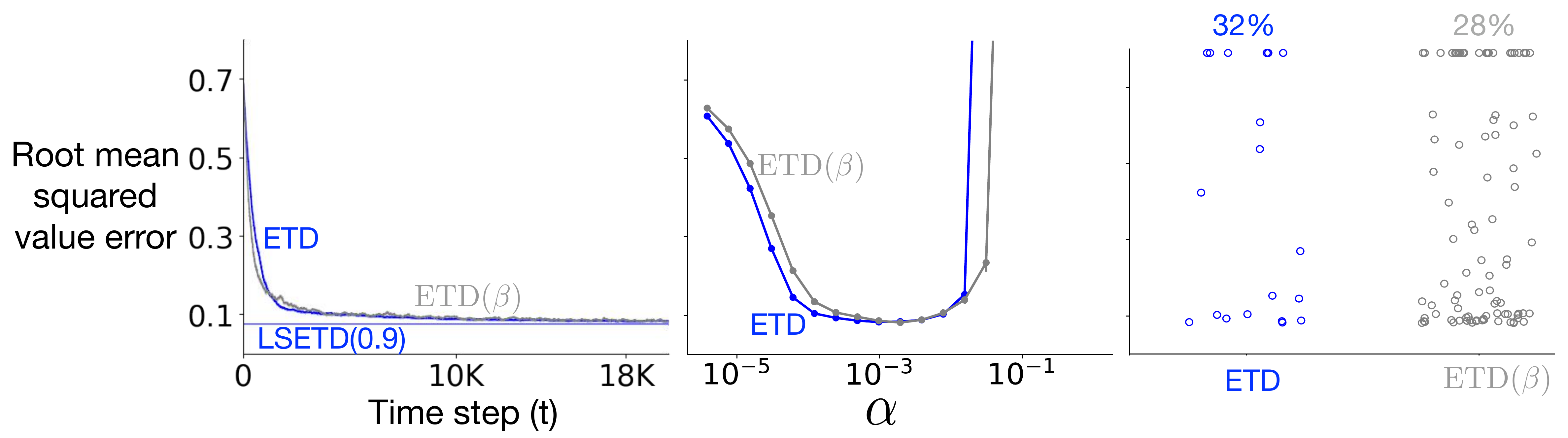}
      \caption{Comparing ETD(0.9) and ETD(0.9, $\beta$) on the {\bf Collision} problem. Free parameters are optimized for final performance. Methods were similar in terms of speed and final performance.}
      \label{fig:}
\end{figure}

\begin{figure}[H]
      \centering
      \includegraphics[width=0.8\linewidth]{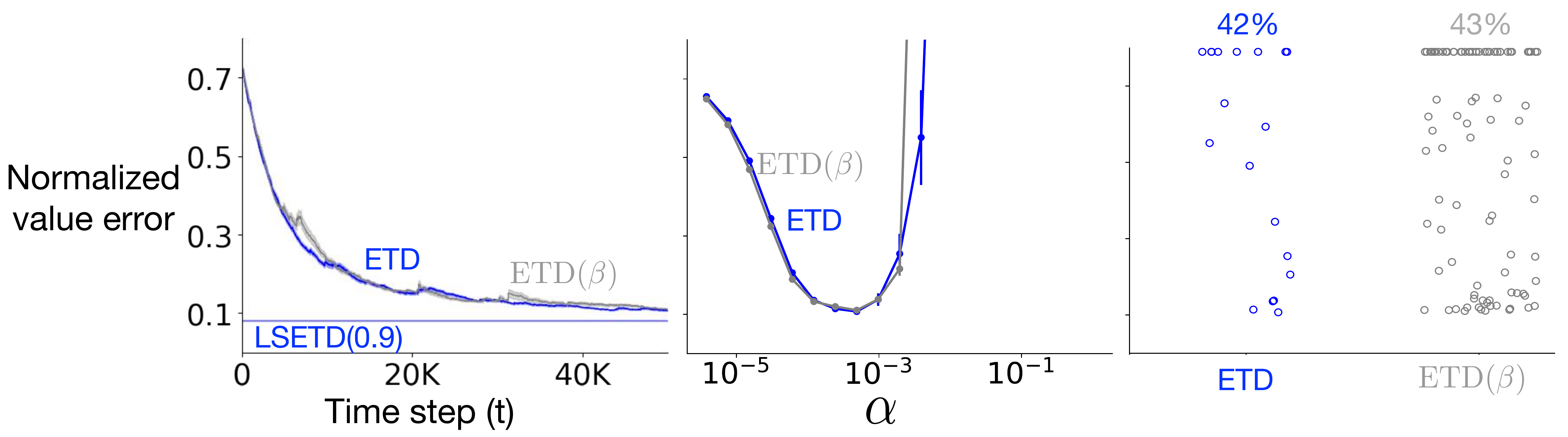}
      \caption{Comparing ETD(0.9) and ETD(0.9, $\beta$) on the {\bf Four Rooms} problem. Free parameters are optimized for final performance. Methods were similar in terms of speed and final performance. Neither method managed to attain the LSETD(0.9) fixed point error in this domain.}
      \label{fig:}
\end{figure}

\begin{figure}[H]
      \centering
      \includegraphics[width=0.8\linewidth]{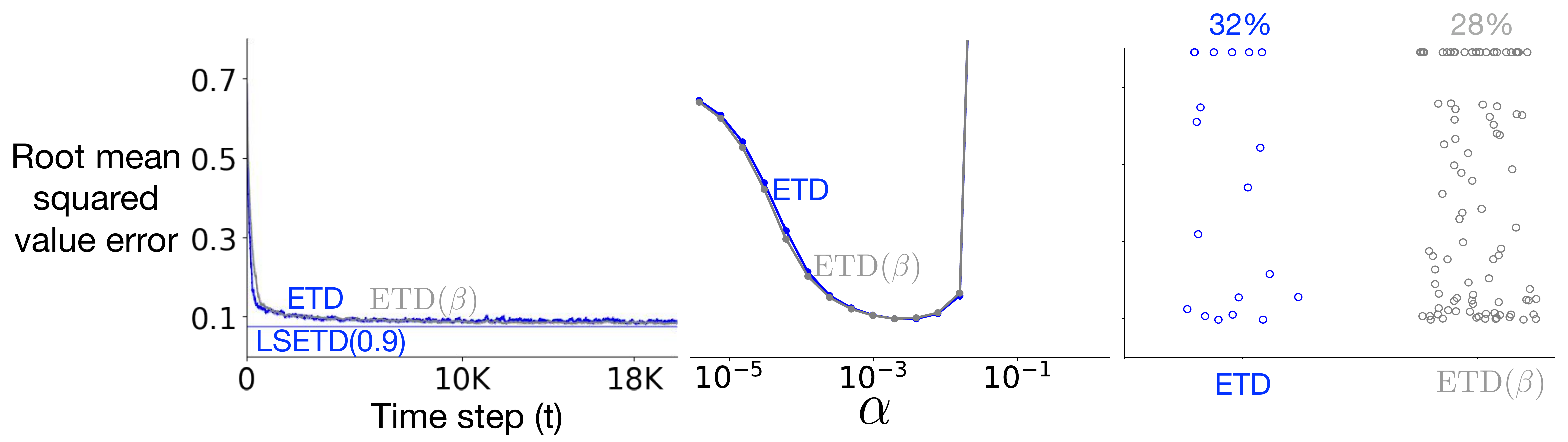}
      \caption{Comparing ETD(0.9) and ETD(0.9, $\beta$) on the {\bf Collision} problem. Free parameters are optimized for area under the learning curve. Methods were similar in terms of speed and final performance, as well as sensitivity to choice of parameters.}
      \label{fig:}
\end{figure}

\begin{figure}[H]
      \centering
      \includegraphics[width=0.8\linewidth]{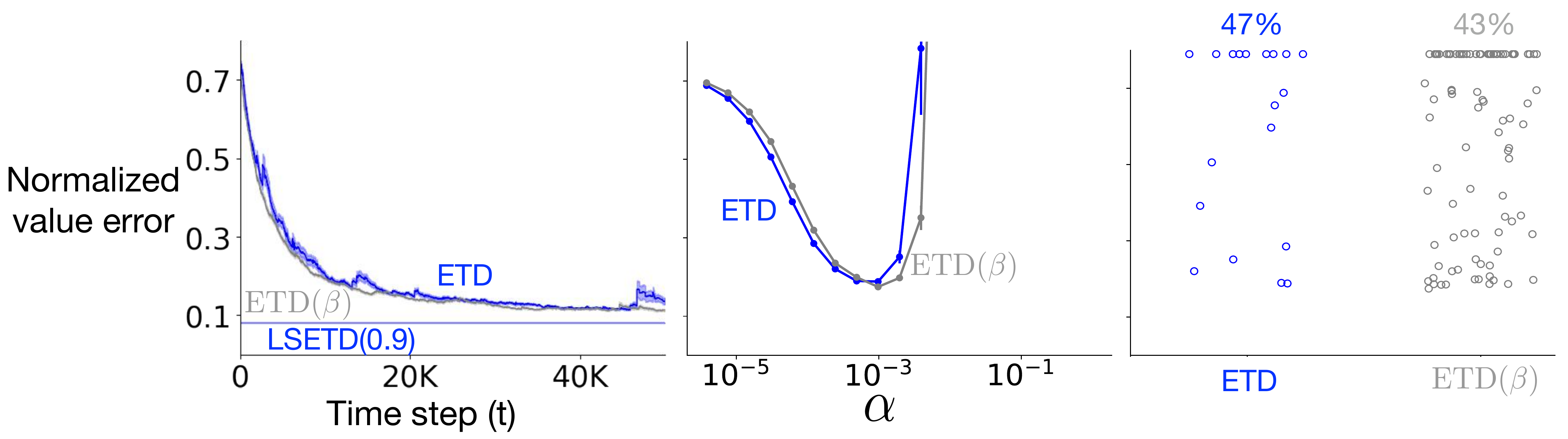}
      \caption{Comparing ETD(0.9) and ETD(0.9, $\beta$) on the {\bf Four Rooms} problem. Free parameters are optimized for area under the learning curve. Methods were similar in terms of speed and final performance. Neither method managed to attain the LSETD(0.9) fixed point error in this domain.}
      \label{fig:}
\end{figure}

\section{Comparing off-policy methods on high variance domain}
\label{app:ComparingTDWithABTDV-traceAndTreeBackup}

In this section, we additionally include all previous performance comparisons from Sections \ref{sct:ComparingTheThreeMainFamilies} through \ref{sct:ComparingETDAndETDBeta} on the high variance domain described in Section \ref{sct:ComparingTDWithABTDV-traceAndTreeBackup}. For clarity, we separate these results on the high variance setting from the other results on the two initial testbeds.

The conclusion that can be made is that methods proposed for reducing variance in off-policy learning (V-trace($\lambda$), Tree Backup($\lambda$), and ABTD($\zeta$)) are well-suited for problems that might have high variance. These methods effectively reduce the value of $\lambda$ in TD($\lambda$) to control the variance. This is why the performance of these methods when $\lambda=0$ and when $\lambda=0.9$ is similar.

\begin{figure}[H]
      \centering
      \includegraphics[width=0.8\linewidth]{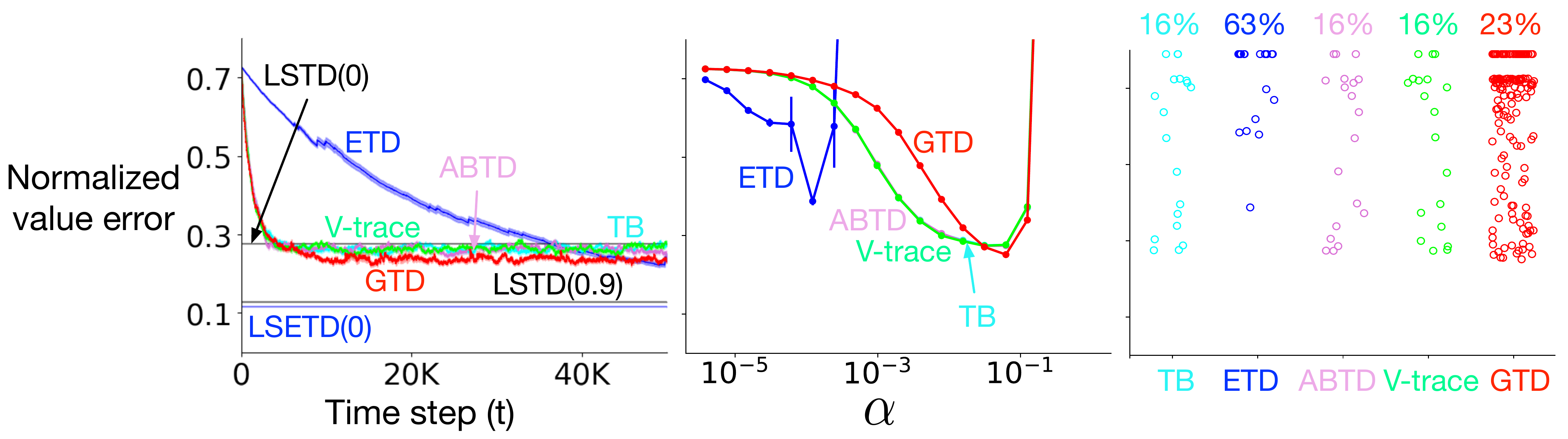}
      \caption{Comparing ETD(0), GTD(0), ABTD(0), TB(0), and V-trace(0) on the {\bf High Variance Four Rooms} problem. Free parameters are optimized for area under the learning curve. ABTD(0), V-trace(0), Tree Backup(0) and GTD(0) outperformed ETD(0). ETD(0) was the slowest method to converge to its final performance. The middle figure shows that ETD(0) was very sensitive to the choice of stepsize. GTD(0) is less sensitive than ETD(0) but more sensitive than ABTD(0), Tree Backup(0) and V-trace(0).}
      \label{fig:}
\end{figure}

\begin{figure}[H]
      \centering
      \includegraphics[width=0.8\linewidth]{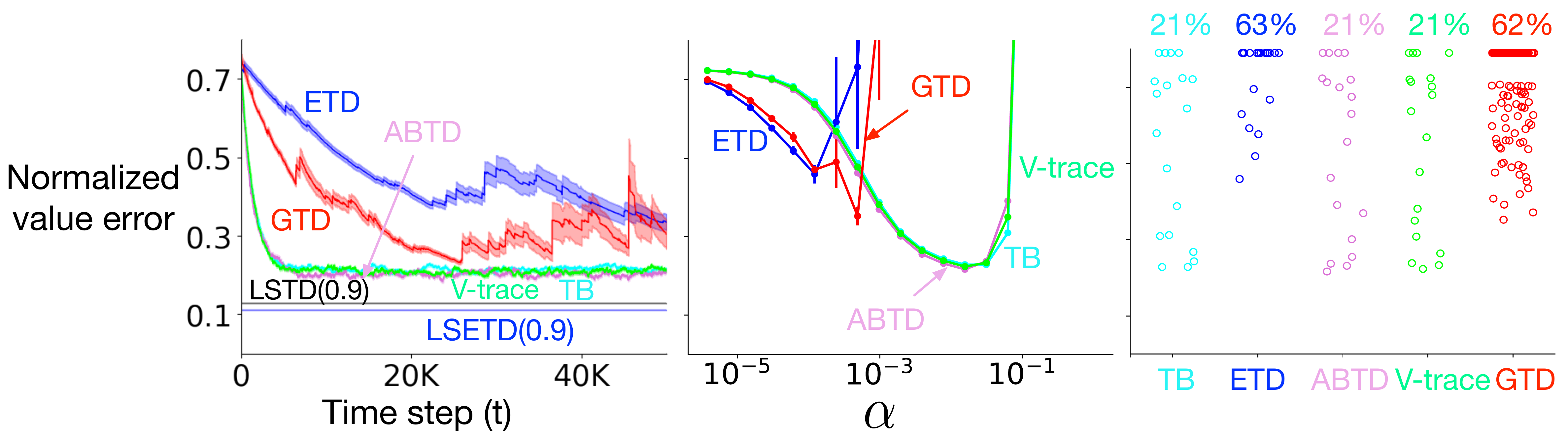}
      \caption{Comparing ETD(0.9), GTD(0.9), ABTD(0.9), TB(0.9), and V-trace(0.9) on the {\bf High Variance Four Rooms} problem. Free parameters are optimized for area under the learning curve. ETD(0.9) and GTD(0.9) both had trouble converging to a good final performance. They were also significantly slower than the other methods. Tree Backup(0.9), ABTD(0.9) and V-trace(0.9) managed to achieve a similar performance as when $\lambda$ was set to 0.}
      \label{fig:}
\end{figure}

\subsection{Comparing prior and posterior correction methods on high variance domain}

In Section \ref{sct:ComparingTheThreeMainFamilies} we compared prior and posterior correction methods on our benchmark domains.
We presented all additional results on these domains in Appendix \ref{app:ComparingTheThreeMainFamilies}.
Here we provide results following the same experimental procedure on the high variance domain.

ETD($\lambda$) has much slower learning than either TD($\lambda$) or GTD($\lambda$) on this domain.
Additionally, ETD($\lambda$) has a much more narrow sensitivity to stepsize and has notably more parameter settings that diverge than the other two algorithms.
TD($\lambda$) has a wide sensitivity to stepsize and has far fewer parameter settings that diverge, but comes to a worse final performance than either ETD($\lambda$) or GTD($\lambda$).

\begin{figure}[H]
      \centering
      \includegraphics[width=0.8\linewidth]{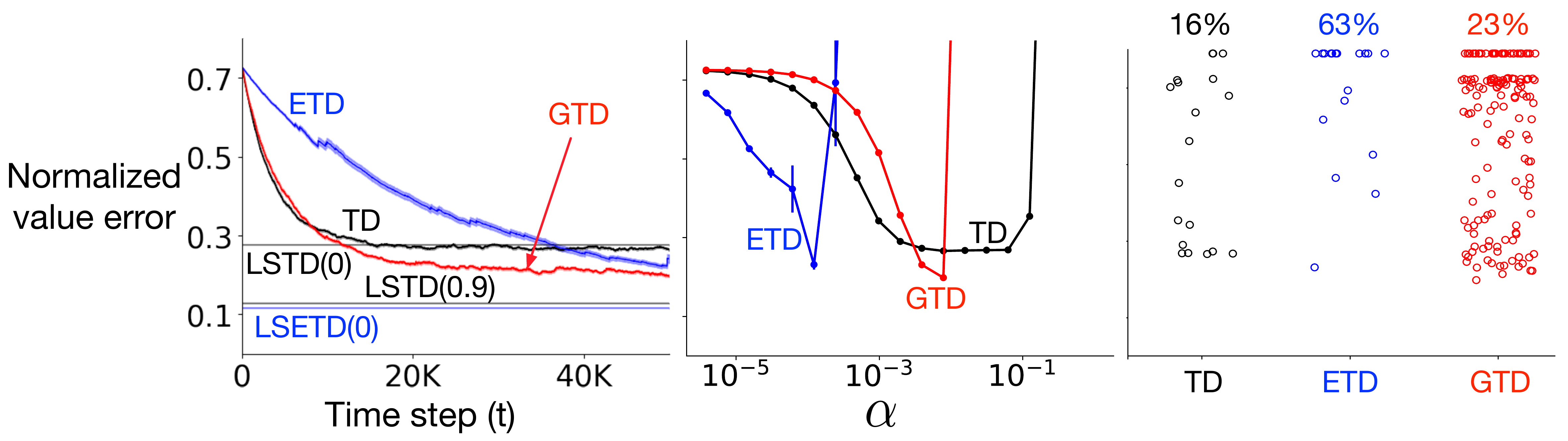}
      \caption{Comparing TD(0), GTD(0), and ETD(0) on the {\bf High Variance Four Rooms} problem. Free parameters are optimized for final performance. TD(0) and GTD(0) outperformed ETD(0) in terms of speed. ETD(0) slowly converged to its best final performance. GTD(0) converged to a better final performance than TD(0) but the parameter sensitivity graph (middle) shows that it is hard to find the parameter set for which GTD(0) converges to a better final performance. However, it is easy to find a parameter set with which GTD(0) converges to a performance similar to that of TD(0).}
      \label{fig:}
\end{figure}

\begin{figure}[H]
      \centering
      \includegraphics[width=0.8\linewidth]{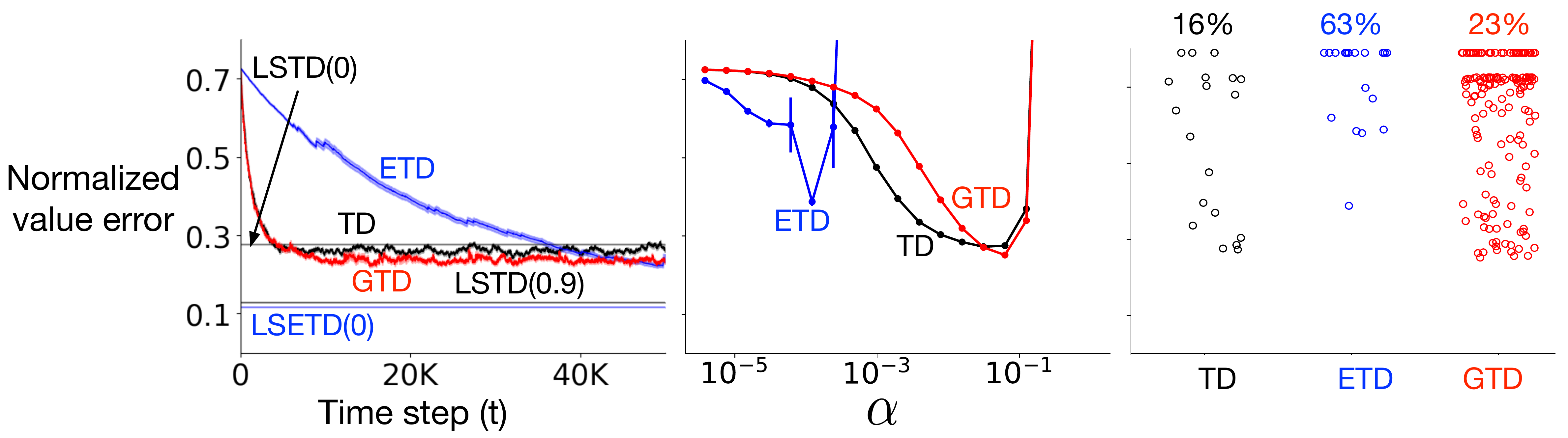}
      \caption{Comparing the learning speed of TD(0), GTD(0), and ETD(0) on the {\bf High Variance Four Rooms} problem. Free parameters are optimized for area under the learning curve. TD(0) and GTD(0) were similar and they both outperformed ETD(0).}
      \label{fig:}
\end{figure}

\begin{figure}[H]
      \centering
      \includegraphics[width=0.8\linewidth]{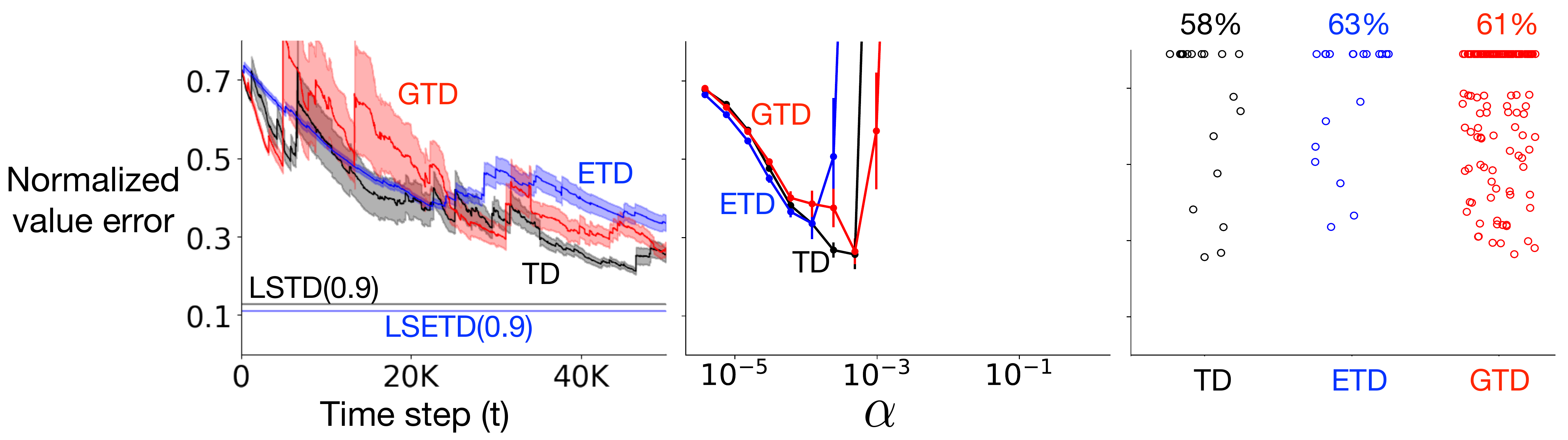}
      \caption{Comparing the final performance of TD(0.9), GTD(0.9), and ETD(0.9) on the {\bf High Variance Four Rooms} problem. Free parameters are optimized for final performance. All methods had trouble converging to their best final performance. All methods were high variance during learning.}
      \label{fig:}
\end{figure}

\begin{figure}[H]
      \centering
      \includegraphics[width=0.8\linewidth]{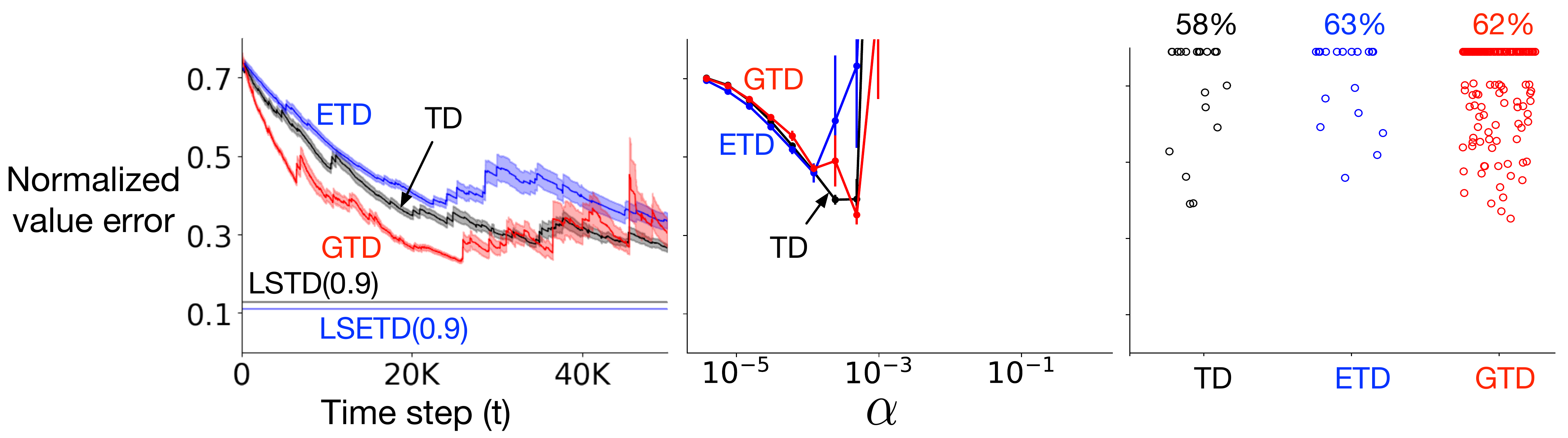}
      \caption{Comparing the learning speed of TD(0.9), GTD(0.9), and ETD(0.9) on the {\bf High Variance Four Rooms} problem. Free parameters are optimized for area under the learning curve. All methods had trouble converging to their best final performance (LSTD(0.9) or LSETD(0.9)). All methods were high variance during learning.}
      \label{fig:}
\end{figure}

\subsection{Comparing accelerated Gradient-TD updates on high variance domain}
Previously, we compared Proximal Gradient-TD methods with Gradient-TD methods in Section \ref{sct:ComparingGradientTDMethodsdWithProximalGradientTDMethods}.
In Appendix \ref{app:ComparingGradientTDMethodsdWithProximalGradientTDMethods} we showed additional results gathered on our two benchmark domains.
Here we provide a comparison between the Proximal Gradient-TD methods and the Gradient-TD methods on the High Variance Four Rooms problem.

\begin{figure}[H]
      \centering
      \includegraphics[width=0.8\linewidth]{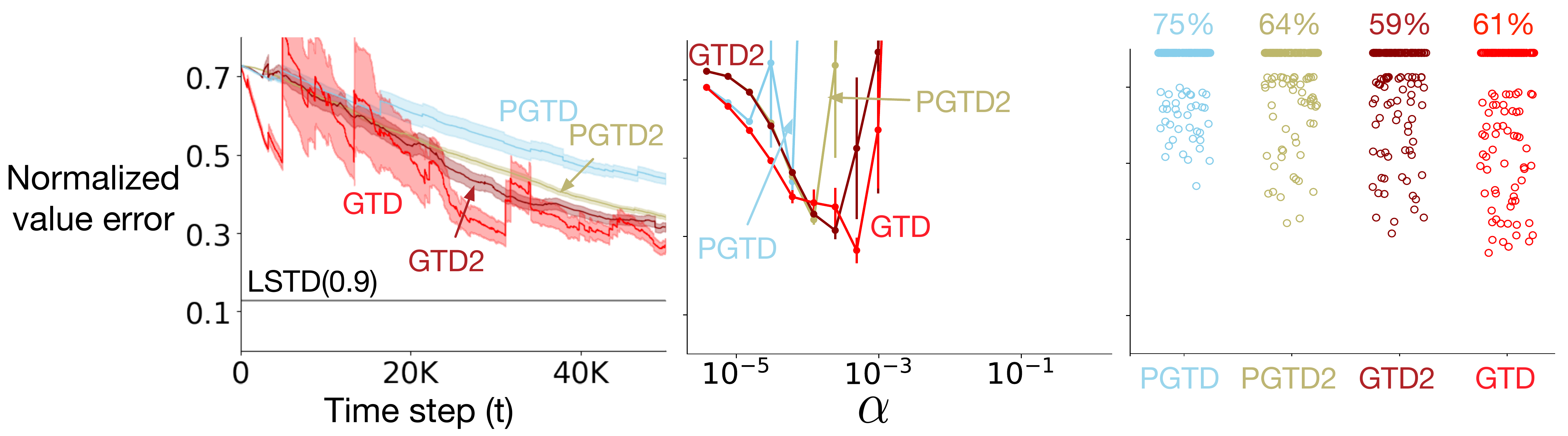}
      \caption{Comparing Proximal-GTD(0.9), Proximal-GTD2(0.9), GTD(0.9), and GTD2(0.9) on the {\bf High Variance Four Rooms} problem. Free parameters are optimized for final performance. Overall, methods performed similarly; however, GTD(0.9) and GTD2(0.9) slightly outperformed proximal-gradient methods (see the parameter sensitivity graph on the right).}
      \label{appfig:ProximalVsGradientHighFourLambda9FinalPerf}
\end{figure}

\begin{figure}[H]
      \centering
      \includegraphics[width=0.8\linewidth]{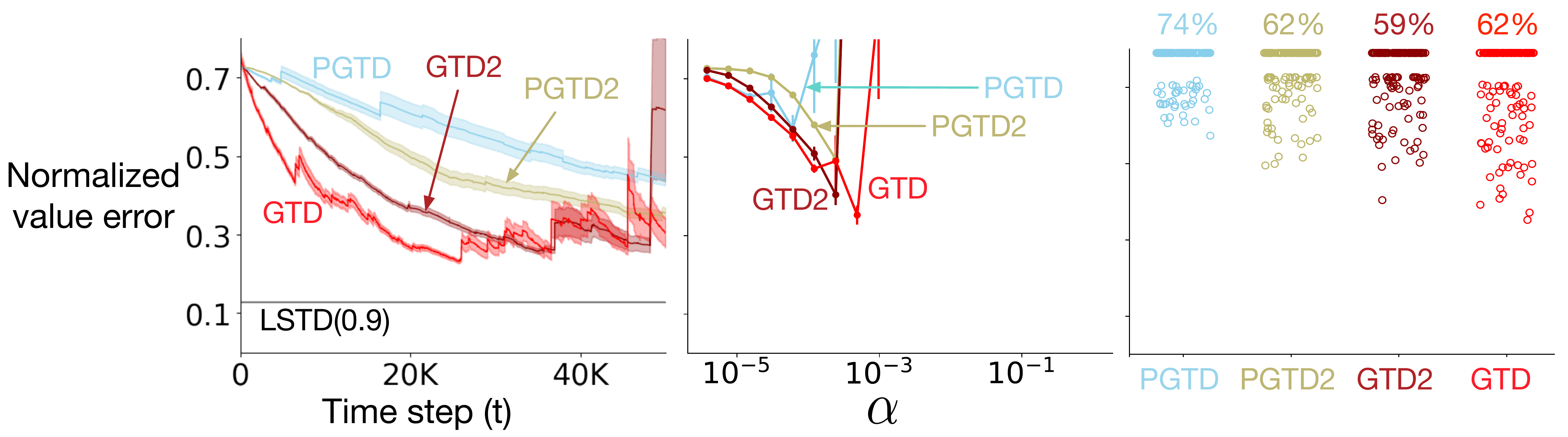}
      \caption{Comparing Proximal-GTD(0.9), Proximal-GTD2(0.9), GTD(0.9), and GTD2(0.9) on the {\bf High Variance Four Rooms} problem. Free parameters are optimized for area under the learning curve. All methods had high variance when learning. The parameter sensitivity graph shows that GTD(0.9) outperformed other methods, followed by GTD2(0.9), Proximal-GTD2(0.9), and Proximal-GTD(0.9).}
      \label{appfig:ProximalVsGradientHighFourLambda9AUC}
\end{figure}

Overall, Gradient-TD methods performed significantly better than Proximal Gradient-TD methods. When using traces, we found that Proximal Gradient-TD methods perform more poorly than Gradient-TD methods in terms of final performance and speed. Proximal Gradient-TD methods also had more parameter settings for which the methods diverged (the percentage of diverged parameter settings is shown in Figures \ref{appfig:ProximalVsGradientHighFourLambda9FinalPerf} and \ref{appfig:ProximalVsGradientHighFourLambda9AUC} on top of the parameter sensitivity graph).
The proximal methods also exhibit more sensitivity to choice of stepsize when $\lambda=0.9$.
Additionally, proximal methods converge to higher fixed-point errors than gradient methods.
When traces are not used, Proximal Gradient-TD methods are slower than Gradient-TD methods and they still diverge for more parameter settings compared to Gradient-TD methods.

\begin{figure}[H]
      \centering
      \includegraphics[width=0.8\linewidth]{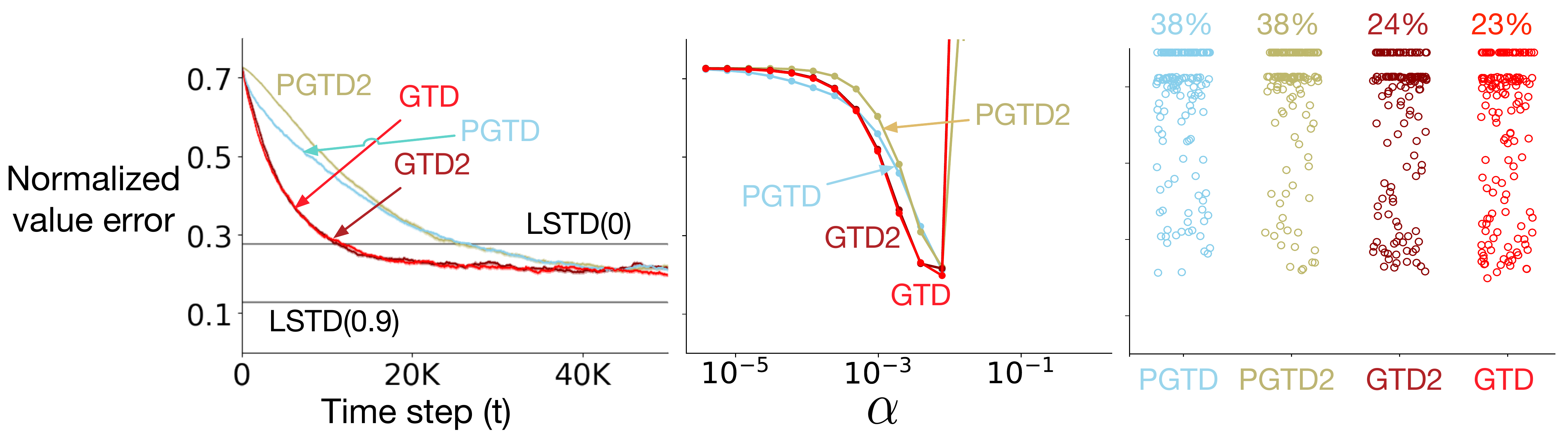}
      \caption{Comparing Proximal-GTD(0), Proximal-GTD2(0), GTD(0), and GTD2(0) on the {\bf High Variance Four Rooms} problem. Free parameters are optimized for final performance. Proximal methods were significantly slower than GTD(0) and GTD2(0). The final performance, however, was similar between all methods. The parameter sensitivity graph (right) shows that GTD(0) and GTD2(0) converged to a point close to their final performance for more parameters than proximal methods (more points are gathered around the bottom of the figure for GTD(0) and GTD2(0)).}
      \label{fig:}
\end{figure}

\begin{figure}[H]
      \centering
      \includegraphics[width=0.8\linewidth]{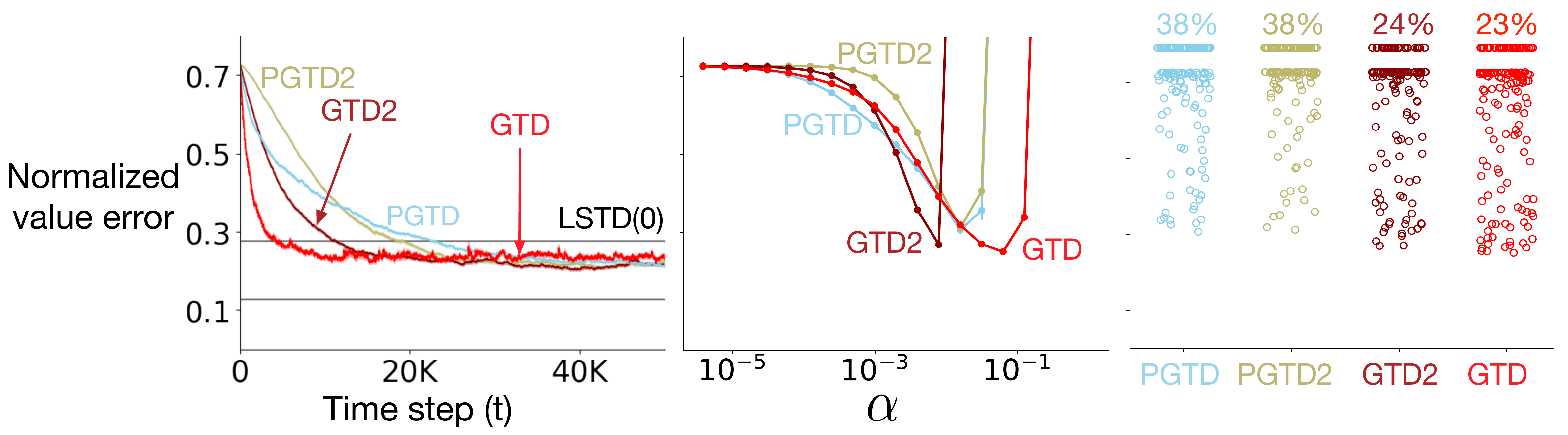}
      \caption{Comparing Proximal-GTD(0), Proximal-GTD2(0), GTD(0), and GTD2(0) on the {\bf High Variance Four Rooms} problem. Free parameters are optimized for area under the learning curve. GTD(0) was the fastest method. Overall, GTD(0) and GTD2(0) seem to be superior to the other two methods since they are faster, it is easier to find the parameter set for which they converge to their best final performance (see all the points gathered around the bottom of the figure for GTD2(0) and GTD(0)).}
      \label{fig:}
\end{figure}

\subsection[dummyLabel]{Comparing Gradient-TD($\lambda$) to Hybrid TD($\lambda$) on high variance domain}
In Section \ref{sct:ComparingHybridTDMethodsWithTD} we compared Hybrid TD($\lambda$) to Gradient-TD($\lambda$).
In Appendix \ref{app:ComparingHybridTDMethodsWithTD} we provide all additional results comparing these methods on our two benchmark domains.
Here we show a comparison of HTD($\lambda$) and GTD($\lambda$) on the high variance domain from Section \ref{sct:ComparingTDWithABTDV-traceAndTreeBackup}.
Both GTD($\lambda$) and HTD($\lambda$) show similar performance on the high variance domain.
HTD($\lambda$) is slightly more sensitive to choice of stepsize parameter on this domain when $\lambda=0.9$. The methods successfully solved the task when $\lambda=0.0$ but showed high variance when the bootstrapping parameter was set to 0.9.

\begin{figure}[H]
      \centering
      \includegraphics[width=0.8\linewidth]{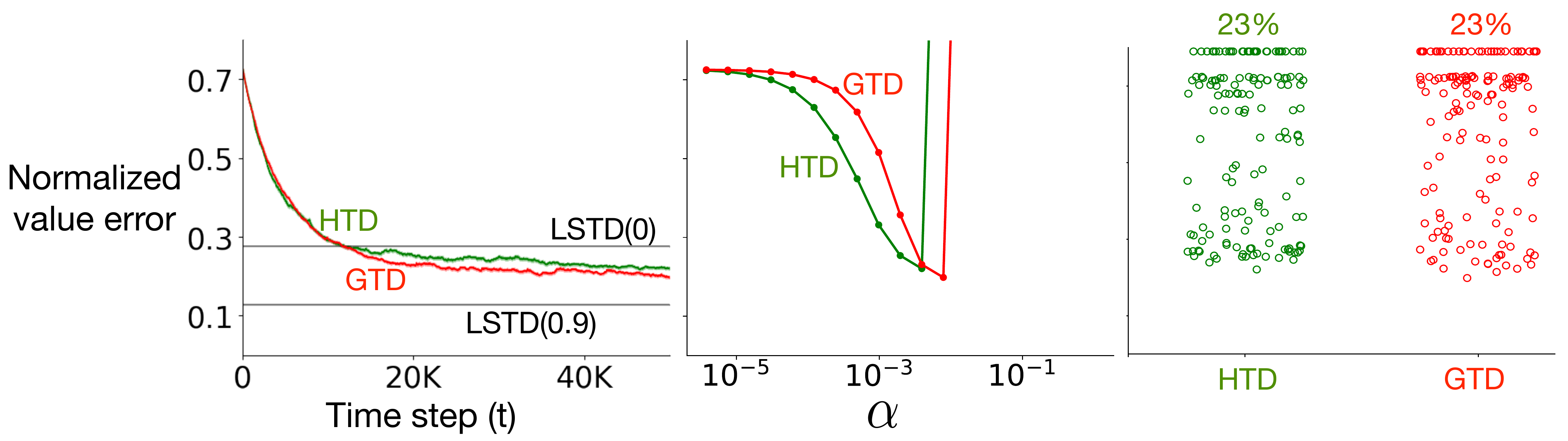}
      \caption{Comparing GTD(0), HTD(0) on the {\bf High Variance Four Rooms} problem. Free parameters are optimized for final performance. Methods performed similarly in terms of speed and final performance. GTD(0) slightly outperformed HTD(0) in terms of final performance for a very specific parameter set. The parameter sensitivity graph however, shows that methods achieved a similar overall performance over different parameters.}
      \label{fig:}
\end{figure}

\begin{figure}[H]
      \centering
      \includegraphics[width=0.8\linewidth]{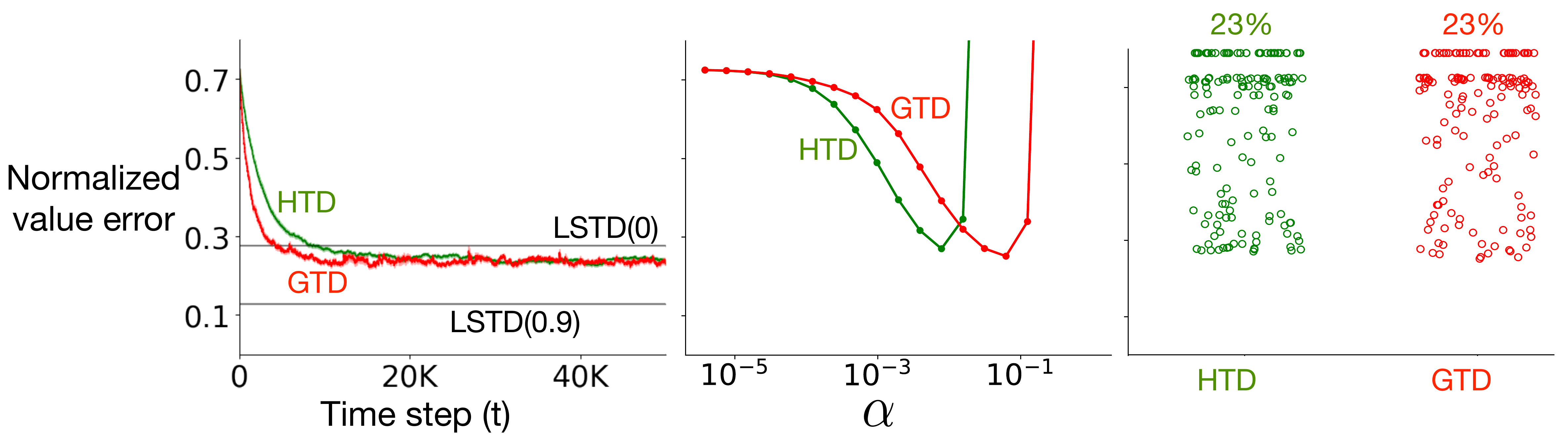}
      \caption{Comparing GTD(0), HTD(0) on the {\bf High Variance Four Rooms} problem. Free parameters are optimized for area under the learning curve. Methods performed similarly.}
      \label{fig:}
\end{figure}

\begin{figure}[H]
      \centering
      \includegraphics[width=0.8\linewidth]{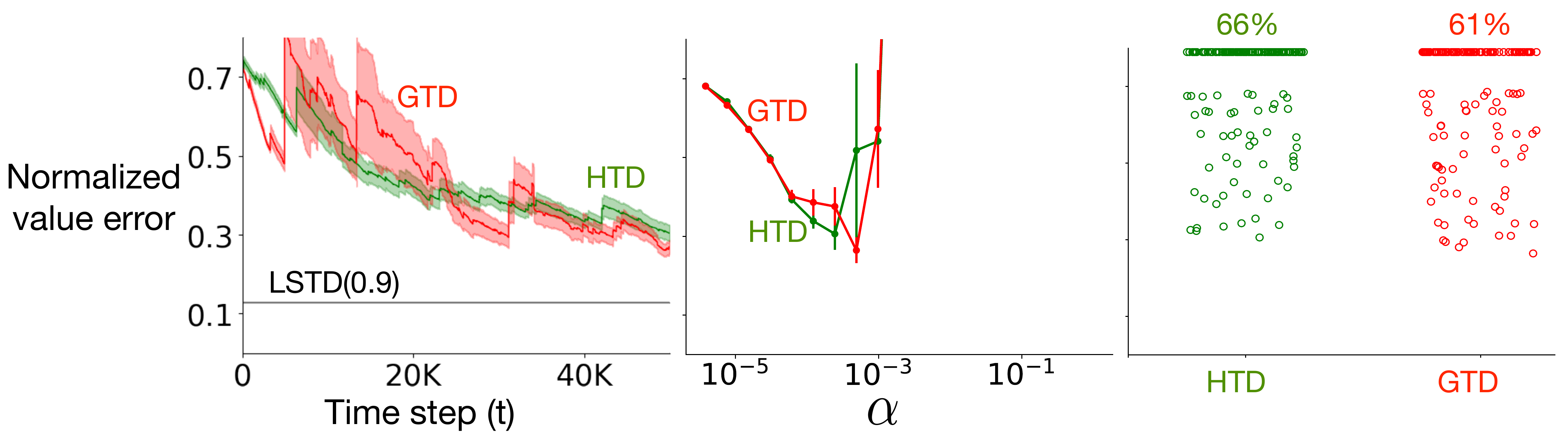}
      \caption{Comparing GTD(0.9), HTD(0.9) on the {\bf High Variance Four Rooms} problem. Free parameters are optimized for final performance. Overall, methods performed similarly; however, GTD(0.9) and GTD2(0.9) slightly outperformed the proximal methods (see the waterfall graph on the right). GTD(0.9) had very high variance in its final performance compared to other methods.}
      \label{fig:}
\end{figure}

\begin{figure}[H]
      \centering
      \includegraphics[width=0.8\linewidth]{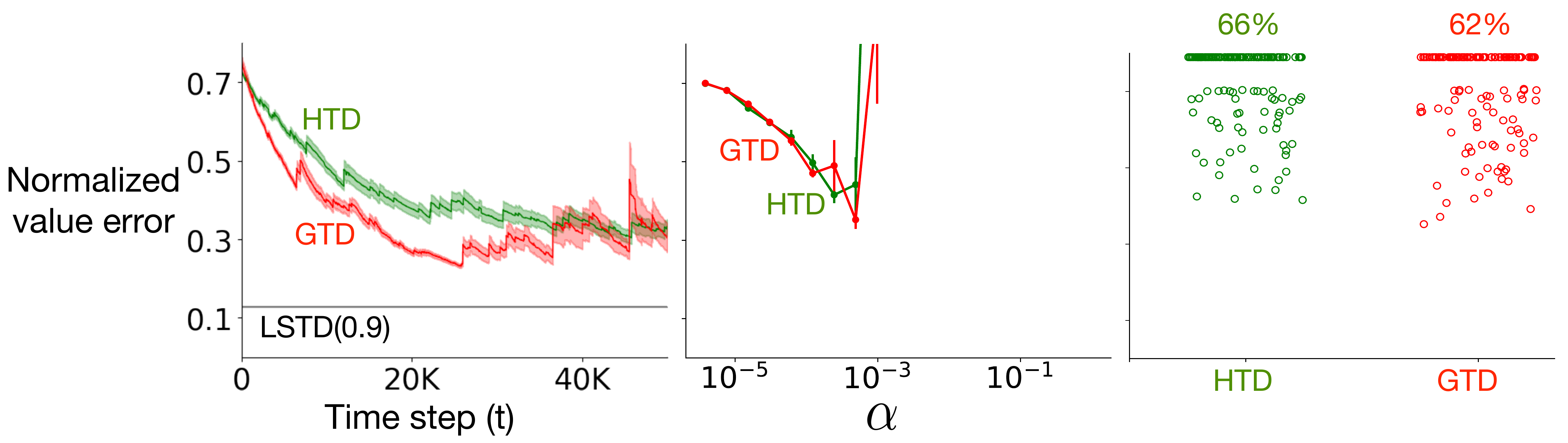}
      \caption{Comparing GTD(0.9), HTD(0.9) on the {\bf High Variance Four Rooms} problem. Free parameters are optimized for area under the learning curve. All methods had high variance when learning, though GTD(0.9) and GTD2(0.9) exhibited greater variance than Proximal GTD(0.9) and Proximal GTD2(0.9). The waterfall graph shows that GTD(0.9) outperformed other methods, followed by GTD2(0.9), Proximal GTD2(0.9), and Proximal GTD(0.9). Proximal GTD(0.9) diverged for most parameter settings, and had the worst final performance.}
      \label{fig:}
\end{figure}

\subsection{Investigating variance reduction for emphatic updates on high variance domain}
In Section \ref{sct:ComparingETDAndETDBeta}, we compared ETD($\lambda$) and ETD($\lambda, \beta$) on our two benchmark domains.
In Appendix \ref{app:ComparingETDAndETDBeta}, we provided all additional results obtained on these domains.
Here we compare ETD($\lambda$) with its lower variance successor, ETD($\lambda, \beta$), on the high variance domain.

ETD($\lambda, \beta$) outperforms ETD($\lambda$) on this domain, both with and without traces.
ETD($\lambda, \beta$) has much less sensitivity to choice of stepsize, and has a lower percentage of parameter settings for which it diverges.
Finally, ETD($\lambda, \beta$) converges faster than ETD($\lambda$) and to a lower final error.

\begin{figure}[H]
      \centering
      \includegraphics[width=0.8\linewidth]{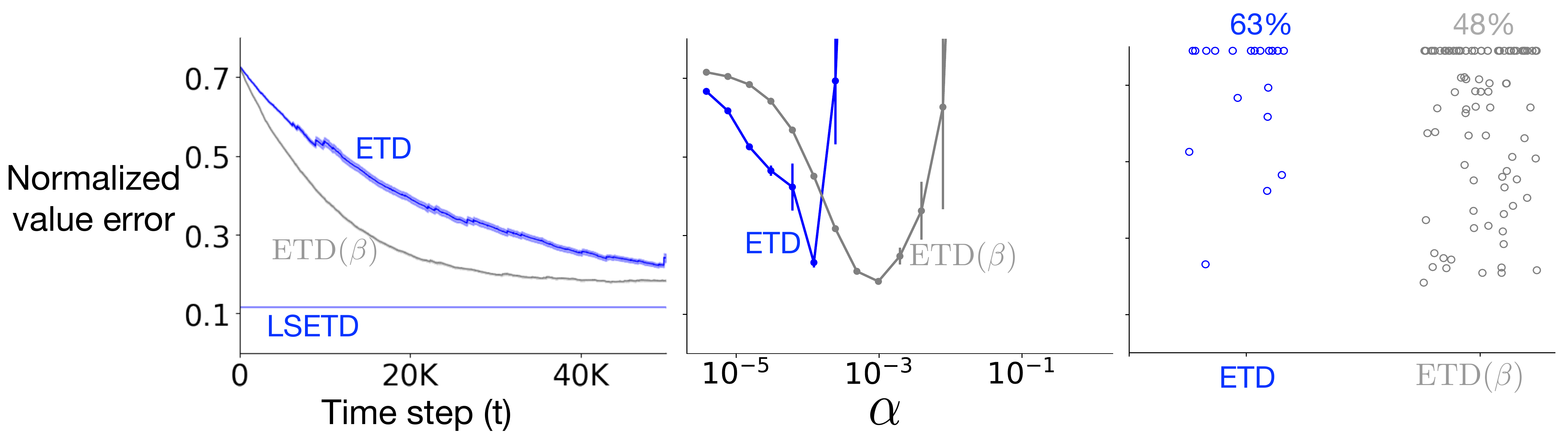}
      \caption{Comparing ETD(0), ETD(0, $\beta$) on the {\bf High Variance Four Rooms} problem. Free parameters are optimized for final performance. ETD(0, $\beta$) was faster than ETD(0) and converged to a final performance that was close to its best final performance for more parameter settings than for ETD(0). ETD(0) had very few parameter settings that achieved its best final performance.}
      \label{fig:}
\end{figure}

\begin{figure}[H]
      \centering
      \includegraphics[width=0.8\linewidth]{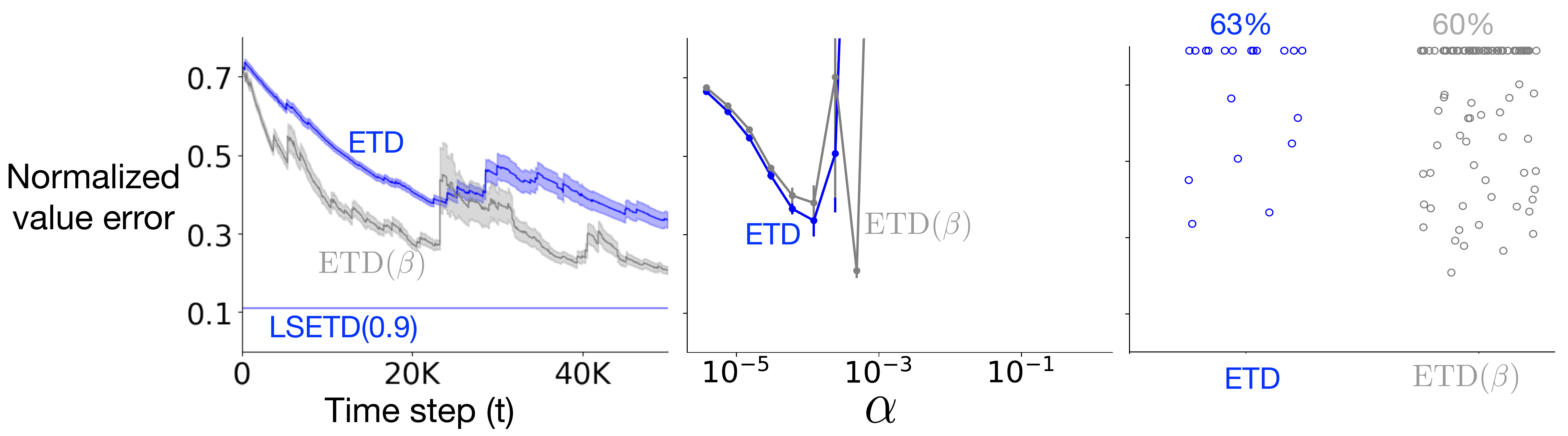}
      \caption{Comparing ETD(0.9), ETD(0.9, $\beta$) on the {\bf High Variance Four Rooms} problem. Free parameters are optimized for final performance. Both methods exhibit high variance, however, ETD(0.9, $\beta$) slightly outperformed ETD(0.9) in terms of speed and final performance. This is likely due to the fact that ETD($\lambda, \beta$) has an additional tunable parameters.}
      \label{fig:}
\end{figure}



\end{document}